\documentclass[10pt,journal,compsoc]{IEEEtran}
\usepackage{float}
\usepackage{url}
\usepackage{amsmath}
\usepackage{amsthm}
\usepackage{amsbsy}
\usepackage{enumitem, amssymb}
\usepackage{pifont}
\usepackage{dsfont}
\usepackage{graphicx}
\usepackage{makecell}
\usepackage[dvipsnames,table]{xcolor}
\usepackage{subcaption}
\usepackage{xspace}
\usepackage[export]{adjustbox}
\usepackage{tcolorbox}
\usepackage{tikz}

\makeatletter
\newcommand*\bigcdot{\mathpalette\bigcdot@{.5}}
\newcommand*\bigcdot@[2]{\mathbin{\vcenter{\hbox{\scalebox{#2}{$\m@th#1\bullet$}}}}}
\makeatother

\usepackage[first=0, last=9]{lcg}
\newcommand{\argmin}{\operatornamewithlimits{argmin}}

\newcommand{\cmark}{\ding{51}}
\newcommand{\xmark}{\ding{55}}

\newcolumntype{P}[1]{>{\centering\arraybackslash}p{#1}}
\newcolumntype{M}[1]{>{\centering\arraybackslash}m{#1}}

\newcommand{\simnormres}[1]{\ensuremath{#1}}
\newcommand{\simboldres}[1]{\ensuremath{\boldsymbol{#1}}}
\newcommand{\simsecores}[1]{\ensuremath{\underline{#1}}}

\newcommand{\changed}[1]{\textcolor{red}{#1}}

\newcommand{\normresvtwo}[2]{\ensuremath{#1} \footnotesize{(\ensuremath{#2})}}
\newcommand{\boldresvtwo}[2]{\ensuremath{\boldsymbol{#1}} \footnotesize{(\ensuremath{\boldsymbol{#2}})}}

\usepackage{booktabs}

\definecolor{lightgray}{gray}{0.95}

\newcommand{\quotationmarks}[1]{``#1''}

\newcommand{\XDER}{X-DER\xspace}
\newcommand{\miniimagenet}{\textit{mini}ImageNet\xspace}

\newcommand{\eg}{\textit{e.g.}\,\,}
\newcommand{\ie}{\textit{i.e.}\,\,}

\newcommand{\cubimage}[1]{\includegraphics[width=0.18\columnwidth]{images/cub/input/#1.png}}
\newcommand{\cubgradimage}[3]{\includegraphics[width=0.18\columnwidth]{images/cub/gradcam/#1_#2_#3.png}}

\usepackage{xstring}

\definecolor{color1bg}{HTML}{E15759}
\definecolor{color2bg}{HTML}{31A354}

\newcommand{\testprediction}[1]{%
\IfEq{#1}{1}{\begin{tikzpicture}\filldraw[fill=color2bg!40!white, draw=black] (0,0) rectangle (1.57,0.17);
\end{tikzpicture}}{\begin{tikzpicture}\filldraw[fill=color1bg!40!white, draw=black] (0,0) rectangle (1.57,0.17);
\end{tikzpicture}}
}

\renewcommand{\eg}{\textit{e.g.};\,\,}
\renewcommand{\ie}{\textit{i.e.};\,\,}

\renewcommand{\changed}[1]{#1}

\usepackage{algorithm}
\usepackage{algorithmic}
\usepackage{mathtools}
\usepackage{tikz}
\usepackage{twoopt}
\renewcommand{\changed}[1]{#1}

\usetikzlibrary{fit}

\newcommand\tikzmark[1]{%
  \tikz[remember picture,overlay]\node[inner xsep=0pt] (#1) {};}

\newcommandtwoopt\TextboxSepCE[5][2.5cm][5cm]{%
\begin{tikzpicture}[remember picture,overlay]
  \coordinate (aux) at ([xshift=#1]#4);
  \node[inner ysep=22pt,yshift=3.5ex,draw=black,thick,
    fit=(#3) (aux),baseline] 
    (box) {};
\end{tikzpicture}%
}

\newcommandtwoopt\TextboxDer[5][2.5cm][2cm]{%
\begin{tikzpicture}[remember picture,overlay]
  \coordinate (aux) at ([xshift=#1]#4);
  \node[inner ysep=15pt,yshift=0.6ex,draw=black,thick,
    fit=(#3) (aux),baseline] 
    (box) {};
\end{tikzpicture}%
}

\newcommandtwoopt\TextboxFutPrep[5][2.5cm][2cm]{%
\begin{tikzpicture}[remember picture,overlay]
  \coordinate (aux) at ([xshift=#1]#4);
  \node[inner ysep=18pt,yshift=2ex,draw=black,thick,
    fit=(#3) (aux),baseline] 
    (box) {};
\end{tikzpicture}%
}

\newcommandtwoopt\TextboxFutConstr[5][2.5cm][2cm]{%
\begin{tikzpicture}[remember picture,overlay]
  \coordinate (aux) at ([xshift=#1]#4);
  \node[inner ysep=53pt,yshift=1.3ex,draw=black,thick,
    fit=(#3) (aux),baseline] 
    (box) {};
\end{tikzpicture}%
}

\newcommandtwoopt\TextboxMemInsert[5][2.5cm][2cm]{%
\begin{tikzpicture}[remember picture,overlay]
  \coordinate (aux) at ([xshift=#1]#4);
  \node[inner ysep=20pt,yshift=2ex,draw=black,thick,
    fit=(#3) (aux),baseline] 
    (box) {};
\end{tikzpicture}%
}

\usepackage{xr}
\makeatletter
\newcommand*{\addFileDependency}[1]{
  \typeout{(#1)}
  \@addtofilelist{#1}
  \IfFileExists{#1}{}{\typeout{No file #1.}}
}
\makeatother

\newcommand*{\myexternaldocument}[1]{
    \externaldocument{#1}
    \addFileDependency{#1.tex}
    \addFileDependency{#1.aux}
}
\myexternaldocument{appendix_tpami_rev}

\ifCLASSOPTIONcompsoc
  \usepackage[nocompress]{cite}
\else
  \usepackage{cite}
\fi
\ifCLASSINFOpdf
\else
\fi
\begin{document}
\title{Class-Incremental Continual Learning\\into the eXtended DER-verse}

\author{Matteo~Boschini,~\IEEEmembership{Member,~IEEE,}
        Lorenzo~Bonicelli,
        Pietro~Buzzega,
        Angelo~Porrello,
        and~Simone~Calderara,~\IEEEmembership{Member,~IEEE}% <-this % stops a space
\thanks{The authors are with the Department of Engineering ``Enzo Ferrari'', The University of Modena and Reggio Emilia, Via Pietro Vivarelli 10,  41125 Modena, Italy.\protect\\
E-mail: \{matteo.boschini, lorenzo.bonicelli, angelo.porrello,\\ simone.calderara\}@unimore.it, pietrobuzzega@gmail.com.\\
Manuscript submitted for review.}%
% \thanks{Manuscript received April 19, 2005; revised August 26, 2015.}}
}
\IEEEtitleabstractindextext{%
\begin{abstract}
The staple of human intelligence is the capability of acquiring knowledge in a continuous fashion. In stark contrast, Deep Networks forget catastrophically and, for this reason, the sub-field of Class-Incremental Continual Learning fosters methods that learn a sequence of tasks incrementally, blending sequentially-gained knowledge into a comprehensive prediction.

This work aims at assessing and overcoming the pitfalls of our previous proposal Dark Experience Replay (DER), a simple and effective approach that combines rehearsal and Knowledge Distillation. Inspired by the way our minds constantly rewrite past recollections and set expectations for the future, we endow our model with the abilities to \textit{i)} revise its replay memory to welcome novel information regarding past data \textit{ii)} pave the way for learning yet unseen classes.

We show that the application of these strategies leads to remarkable improvements; indeed, the resulting method -- termed eXtended-DER (\XDER) -- outperforms the state of the art on both standard benchmarks (such as CIFAR-100 and \miniimagenet) and a novel one here introduced. To gain a better understanding, we further provide extensive ablation studies that corroborate and extend the findings of our previous research (\eg the value of Knowledge Distillation and flatter minima in continual learning setups).

We make our results fully reproducible; the codebase is available at \url{https://github.com/aimagelab/mammoth}.
\end{abstract}

}

\maketitle

\IEEEdisplaynontitleabstractindextext

% \fi
\IEEEpeerreviewmaketitle

\IEEEraisesectionheading{\section{Introduction}\label{sec:introduction}}

\IEEEPARstart{H}{uman} intelligence allows us to acquire new knowledge about the surrounding world in a natural way. Thanks to the extraordinary and still unclear capabilities of the human brain, we can learn new and complex tasks (\eg driving cars) and, at the same time, remember the old ones (\eg cycling) without experiencing either interference or forgetting. Moreover, human beings exhibit an ability called \textbf{fluid intelligence}~\cite{cochrane2019fluid}: according to this construct, humans can reason about and engage with novel problems in a manner that does not explicitly rely on prior learning. Namely, we can recognize new patterns and plan new strategies in unseen environments in a manner that only minimally depends upon specific previous experience or acculturation.

Despite the long-standing parallelism between the human brain and Artificial Neural Networks (ANNs), the latter do not support these abilities and struggle~\cite{mccloskey1989catastrophic}: indeed, novel knowledge tends to overwrite the old one, thus leading to a disruptive degradation of performance in previous tasks. Such an issue -- which is widely known as \textit{catastrophic forgetting}~\cite{mccloskey1989catastrophic} -- currently represents a hindrance towards the broad applicability of ANNs: to overcome this limitation, the field of Continual Learning (CL) includes a wide array of methods to let ANNs retain their performance~\cite{de2019continual,parisi2019continual}.
Whether by loss terms that prevent the model from changing~\cite{kirkpatrick2017overcoming, zenke2017continual, li2017learning}, by explicitly using distinct parts of the model at distinct times~\cite{rusu2016progressive, mallya2018packnet}, or by revisiting past data~\cite{french1991using, rebuffi2017icarl, aljundi2019gradient}, CL approaches let current training be influenced by previously learned information to preserve it.

The exploitation of an episodic memory (\ie a subset of past data that are continuously revisited) is undoubtedly one of the most reliable ways to face the aforementioned problem~\cite{aljundi2019gradient,prabhu2020gdumb,buzzega2020dark}. Due to its effectiveness, a plethora of approaches deal with its design and differ in the following facets: \textit{when} the memory has to be populated~\cite{chaudhry2019tiny} (\eg at discrete intervals \textit{vs} continuously); \textit{which} elements we keep in memory and \textit{which} ones we move out~\cite{aljundi2019online,aljundi2019gradient,buzzega2020rethinking}; \textit{what} kind of regularization we apply on these examples~\cite{riemer2018learning,lopez2017gradient} and, consequently, \textit{what} information we store in it (\eg also old model responses~\cite{buzzega2020dark}). However, there is a promising direction that is still unexplored: \textit{how} the memory has to be updated, \ie the \textbf{rewriting} of past experiences to meet new insights regarding old events. This activity is peculiar to human beings~\cite{bridge2014hippocampal,marla2014how}: memory, indeed, is not to be understood as a mere video-camera recording events; instead, the hippocampus reframes past events to create a tale that fits the current world, thus helping us to take good decisions and focus on what is important in the here and now. Such a capability represents a source of inspiration for this work: in fact, we firstly extend our previous proposal~\cite{buzzega2020dark} -- called \textbf{Dark Experience Replay (DER)} -- by equipping its memory with an update procedure that implants information from the present into the retained memories.

The manner memory is kept up-to-date is nevertheless only one of the two directions this work investigates. Recent studies highlight that the episodic memory also plays a key role in the mental simulation of future events~\cite{addis2007remembering,schacter2007remembering}; in other words, previously learned concepts influence our expectations about the future. We try to replicate this effect in our CL algorithm by further devising a \textbf{future preparation} strategy: \ie a technique that exploits past and present data to prepare future classification heads to accommodate meaningful information.

To sum up, this work identifies some shortcomings in the way DER~\cite{buzzega2020dark} organizes present and future knowledge. Consequently, we address them by a two-fold enhancement: on the one hand, we propose a procedure that maintains the memory buffer up-to-date by inserting secondary information from the present into memories of the past; on the other, we evaluate the benefits of preparing the underlying model to incoming tasks. Thanks to these improvements, our revised method -- which we call \textbf{eXtended-DER} (\textit{a.k.a.} \textbf{\XDER}) -- achieves a remarkable increase in accuracy w.r.t.\ the current state of the art on two standard benchmarks (Split CIFAR-100 and Split \miniimagenet) and on the newly introduced Split NTU-60.

In addition to extensive ablation studies highlighting the rationale behind our intuitions, we remark the following contributions:
\begin{itemize}
    \item We shed light on some pitfalls of our previous proposal and, on this basis, propose \XDER, a novel CL method that embraces memory update and future preparation.
    \item In light of recent advances regarding the foundations of Knowledge Distillation~\cite{menon2021statistical}, we review and provide new insights on the benefits of logits-replay against catastrophic forgetting.
    \item We deepen the discussion of our previous work about the geometry of local minima in CL, thus strengthening what we and other authors~\cite{mirzadeh2020understanding} have recently stated. From this perspective, we conduct several evaluations on \XDER and show that it favorably attains flatter minima.
\end{itemize}

\section{Related Work}
\label{sec:rel}
\subsection{Continual Learning}
Recent years have witnessed a surge in the interest for methods that can alleviate the phenomenon of catastrophic forgetting~\cite{mccloskey1989catastrophic}, where the goal is to obtain a model that can adapt to changes in the distribution of input data (plasticity) while retaining the previously learned knowledge (stability).
This problem is modeled in literature by means of different \textit{settings}~\cite{van2019three}, which typically unfold a base classification problem in successively learned \textit{tasks}.
In the \textbf{task incremental} setting (\textbf{Task-IL} or \textbf{multi-head}), the learner must learn and remember how to classify within each task, as it is given access to the task identity of test samples at inference time.
\textit{Vice-versa}, the \textbf{class incremental} scenario (\textbf{Class-IL} or \textbf{single-head}) requires the task identifier to be predicted along with the sub-class label. Lastly, the \textbf{domain incremental} setting (\textbf{Domain-IL}) does not alter the distribution of classes; instead, it characterizes task changes by introducing a shift in the input distribution. 

Methods specifically designed to tackle the Task-IL scenario usually exhibit a multi-headed architecture~\cite{rusu2016progressive} or use the provided task information to map portions of the model to specific tasks~\cite{serra2018overcoming}. While these strategies often prove effective, their use is limited to the multi-head setting.

By contrast, recent CL works focus mainly on Class-IL~\cite{aljundi2019gradient,caccia2021reducing,mittal2021essentials,belouadah2019il2m,hou2019learning}, as it is more general and regarded as more realistic than the other scenarios~\cite{farquhar2018towards,aljundi2019gradient}. 
Methods capable of working in a single-head assumption are usually divided into two families: \textit{regularization-based} and \textit{rehearsal-based}.

The former use specifically designed regularization terms to lead the optimization towards a good balance between stability and plasticity. Typically, they introduce a penalty term that discourages alterations in the weights that are vital for the previous tasks while the new ones are being learned~\cite{kirkpatrick2017overcoming,zenke2017continual}. These methods can be effective for short sequences of tasks but usually fail to scale to complex problems~\cite{farquhar2018towards, aljundi2019gradient}.

On the other hand, \textit{rehearsal} models take advantage of a memory buffer to store exemplar elements from previous distributions.
Experience Replay~\cite{ratcliff1990connectionist,robins1995catastrophic} (ER) simply replays the stored elements along with the input stream to simulate training over an independent and identically distributed task (joint training). Despite its simplicity, this method has proven to be highly effective even with a minimal memory footprint~\cite{chaudhry2019tiny} and serves as the basis for recent methods that propose variations on the strategies for the selection of samples to include in the memory buffer~\cite{aljundi2019gradient} or the sampling of examples from it~\cite{aljundi2019online}. The retained knowledge can also be used as a mean to revise the optimization procedure: MER~\cite{riemer2018learning} employs meta-learning to discourage interference and maximize knowledge transfer between tasks, while GEM~\cite{lopez2017gradient} and A-GEM~\cite{chaudhry2018efficient} use old training data to minimize the gradient interference in an explicit fashion.

\subsection{Self-Distillation}
\textbf{Knowledge Distillation (KD)}~\cite{hinton2015distilling} is a training methodology that allows transferring the knowledge of a teacher model into a separate student model. While \textit{Hinton et al.\ }originally proposed to distillate large teachers -- possibly ensembles -- into smaller students, further studies revealed additional interesting properties about this technique. In particular, \textit{Furlanello et al.}~\cite{furlanello2018born} show that multiple rounds of distillation between models with the same architecture (termed \textbf{self-distillation}) can surprisingly improve the performance of the student. More recently, other works~\cite{zhang2019your,luan2019msd} explore an interesting variation of self-distillation that distills knowledge from the deeper layers of the network to its shallower ones to accelerate convergence and attain higher accuracy.

\textbf{Knowledge Distillation and Continual Learning.}~Distillation can be used to hinder catastrophic forgetting by appointing a previous snapshot of the model as teacher and distilling from it while new tasks are learned. Learning Without Forgetting~\cite{li2017learning} uses teacher responses to new exemplars to constrain the evolution of the student. Several other works combine distillation with \textit{rehearsal}: iCaRL~\cite{rebuffi2017icarl} distills the responses of the model at the previous task boundary, learning latent representations to be used in a \textit{nearest mean-of-exemplars} classifier; EtEIL~\cite{castro2018end}, LUCIR~\cite{hou2019learning} and BiC~\cite{wu2019large} focus on contrasting the prediction bias that comes from incremental classification; IL2M~\cite{belouadah2019il2m} stores additional statistics to facilitate distillation and compensate bias.

\section{background}
\label{sec:background}
\subsection{Class-Incremental Continual Learning}
In Class-Incremental Continual Learning (CiCL), a model $f(\cdot;\theta)$ is trained on a sequence of $T$ tasks $\{\mathcal{T}_0, \dots, \mathcal{T}_{T-1}\}$, having access to one at a time. The $i^{\text{th}}$ task consists of datapoints $\{x_i^{(n)}, y_i^{(n)} \}_{n=1}^{|\mathcal{T}_i|}$, where $y_i^{(n)} \in \mathcal{Y}_i$, with disjoint ground-truth values for different tasks, \ie $\mathcal{Y}_i\cap \mathcal{Y}_j=\varnothing \ \text{s.t.} \ i\neq j$. For the sake of simplicity, we assume all tasks having the same number of classes, \ie $|\mathcal{Y}_i| = |\mathcal{Y}_j| = |\mathcal{Y}|$. While all $x_i$ are \textit{i.i.d.}\ within $\mathcal{T}_i$, the overall training procedure does not abide by the \textit{i.i.d.}\ assumption, as input distribution shifts between tasks and labels change. 
The objective of CiCL is the minimization of the risk over all tasks:
\begin{equation}
\label{eq:cont}
    \mathcal{L}_{\text{CiCL}} \triangleq \sum_{i=0}^{T-1}{
    \mathop{{\mathds{E}}}_{(x,y) \sim \mathcal{T}_i}\big[\mathcal{L}(f(x; \theta), y)}\big],
\end{equation}
where $\mathcal{L}$ stands for the loss (\eg the categorical cross-entropy) of predicting $f(x; \theta)$ given $y$ as the true label. The optimal solution $\theta^{*}$ should provide accurate predictions for all tasks; however, this has to be pursued by observing one task at a time. To account for this, its actual learning objective should combine the empirical risk on the current task $\mathcal{T}_c$ with a separate regularization term $\mathcal{L}_{R}$:
\begin{equation}
\label{eq:true_cont}
    \hat{\mathcal{L}}_{\text{CiCL}} \triangleq \mathop{\mathds{E}}_{(x,y) \sim \mathcal{T}_c}{\big[\mathcal{L}(f(x; \theta), y)\big]} + \mathcal{L}_{R}.
\end{equation}
The second term $\mathcal{L}_{R}$ serves a twofold purpose: \textit{i)} it prevents the model from forgetting past knowledge while fitting new data~\cite{robins1995catastrophic}; \textit{ii)} it encourages the learner to gather per-task classifiers into a single and harmonized one~\cite{hou2019learning}.
\subsection{A Self-KD Approach: Dark Experience Replay}
\label{sec:der}
To design the regularization term of Eq.~\ref{eq:true_cont} for preserving the capabilities on old tasks, the approaches~\cite{rebuffi2017icarl,li2017learning} based on Knowledge Distillation ($\operatorname{KD}$) use past snapshots $f(\cdot; {\theta}^{(i)})$ as teachers for the model engaged on the current task $c$:
\begin{equation}
    \mathcal{L}_{R} = \sum_{i=0}^{c - 1}{\mathop{\mathds{E}}_{x \sim \mathcal{T}_i}\big[\operatorname{KD}(f(x; \theta^{(i)}), f(x; \theta)})\big].
\end{equation}
Logit matching~\cite{hinton2015distilling} constitutes a straightforward and effective approach to pursue the objective above:
\begin{equation}
    \label{eq:kd}
    \mathcal{L}_{R} = \sum_{i=0}^{c - 1}\mathop{\mathds{E}}_{x \sim \mathcal{T}_i}\big[||(f(x; \theta^{(i)}) - f(x; \theta))||^2_2\big].
\end{equation}
Notably, Eq.~\ref{eq:kd} violates CiCL, as it assumes the availability of data-points from previous tasks. To approximate it, our previous proposal~\cite{buzzega2020dark} -- termed \textbf{Dark Experience Replay} (DER) -- introduced a small replay buffer $\mathcal{M}$ that stores a limited amount of past examples $x$ along with the model outputs $\ell \triangleq f(x, \theta^{(t)})$, where $t$ indicates the time of memory insertion. Eq.~\ref{eq:kd} can therefore be recast as:
\begin{equation}
    \label{eq:der}
    \mathcal{L}_{R} = \mathcal{L}_{\operatorname{DER}} = \alpha \cdot \mathds{E}_{(x,\ell) \sim \mathcal{M}}\big[||(\ell - f(x; \theta))||^2_2].
\end{equation}
\noindent We further introduced \textbf{Dark Experience Replay++} (DER++), which replays both logits and ground-truth labels:
\begin{equation}
\mathcal{L}_{R} = \mathcal{L}_{\operatorname{DER++}} = \mathcal{L}_{\operatorname{DER}} + \beta\cdot\mathop{\mathds{E}}_{(x,y) \sim \mathcal{M}}\big[\mathcal{L}(f(x; \theta), y)].
\label{eq:loss_derpp}
\end{equation}
\noindent\textbf{Pre-allocation of future heads.}~In several CiCL works, the model ends with a linear layer projecting into a space with as many features as classes have been encountered so far: as the number of the latter increases one task after the other, some approaches~\cite{hou2019learning,li2017learning} extend the classifier by instantiating a new classification layer (\textit{a.k.a.} \textbf{head}); while some others~\cite{riemer2018learning,rebuffi2017icarl} initialize the network by providing as many classification heads as tasks are encountered from the beginning to the end\footnote{Sec.~\ref{sec:preallocation} shows that knowing in advance the total number of tasks does not represent a crucial hypothesis and can be easily removed.}. Both our previous proposals, DER(++), belong to this second line of approaches: indeed, they provide all necessary classification heads from the beginning and let their parameters be subject to optimization. It is noted that this practice -- which we show in the following opens up interesting possibilities -- is not entirely novel to the field:~\cite{aljundi2019gradient, aljundi2019online} involve a cross-entropy term spanning over all classes (hence, already seen and yet unseen), which silences the activity of future heads (as we discussed later).
\subsection{Limitations of Dark Experience Replay}
\label{sec:limitations}
\begin{figure}[t]
    \centering
    \includegraphics[width=0.98\linewidth]{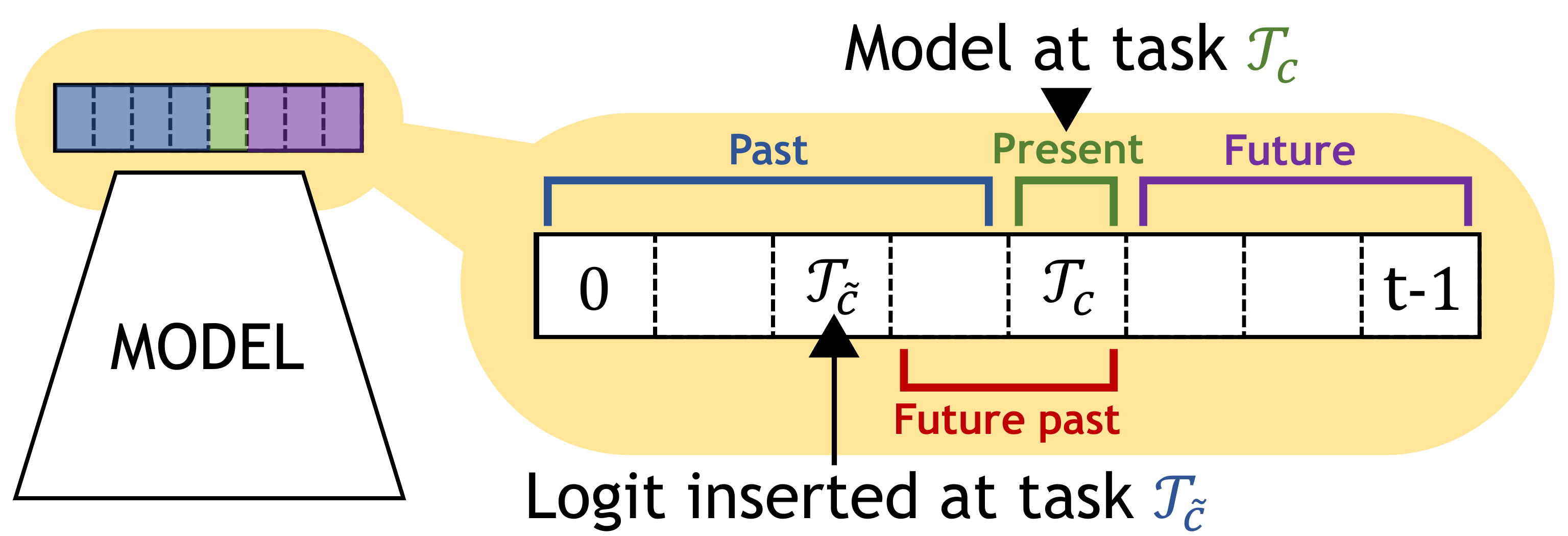}
    \caption{A view of CiCL timelines. $\mathcal{T}_c$ is the task that is currently being learned by the model; $\mathcal{T}_{\tilde{c}}$ indicates the task at which an example entered the memory buffer.}
    \label{fig:timelines}
    \vspace{-1em}
\end{figure}
\textbf{Terminology.} To facilitate the discussion, we introduce a categorization that splits the output space of the model at task $\mathcal{T}_c$ into the following partitions (see also Fig.~\ref{fig:timelines}):
\vspace{0.3em}
\begin{center}
\begin{tabular}{p{1.8cm}p{6.2cm}} 
\makecell[tc]{\textbf{Past}\\\large{$\ell_{\operatorname{pa}[c]}$}} & $\operatorname{pa}[c] \equiv \{0,1,\dots, c\cdot|\mathcal{Y}| - 1\}$ \hfill\hfill\hfill\hfill\vspace{0.3em}\linebreak \ie logits modeling the probabilities of classes observed \textbf{up to} the current task.\\ 
\end{tabular}
\vspace{0.3em}
\begin{tabular}{p{1.8cm}p{6.2cm}}
\makecell[tc]{\textbf{Present}\\\large{$\ell_{\operatorname{pr}[c]}$}} & $\operatorname{pr}[c] \equiv \{c\cdot|\mathcal{Y}|,\dots,(c+1)\cdot|\mathcal{Y}| - 1\}$ \hfill\hfill\hfill\hfill\vspace{0.3em}\linebreak \ie logits of the head associated to the \textbf{current} task $c$.\\
\end{tabular}
\vspace{0.3em}
\begin{tabular}{p{1.8cm}p{6.2cm}}
\makecell[tc]{\textbf{Future}\\\large{$\ell_{\operatorname{fu}[c]}$}} &$\operatorname{fu}[c] \equiv \{(c+1)\cdot|\mathcal{Y}|,\dots,T\cdot|\mathcal{Y}|-1\}$ \hfill\hfill\hfill\hfill\vspace{0.3em}\linebreak \ie logits corresponding to \textbf{unseen} classes. They are not useful for classifying examples seen thus far, but will be needed during the following tasks.\\
\end{tabular}
\end{center}
It is noted that the composition of these partitions depends upon the specific task the model is learning (some logits move from one partition to the other when passing to the subsequent task). 
\begin{figure*}[t]
    \begin{minipage}{.5\linewidth}
    \centering
    \subfloat[Considering only examples of previous tasks misclassified by DER++, the percentages of predictions won by each classification head. We omit later tasks as they entail the same issue.]{\label{fig:derpperrors}\includegraphics[width=.85\linewidth]{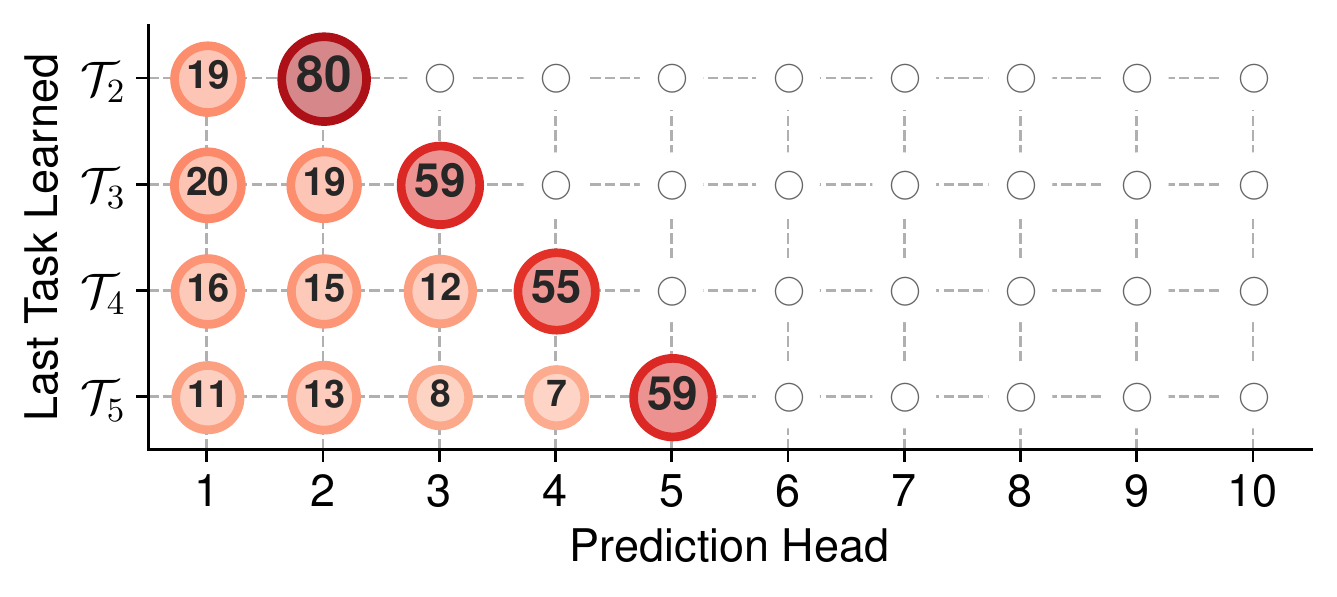}}
    \end{minipage}%
    \begin{minipage}{.5\linewidth}
    \centering
    \subfloat[Average norm of DER++ gradients w.r.t. stream and buffer data: the contribution of new examples on the stream significantly outweighs the one of replay items.]{\label{fig:derppgradients}\includegraphics[width=.85\linewidth]{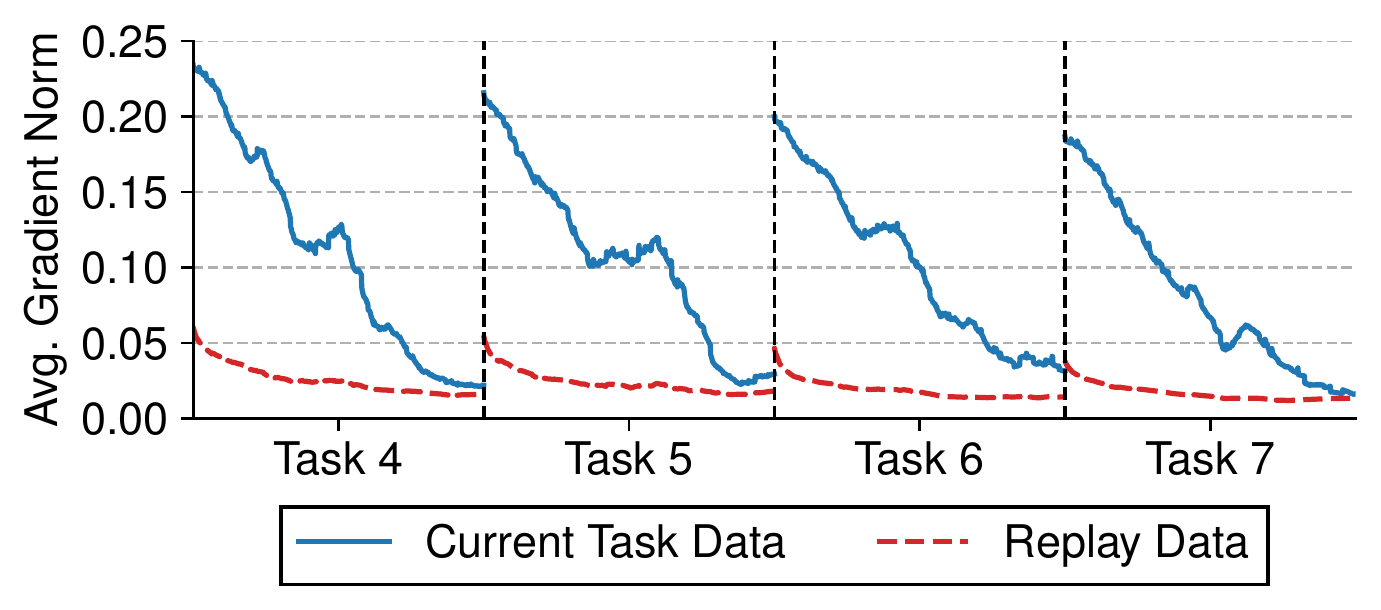}}
    \end{minipage}
    \vspace{-0.3em}
    \centering
    \subfloat[Average logit values stored in the memory buffer by DER++ (left) and \XDER (right) for the first 50 classes of Split CIFAR-100. For DER++, future logits (\textit{red}) feature values that are strongly biased towards negative values. On the account of future preparation adopted by \XDER, its future logits (\textit{green}) are comparable to past ones (\textit{blue}).]{\label{fig:derpplogits}\includegraphics[width=.9\linewidth]{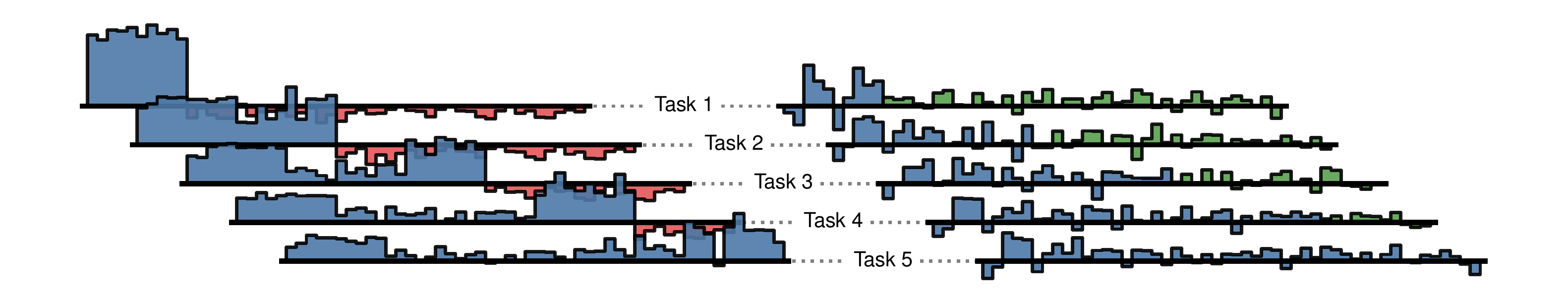}}
    \caption{Three visuals that depict some facets of bias accumulation inherent Dark Experience Replay++.}
    \label{fig:weak}
    \vspace{-0.5em}
\end{figure*}
We then identify \textbf{only for buffer data-points} the logits of \textbf{future past}: namely, the set of classes the model discovered after the example was inserted in $\mathcal{M}$.
\begin{center}
    \begin{tabular}{p{1.8cm}p{6.2cm}}
    \rowcolor{lightgray}
    \makecell[tc]{\textbf{Future past}\\
    \large{$\ell_{\operatorname{fp}[c;j]}$}\vspace{0.7em}\\$j \in \{\tilde{c}+1,$ \\ $\dots,\ c\}$}
    & $\operatorname{fp}[c;j] \equiv \{j\cdot|\mathcal{Y}|,\dots,(j+1)\cdot|\mathcal{Y}| - 1\}$ \hfill\hfill\hfill\hfill\vspace{0.3em}\linebreak given an $(x, y, \ell) \in \mathcal{M}$ stored at task $\tilde{c}$ ($\tilde{c} < c$), these logits model the classes of the $j^{\text{th}}$ task discovered \textbf{after} the insertion of the example into the buffer.\\
    \end{tabular}    
\end{center}
\noindent We now discuss two weaknesses of DER(++) concerning the partitions of future and future past logits. Afterwards -- in Sec.~\ref{sec:mod} -- we discuss of those could be overcome.

\vspace{0.5em}
\noindent\textbf{(L1) DER(++) have a blind spot for future past}.~Our previous work showed that the memory buffer of DER(++) provides a more efficient and informative way to refresh old tasks. However, we acknowledge here that the information it carries is limited solely to the classes already seen at the time an example was inserted in $\mathcal{M}$.

Indeed, when inserting an example in the rehearsal memory, it stands to reason that past and present logits encode all the information useful for later replay. However, by the time we move to subsequent tasks, the model discovers new classes and, with them, their relations with the old ones (\ie future past information). Unfortunately, DER(++) replaying do not profit from this incoming information, as they pin as target a version of future past logits that precedes the effective observation of the corresponding classes; differently, we could \textbf{update} the memory buffer to capture this emerging knowledge. As reported in the following, this delivers a remarkable effect in the prevention of forgetting. \changed{For an in-depth experiment showing that DER++ is blind towards future past classes and a comparison with our proposal, we refer the reader to App.~\ref{app:l1otherexamples}}.

It is noted that this limitation does not apply to those distillation-based models that use previous network checkpoints to compute the regularization objective~\cite{li2017learning,rebuffi2017icarl,hou2019learning}. In fact, as the teacher is updated at every task boundary, future past logits are naturally made available to the student network. The downside of this strategy, however, is that the update does not only concern future past logits, but also past ones; thus leading the teacher itself to forget.

\vspace{0.5em}
\noindent\textbf{(L2) DER(++) overemphasize the classes of the current task}. Several works~\cite{wu2019large,ahn2020ssilss,caccia2021reducing,mittal2021essentials} have recently shed light on the accumulation of bias towards present classes and the negative impact it has on performance. We have found that also DER(++) are prone to such a pitfall: similarly to~\cite{wu2019large}, we can quantitatively characterize it by evaluating how predictions distribute across different classification heads (as training progresses). In particular, we limit the analysis on misclassified examples belonging to tasks prior to the current one: Fig.~\ref{fig:derpperrors} highlights which task comprises the predicted class (on average); as can be seen, the majority of wrong predictions end up in the last observed task. 

On the one hand, the negative bias towards past classes can be ascribed to the optimization of the cross-entropy loss on examples from the current task. As pointed out in~\cite{caccia2021reducing}, when a new task is presented to the net, an asymmetry arises between the contributions of replay data and current examples to the weights updates: indeed, the gradients of new (and poorly fit) examples outweigh (Fig.~\ref{fig:derppgradients}). If we aim at learning the current task solely, this is desirable as it favorably dampens logits of past classes. However, a hasty attenuation of earlier classes clashes with the second goal of avoiding forgetting of past concepts. In order to achieve a unified classifier, it is important to take countermeasures against such a phenomenon.
\begin{figure*}[t]
    \centering
    \includegraphics[width=0.9\linewidth,keepaspectratio]{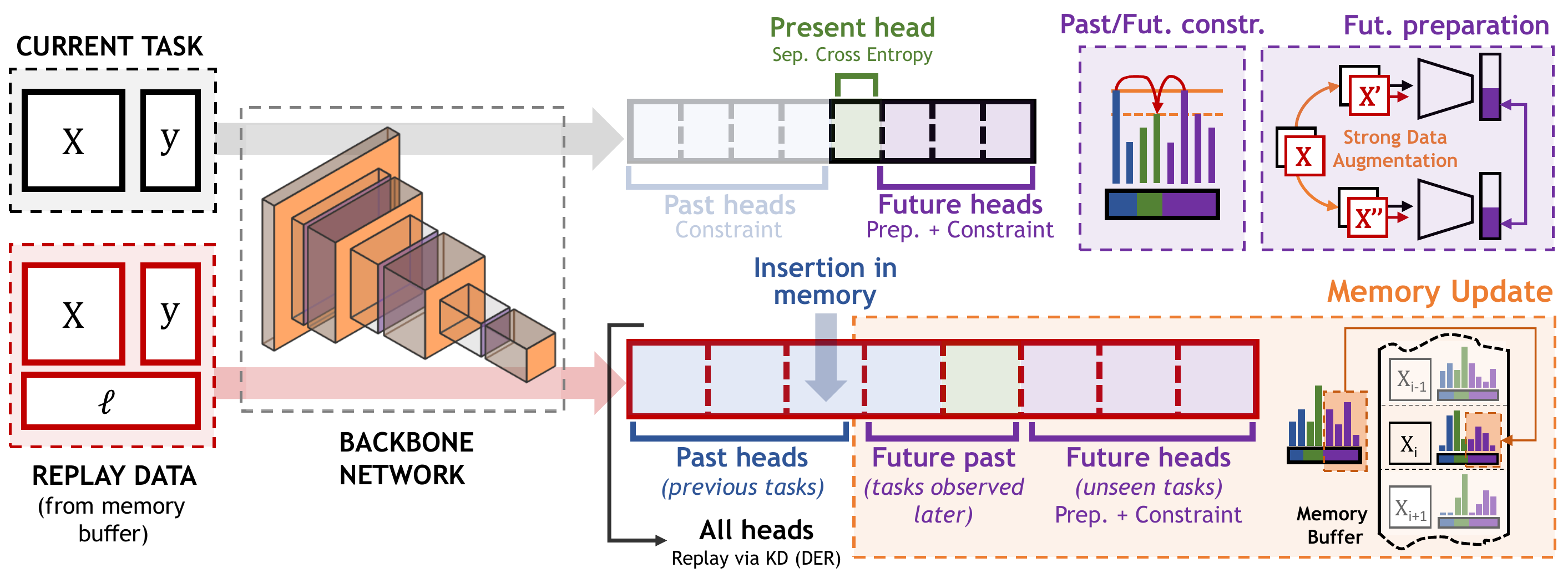}
    \caption{
    % Given a mix of examples from the current task and the memory buffer $\mathcal{M}$, 
    \XDER reckons on distinct objectives for different partitions of the output space: \textit{i}) it applies the Cross-Entropy loss in a way that isolates the head of the current task; \textit{ii}) it relieves forgetting by applying Knowledge Distillation on examples from $\mathcal{M}$; \textit{iii}) it warms (future) logits tied to unseen classes; Meanwhile, the predictions stored in $\mathcal{M}$ are regularly updated to deal with secondary information (\quotationmarks{future past}) relating old examples with the classes emerging later on the stream.
    }\vspace{-0.5em}
    \label{fig:overview}
\end{figure*}

Similarly, we observe a consistent negative bias towards future classes. We ascribe this behavior to the cross-entropy loss: as its application spans all heads, the future ones always have zeroes as post-softmax targets and hence are strongly pushed towards pre-softmax negative values. On the one hand, this desirably prevents the model from trivial errors; however, we mean future heads to accommodate the learning of future tasks. Therefore -- if the negative bias accumulates so strongly on these heads -- the recovery from that situation may slow down and complicate the learning of new tasks. In this regard, Fig.~\ref{fig:derpplogits} illustrates the behavior of future logits and compares the average responses of both DER++ (Fig.~\ref{fig:derpplogits}, left) and the approach discussed in Sec.~\ref{sec:mod} (Fig.~\ref{fig:derpplogits}, right). As can be seen, the former consistently exhibits negative values for unseen classes; on the contrary, the latter avoids bias accumulation on the account of the regularization it imposes on future logits.

\section{Proposed Approach}
\label{sec:mod}
In this section, we discuss how the above-discussed limitations of DER(++) can be addressed. We refer the reader to Fig.~\ref{fig:overview} for a visual overview of the model thus enhanced, which we dub \textbf{eXtended-DER} (\textit{a.k.a.} \textbf{\XDER}).
\subsection{\XDER: Logits of Future Past}
\label{sec:xder}
To prevent DER(++) from losing valuable secondary information, we devise a simple procedure that keeps its memory buffer updated. Let us suppose the model is learning the $c^{\text{th}}$ task and an example $(x, y, \ell^{{\scriptscriptstyle{\mathcal{M}}}}) \in \mathcal{M}$ from a previous $\tilde{c}^{\text{th}}$ task is sampled from the memory buffer for replay. The current network output $\ell \triangleq f(x; \theta)$ now contains the secondary information of task $\mathcal{T}_c$ for $x$: therefore, we propose to \textbf{implant} the corresponding logits $\ell_{\operatorname{fp}[c;c]}$ into the memory entry containing $\ell^{{\scriptscriptstyle{\mathcal{M}}}}$. Such an operation only involves the head of the current task and is applied both while learning (in an ongoing manner) and at the end of it.

From a technical perspective, we do not simply overwrite previous logits with the new ones. Indeed, as the net suffers from bias towards present classes, simply implanting their values in the memory buffer and using these for later replay would exacerbate the issue even more. Instead, we take care of re-scaling the portion tied to future past in a way such that its maximum logit $\ell_{\operatorname{fpmax}} = \operatornamewithlimits{max}_{j\in{\operatorname{fp}[c;c]}} \ell_j$ is lower than the ground-truth one $\ell^{\scriptscriptstyle{\mathcal{M}}}_{\operatorname{gt}} = \operatorname{one-hot}(y)\bigcdot\ell^{\scriptscriptstyle{\mathcal{M}}}$ already in memory. Formally:
\begin{equation}
    \ell^{\scriptscriptstyle{\mathcal{M}}}_{k} \longleftarrow \ell_{k} \cdot \operatorname{min} (\gamma \frac{\ell^{\scriptscriptstyle{\mathcal{M}}}_{\operatorname{gt}}}{\ell_{\operatorname{fpmax}}}, 1),\quad k \in \operatorname{fp}[c;c]
\end{equation}
where $\gamma\in [0,1]$ is a hyperparameter controlling the attenuation rate (which we typically set to $0.75$).
\subsection{Future Preparation}
\label{sec:futurepreparation}
\begin{figure}[t]
    \centering
    \includegraphics[width=\linewidth,keepaspectratio]{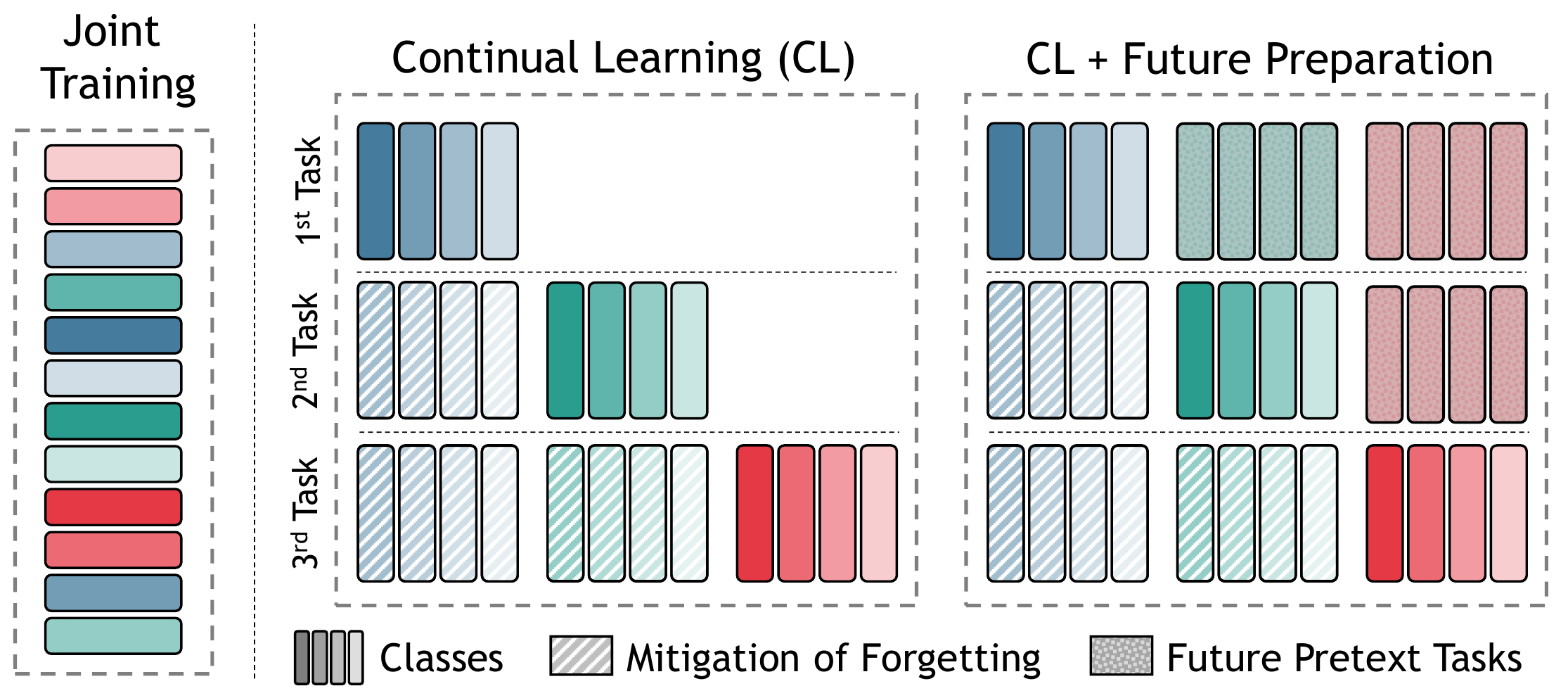}
    % \vspace{-0.5em}
    \caption{Unlike standard approaches, \XDER introduces pretext tasks for anticipating unseen classes.}
    % \vspace{-1em}
    \label{fig:jointsimulation}
\end{figure}

Most CiCL methods exploit the information available up to the current task to prevent the leak of past knowledge. Here, we take an extra step and argue that the same care should be placed on preparing future heads to accommodate future classes. In this respect, Fig.~\ref{fig:jointsimulation} depicts the underlying intuition: considering the joint training on all tasks (Fig.~\ref{fig:jointsimulation}, left) as the optimal solution we have to approximate, standard CL approaches (Fig.~\ref{fig:jointsimulation}, center) seem to focus only on a (growing) part of the overall problem, \ie what concerns the tasks seen up to the current one, as embodied by Eq.~\ref{eq:true_cont}. Instead, we claim that even a coarse guess regarding unseen tasks can lead to a better estimate of Eq.~\ref{eq:cont}.

To the best of our knowledge, \XDER is the first method pursuing this goal through optimization (Fig.~\ref{fig:jointsimulation}, right): it conjectures about the semantics of logits corresponding to unseen classes, which are encouraged to be consistent across instances of the same class. As outlined by the field of contrastive self-supervised learning~\cite{chen2020simple,zbontar2021barlow}, the skillful use of data augmentation techniques can lead towards useful representations even when no label information is made available to the learner. In a sense, this resembles the case of our future classes: hence, we expect it to be an effective warm-up strategy for future tasks.

Intuitively, the auxiliary objective we present in the following encourages each individual future classification head to yield similar responses for ``similar'' examples. However, as we dispose of label information for both the examples from the current task and the memory buffer, we refine the contrastive objective by incorporating class supervision. As shown in~\cite{khosla2020supervised}, we can leverage it to pull together representations of examples from the same class and to do the opposite for different classes.

In practice, given a batch of $N$ examples, we extend it by appending $N$ variants of the original items (obtained through strong data augmentation). We then consider the output response $f(x_i; \theta) \triangleq \ell(x_{i})$ for the $i^{\text{th}}$ example: in particular, we firstly focus on the (normalized) $j^\text{th}$ future head (s.t.\ $j \in \{c+1,\dots,T\}$), which we denote with $\tilde\ell_{\operatorname{fu}[c;j]}(x_{i}) \triangleq \operatorname{L2Norm}(\ell_{\{j\cdot|\mathcal{Y}|,\dots,(j+1)\cdot|\mathcal{Y}|-1\}}(x_i))$. On top of this, we compute the following loss term:
\begin{equation}
\label{eq:selfsupith}
    \mathcal{L}_{\text{SC}}(x_i, y_i; j) = -\hspace{-0.5em}\sum_{p\in P(i)}{log \frac{\operatorname{exp}({\tilde\ell_{\operatorname{fu}[c;j]}(x_i) \bigcdot \tilde\ell_{\operatorname{fu}[c;j]}(x_p)}/{\tau})}{\sum\limits_{\substack{k = 1\\k \neq i}}^{2N}{\operatorname{exp}({\tilde\ell_{\operatorname{fu}[c;j]}(x_i) \bigcdot \tilde\ell_{\operatorname{fu}[c;j]}(x_k)}/{\tau})}} },
\end{equation}
where $P(i) = \{p \in \{1,\dots,2N\}\ |\ i \neq p \land y_i = y_p\}$ stands for the positive set (\ie the indices of examples sharing the label of the $i^{th}$ item) and $\tau$ is a positive scalar value that acts as a temperature. The full objective simply consists in averaging the values of Eq.~\ref{eq:selfsupith} across all future heads:
\begin{equation}
\label{eq:selfsup}
\mathcal{L}_{\text{FP}}(x_i, y_i) = \frac{1}{T-(c+1)} \sum_{j=c+1}^{T} \frac{1}{|P(i)|} \mathcal{L}_{\text{SC}}(x_i, y_i; j).
\end{equation}
Since Eq.~\ref{eq:selfsup} encourages unused heads to convey additional semantics about the examples seen so far, we profitably exploit also future logits during replay. Moreover, as new classes emerge in later tasks, we account for the corresponding semantic drift by applying the update procedure outlined in Sec.\ref{sec:xder} also on these logits.
We empirically found it beneficial to apply logits-replay also for future heads; therefore, the outlined update procedure also extends to the future logits retained in the memory buffer.
\subsection{Bias Mitigation}
\label{sec:biasmitigation}
Sec.~\ref{sec:limitations} reports that one of the main weaknesses of our previous proposal regards the bias it accumulates on classes of the current task. This unfolds in two directions: on the one hand, most errors regarding past tasks are due to a blind preference of the model towards novel classes; on the other hand, future heads are only provided negative samples and therefore collapse to bad configurations that may hurt or slow down later learning. 

\vspace{0.5em}
\noindent\textbf{Preventing penalization of past classes.}~As also observed in other recent works~\cite{ahn2020ssilss,mittal2021essentials,caccia2021reducing}, this issue can be mitigated by revising the way the cross-entropy loss is applied during training. Given an example from the current task, we avoid computing the softmax activation on all logits and instead restrict it on those modeling the scores of the current task classes. This way, the predictions of past classes are removed from the equation and thus not penalized by the outweighing gradients of novel examples. In formal terms, we compute the following objective:
\begin{equation}
\label{eq:sce}
\mathcal{L}_{\operatorname{S-CE}}(x_i, y_i) = \operatorname{CE}(\operatorname{softmax}(\ell_{\operatorname{pr[c]}}(x_i)), y_i) 
\end{equation}
where $c$ indicates the index of the current task. We remark that, for the class-balanced buffer datapoints, such modification is not strictly necessary; hence, we naturally compute the softmax across the logits of all classes seen so far. 

\vspace{0.5em}
\noindent\textbf{Restraining past and future activations.}~In light of the considerations above, we apply the cross-entropy term in isolation on present logits. This favorably prevents the dampening of both past and future heads; however, if left unchecked, nothing prevents their responses from outgrowing present ones and causing trivial classification errors.

To avoid this issue, we provide an optimization constraint to limit the activations of past and future heads: for current task examples, we require their corresponding responses to be lower than a safeguard threshold, identified as the logit $\ell_{\operatorname{gt}}(x_{i}) \triangleq \operatorname{one-hot}(y_{i}) \bigcdot \ell(x_{i})$ corresponding to the ground-truth class. In doing so, we penalize the maximum past (future) logit $\ell_{\operatorname{pa-max}}(x_{i}) = \operatornamewithlimits{max}_{j\in\operatorname{pa}[c]}\ell_{j}(x_i)$ ($\ell_{\operatorname{fu-max}}(x_{i}) = \operatornamewithlimits{max}_{j\in\operatorname{fu}[c]}\ell_{j}(x_i)$) if it oversteps $\ell_{\operatorname{gt}}(x_{i})$:
\begin{align}
\label{eq:fupaconstr}
\mathcal{L}_{\text{PFC}}(x_{i}, y_{i}) = &\max(0,\ell_{\operatorname{pa-max}}(x_{i}) - \ell_{\operatorname{gt}}(x_{i}) + m) +\nonumber\\ &\max(0,\ell_{\operatorname{fu-max}}(x_{i}) - \ell_{\operatorname{gt}}(x_{i}) + m),
\end{align}
where $m$ is a hyper-parameter (we typically set it to $0.3$ in our experiments) that controls the strictness of the penalty.
\subsection{Overall Objective}
\begin{figure}
    \centering
    \includegraphics[width=0.95\linewidth]{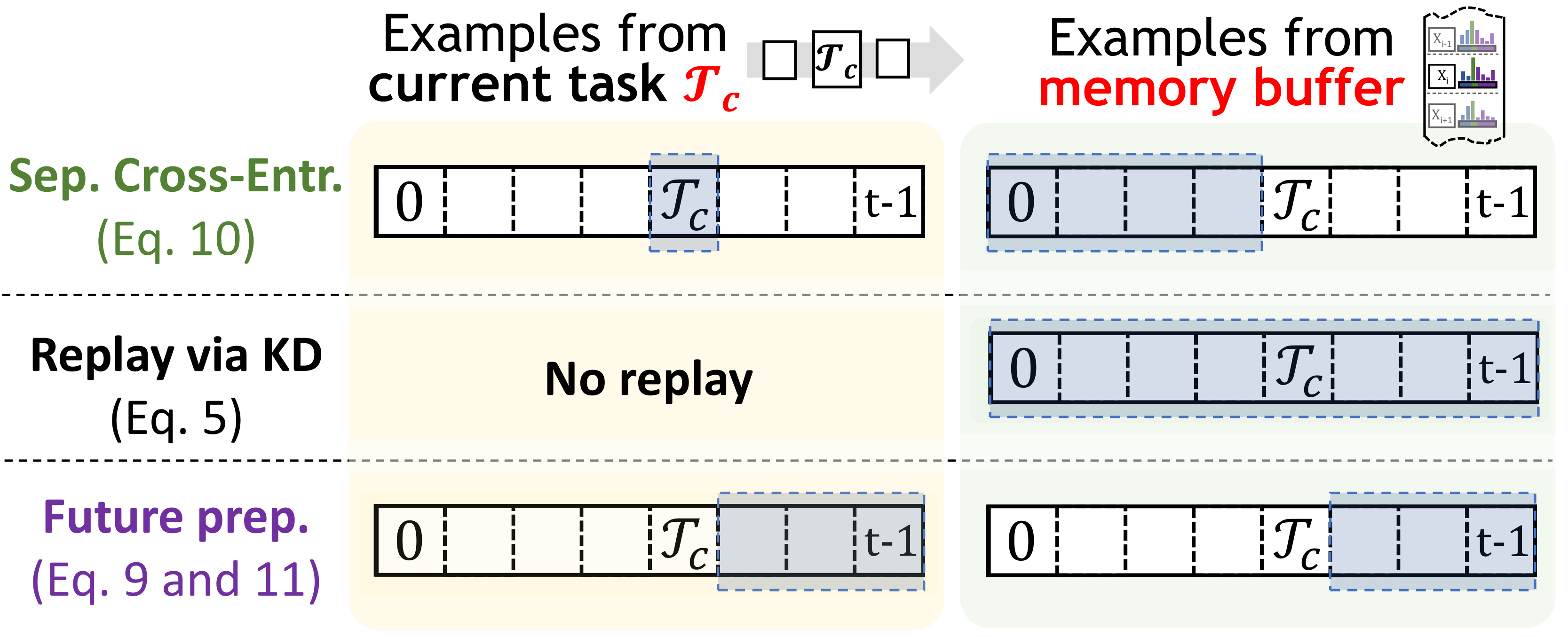}
    \caption{\changed{Visual summary of which loss terms cover each classification heads.}}
    % \vspace{-0.5em}
    \label{fig:xder_losses}
\end{figure}
To sum up, \XDER seeks to optimize the following minimization problem:
\begin{equation}\label{eq:loss_total}
    \argmin_{\theta}\quad\mathcal{L}_{\operatorname{X-DER}} \triangleq \mathcal{L}_{\operatorname{DER}} + \mathcal{L}_{\operatorname{S-CE}} + \mathcal{L}_{\operatorname{F}}, 
\end{equation}
where $\mathcal{L}_{\operatorname{DER}}$ denotes the objective reported in Eq.~\ref{eq:der}, $\mathcal{L}_{\operatorname{S-CE}}$ rewrites Eq.~\ref{eq:sce} to take into account examples from both the current task $c$ and the memory buffer 
\begin{equation}
\mathcal{L}_{\operatorname{S-CE}} = \mathop{\mathds{E}}_{\substack{(x,y) \sim \mathcal{T}_c \\ (x',y') \sim \mathcal{M}}} \big[\mathcal{L}_{\operatorname{S-CE}}(x,y) + \beta \cdot \mathcal{L}_{\operatorname{S-CE}}(x',y')\big],
\end{equation}
and $\mathcal{L}_{\operatorname{F}}$ groups together the goals outlined in Sec.~\ref{sec:futurepreparation} and~\ref{sec:biasmitigation}
\begin{equation*}
\mathcal{L}_{\operatorname{F}} = \mathop{\mathds{E}}_{\substack{(x,y) \sim \mathcal{T}_c \\ (x',y') \sim \mathcal{M}}} \big[
\lambda\underbrace{\mathcal{L}_{\operatorname{FP}}(x||x',y|| y')}_{\substack{ \text{Eq.~\ref{eq:selfsup}} \\ \text{Future Preparation}}}+
\eta\underbrace{\mathcal{L}_{\operatorname{PFC}}(x||x',y||y')}_{\substack{\text{Eq.~\ref{eq:fupaconstr}} \\\text{Past/Future Constraint}}}\big].
\end{equation*}
It is noted that $\beta$, $\lambda$, and $\eta$ are three hyperparameters weighing each contribution to the overall loss. For the sake of clarity, \changed{Fig.~\ref{fig:xder_losses}} proposes a visual breakdown of the loss terms and the involved partitions of the classifier. \changed{For a deeper technical understanding of \XDER, we refer the reader to the pseudo-code provided in App.~\ref{app:pseudocode}.}

\section{Experimental Analysis}
\subsection{Experimental Settings}
\noindent\textbf{Datasets}.~To assess the merits of our proposal, we firstly focus our experiments on well-known and challenging image classification datasets, whose classes undergo a split to form the sequences of tasks.
\begin{itemize}
    \item \textbf{Split CIFAR-100}~\cite{zenke2017continual,rebuffi2017icarl,chaudhry2019tiny} is obtained by splitting the CIFAR-100 dataset~\cite{krizhevsky2009learning} into $10$ consecutive tasks; each comprises of $10$ classes with $500$ and $100$ $32\times32$ images each for training and testing respectively; 
    \item \textbf{Split\,\,\,\miniimagenet}~\cite{chaudhry2019tiny,ebrahimi2020adversarial,derakhshani2021kernel} derives from \miniimagenet~\cite{vinyals2016matching}, a $100$-class subset of the popular ImageNet dataset, split in $20$ classification tasks. Each task presents $84\times 84$ RGB images out of $5$ disjoint classes: by so doing, we can assess our findings on a more complex problem, in terms of both the number of tasks and input resolution. 
\end{itemize}
 
In addition, we set images aside and conduct experiments on Action Recognition: to this aim, we introduce \textbf{Split NTU-60}, a sequential classification benchmark built on top of the 3D skeletal data from the NTU-RGB+D dataset~\cite{shahroudy2016ntu}. To the best of our knowledge, this is the first Continual Learning experimental setting that targets graph-based action classification. We consider this an interesting complement to traditional settings for a threefold reason: i) it deals with a data type (\ie skeletal graphs that expand in time) that radically differs from images; ii) it sheds light on the tendency of different deep architectures to incur forgetting -- hence, not only the common MLPs and CNNs, but also Graph CNNs (GCNNs)~\cite{kipf2017semi,porrello2019classifying}; iii) as it still tackles classification, we can provide novel forgetting measurements that characterize existing and well-established approaches. 
In our experiments, we split the data of NTU-RGB+D into 6 tasks, each of which contains 10 classes. We refer the reader to App.~\ref{app:ntu} for further details regarding this dataset.

\vspace{0.5em}
\noindent\textbf{Architectures.}~For Split CIFAR-100, we use ResNet18~\cite{he2016deep} as in~\cite{rebuffi2017icarl}. For Split \miniimagenet we opt instead for EfficientNet-B2~\cite{tan2019efficientnet}, which has recently arisen as a more suitable baseline network that allows better performance with fewer parameters and faster inference: we argue that the resulting considerations are therefore more indicative of the current advances in deep learning. For the same reason, we opt for EfficientGCN-B0~\cite{song2021constructing} when dealing with Split NTU-60. Further details can be found in App.~\ref{app:backbone}.

\vspace{0.5em}
\noindent\textbf{Training details.}~All models are trained from scratch (no pre-training has been used). While there are some works~\cite{lopez2017gradient,aljundi2019gradient,aljundi2019online} that have recently investigated the single-epoch scenario (no more than one pass on training data), we place our experiments in the multi-epoch setup~\cite{rebuffi2017icarl,wu2019large,zenke2017continual}. As argued by our previous work~\cite{buzzega2020dark}, this leads to easier-to-read results, in which the under-fitting linked to few iterations is removed from the equation. In more detail, we always use Stochastic Gradient Descent as optimizer and a number of epochs that varies according to the dataset (50 for CIFAR-100, 80 for \miniimagenet and 70 for NTU); in line with~\cite{rebuffi2017icarl,wu2019large,hou2019learning}, we also define a set of epochs at which the learning rate is divided by $10$ ($[35,45]$ for CIFAR-100, $[35,60,75]$ for \miniimagenet). For NTU, we use a cosine scheduler with a 10-epoch warm-up.

For all the evaluated approaches, we select their hyperparameters through grid-search. \changed{We refer the reader to the Appendix for: \textit{i)} a full description of the considered parameter combinations, the chosen ones, and other training details (App.~\ref{app:hyper}); \textit{ii)} an empirical evaluation of the sensitivity of \XDER to different choices of hyperparameters (App.~\ref{app:sensitivity}).}

It is worth noting that the experimental settings of different works are often subtly but meaningfully dissimilar, which makes drawing direct comparisons among them difficult. Therefore, we avoid taking results directly from other works and instead re-run all experiments in a common and unified experimental environment (for which we make the code-base available at \url{https://github.com/aimagelab/mammoth}). 

\vspace{0.5em}
\noindent\textbf{Metrics.}~We firstly assess the performance in terms of Final Average Accuracy (FAA). Let $a_i^t$ be the model accuracy on the $\text{i}^{\text{th}}$ task after training on task $\mathcal{T}_t$, we define FAA as:
\begin{equation}
    \operatorname{FAA} \triangleq \frac{1}{T} \sum_{i=0}^{T-1} a_i^{T-1},
\end{equation}
where $T$ denotes the total number of tasks. FAA represents the most immediate summarizing measure that allows direct comparisons. However, it provides only a snapshot of the state after the last task: to account for what happens during the entire sequence, we follow other works~\cite{rebuffi2017icarl,hou2019learning,wu2019large} and exploit the Final Forgetting (FF)~\cite{chaudhry2018riemannian} metric:
\begin{equation}
\begin{gathered}
    \operatorname{FF} \triangleq \frac{1}{T-1} \sum_{j=0}^{T-2} f_j,~~\text{s.t.}~~f_j=\operatornamewithlimits{max}_{l\in\{0,\dots,T-2\}} a_j^l - a_j^{T-1}.
\end{gathered}
\end{equation}
FF is bound in $[-100,100]$ and measures the average accuracy degradation, \ie the maximum discrepancy in performance observed for a given task through training.
\changed{To complement our analysis of performance, we also examine \XDER both in terms of its sensitivity to hyperparameters, memory footprint and training time. A comparative evaluation w.r.t.\ SOTA methods can be found in App.~\ref{app:memtim}}.
\begin{table*}[t]
    \setlength{\tabcolsep}{1.3em}
    \renewcommand{\arraystretch}{1.3}
    \caption{Class-Incremental Continual Learning Final Average Accuracy (FAA) and Final Forgetting (FF) (in parentheses).}
    \label{table:faa}
    \centering
    \rowcolors{1}{}{lightgray}
    \begin{tabular}{lccccc}
    \hline
    \textbf{FAA} [$\boldsymbol{\uparrow}$\hspace{-.7em}] \textbf{(FF} [$\boldsymbol{\downarrow}$]\textbf{)} & \multicolumn{2}{c}{\textbf{CIFAR-100}} & \multicolumn{2}{c}{\textbf{\miniimagenet}} & \multicolumn{1}{c}{\textbf{NTU-60}} \\
    \hline
    \textbf{JT} (upper bound) & \multicolumn{2}{c}{\normresvtwo{70.44}{-}} & \multicolumn{2}{c}{\normresvtwo{53.55}{-}} & \multicolumn{1}{c}{\normresvtwo{85.75}{-}}\\
    \hline
    \textbf{FT}  (lower bound)              & \multicolumn{2}{c}{\normresvtwo{9.43}{89.82}}  &  \multicolumn{2}{c}{\normresvtwo{4.51}{77.38}}  & \normresvtwo{15.74}{92.85} \\
    \textbf{LwF.MC}~\cite{rebuffi2017icarl} & \multicolumn{2}{c}{\normresvtwo{16.22}{54.89}}  &  \multicolumn{2}{c}{\normresvtwo{12.20}{23.96}}  & \normresvtwo{28.24}{46.50} \\
    \hline
    \rowcolor{white}{}
    \textbf{$\mathcal{M}_\text{size}$} & 500 & 2000 & 2000 & 5000 & 500 \\
    \hline
    \textbf{ER}~\cite{riemer2018learning}     & \normresvtwo{22.10}{73.64} & \normresvtwo{38.58}{53.28} & \normresvtwo{14.57}{64.49} & \normresvtwo{21.42}{50.36} & \normresvtwo{51.77}{48.54} \\
    \textbf{GDumb}~\cite{prabhu2020gdumb}     & \normresvtwo{9.98}{-} & \normresvtwo{20.66}{-} & \normresvtwo{15.22}{-}              & \normresvtwo{27.79}{-} & \normresvtwo{27.59}{-} \\
    \textbf{ER-ACE}~\cite{caccia2021reducing} & \normresvtwo{38.75}{40.04} & \normresvtwo{49.72}{25.71} & \normresvtwo{22.60}{23.74} & \normresvtwo{27.92}{19.72} & \normresvtwo{52.14}{23.33} \\
    \textbf{RPC}~\cite{pernici2021class}      & \normresvtwo{22.34}{71.94} & \normresvtwo{38.33}{52.33} & \normresvtwo{15.60}{61.00} & \normresvtwo{24.69}{46.34} & \normresvtwo{49.40}{48.10} \\
    \textbf{BiC}~\cite{wu2019large}           & \normresvtwo{36.02}{51.85} & \normresvtwo{46.39}{40.49} & \normresvtwo{12.96}{57.19} & \normresvtwo{14.45}{56.55} & \normresvtwo{29.20}{66.16} \\
    \textbf{iCaRL}~\cite{rebuffi2017icarl}    & \normresvtwo{46.52}{22.06} & \normresvtwo{49.82}{18.07} & \normresvtwo{22.58}{16.46} & \normresvtwo{22.78}{16.37} & \normresvtwo{45.83}{21.48} \\
    \textbf{LUCIR}~\cite{hou2019learning}     & \normresvtwo{40.59}{34.55} & \normresvtwo{41.73}{25.41} & \normresvtwo{14.97}{43.83} & \normresvtwo{17.61}{39.01} & \normresvtwo{58.06}{32.58} \\
    \hline
    \textbf{DER}~\cite{buzzega2020dark}       & \normresvtwo{36.60}{54.99} & \normresvtwo{51.89}{34.54} & \normresvtwo{22.96}{48.78} & \normresvtwo{29.83}{36.38} & \normresvtwo{49.49}{43.09} \\
    \textbf{DER++}~\cite{buzzega2020dark}     & \normresvtwo{38.25}{50.54} & \normresvtwo{53.63}{33.66} & \normresvtwo{23.44}{46.69} & \normresvtwo{30.43}{37.11} & \normresvtwo{55.32}{35.95} \\
    \hline
    \textbf{\XDER w/o memory update}          & \normresvtwo{42.67}{24.03} & \normresvtwo{56.55}{9.24} & \normresvtwo{25.76}{16.76} & \normresvtwo{31.40}{13.50}  & \normresvtwo{57.66}{12.52} \\
    \textbf{\XDER w/o future heads} & \normresvtwo{45.61}{33.31} & \normresvtwo{55.00}{22.94} & \normresvtwo{21.71}{36.92}  & \normresvtwo{27.45}{18.39} & \normresvtwo{61.02}{9.80} \\
    \textbf{\XDER w/ CE future heads}      & \normresvtwo{47.67}{25.12} & \normresvtwo{55.61}{10.52}  & \normresvtwo{27.18}{36.12} & \normresvtwo{30.69}{16.80} & \normresvtwo{61.58}{10.94} \\
    \textbf{\XDER w/ RPC future heads}          & \normresvtwo{48.53}{26.94} & \normresvtwo{57.00}{12.65} & \normresvtwo{26.38}{38.33} & \normresvtwo{29.91}{28.29}  & \normresvtwo{62.41}{8.88} \\
    \textbf{\XDER}        & \boldresvtwo{49.93}{19.90} & \boldresvtwo{59.14}{12.58} & \boldresvtwo{28.19}{20.45} & \boldresvtwo{31.70}{15.87} & \boldresvtwo{64.86}{9.95} \\
    \hline
    \end{tabular}
\end{table*}
\subsection{Baselines and Competing Methods}
\label{sec:baselines}
To gain a clear understanding, we provide two methods that act as \textbf{lower-} and \textbf{upper-bound}: the former continually fine-tunes on the most recent task with no remedy to catastrophic forgetting (\textit{a.k.a.} Fine-Tuning, FT); the latter trains a model jointly on all data (\textit{a.k.a.} Joint-Training, JT).

\noindent We use the following CiCL approaches as competitors:
\begin{itemize}
    \item \textbf{Learning without Forgetting (LwF)}~\cite{li2017learning} is a regularization approach that exploits Knowledge Distillation: it uses the model learned in the previous task as a teacher during the current one. As done in~\cite{rebuffi2017icarl}, we use LwF.MC (the CiCL adaptation of LwF);
    \item \textbf{Experience Replay (ER)}~\cite{ratcliff1990connectionist, robins1995catastrophic} is a simple rehearsal method that stores previously encountered examples in a memory buffer for later replay. In spite of its simplicity, it still stands as a strong baseline~\cite{riemer2018learning, chaudhry2019tiny};
    \item \textbf{ER with Asymmetric Cross-Entropy (ER-ACE)}~\cite{caccia2021reducing} is a recently proposed modification of ER that addresses stream imbalance w.r.t.\ to the memory buffer by optimizing separate cross-entropy loss terms;
    \item \textbf{Incremental Classifier and Representation Learning (iCaRL)}~\cite{rebuffi2017icarl} combines a carefully-designed \textit{herding} buffer with a \textit{nearest mean-of-exemplars} classifier; it is often regarded as a strong performer on complex datasets;
    \item \textbf{Bias Correction (BiC)}~\cite{wu2019large} pairs Experience Replay with a regularization term that resembles the objective of LwF. Most notably, it leverages a separate layer that aims at counteracting bias in the backbone network;
    \item \textbf{Learning a Unified Classifier Incrementally via Rebalancing (LUCIR)}~\cite{hou2019learning} is a rehearsal strategy proposing several modifications that promote separation in feature space and result in a more harmonized incrementally-learned classifier;
    \item \textbf{Greedy Sampler and Dumb learner (GDumb)}~\cite{prabhu2020gdumb} is an experimental method meant to question the advances in CL. It totally avoids training steps on data from the current task and just fills up the memory buffer: when an evaluation is required, it then trains a new model on the memory buffer from scratch;
    \item Our previous proposals: Dark Experience Replay (DER) and its variant that uses also labels (DER++).
\end{itemize}

\vspace{0.5em}
\noindent\textbf{Ablative studies} Besides reviewing the performance of \XDER in light of the state of the art, we provide additional comparisons to validate the design choices of \XDER:
\begin{itemize}
    \item \textbf{X-DER w/o memory update}, which does not update logits through the sequence of tasks;
    \item \textbf{X-DER w/o future heads}, which represents the simplest way to handle new classes: namely, it simply adds a new classification head once the new task is presented;
    \item \textbf{X-DER w/ CE on future heads}, a baseline that devises future heads. In line with what is done by DER(++), it includes them in the computation of the stream-specific portion of the separated cross-entropy loss by targeting them to zero probabilities;
    \item \textbf{X-DER w/ RPC}, an alternative to the semi-supervised strategy devised in Sec.~\ref{sec:futurepreparation}. It builds future preparation upon the \textbf{Regular Polytope Classifier} proposed in~\cite{pernici2021class}. Briefly, it appoints the parameters of the final classification layer to constant weights values, the latter being designed to keep them equally distributed in output space. This way, the authors meant to ensure that all classes (both seen and unseen) are all at equal distance.
\end{itemize}

\noindent For completeness, we also evaluate the original \textbf{Regular Polytope Classifier (RPC)}~\cite{pernici2021class}, which complements ER with the above-mentioned fixed final classifier.

\subsection{Discussion}
\label{sec:exp}
\begin{figure*}[t]
    \centering
    \includegraphics[width=1.\linewidth,keepaspectratio]{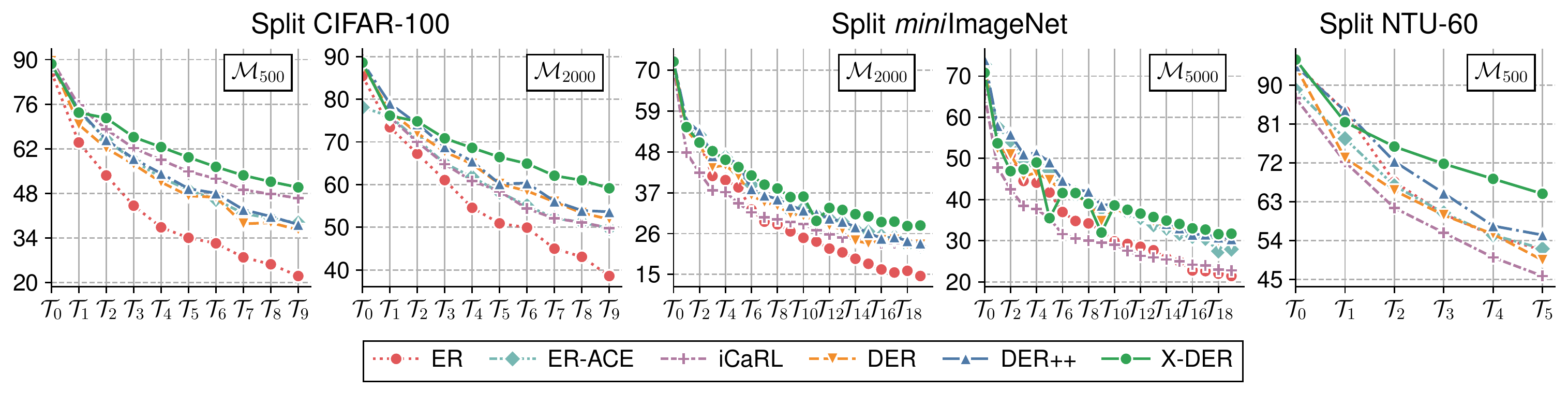}\vspace{-0.35em}
    \caption{For the experimental settings reported in Tab.~\ref{table:faa}, the trend of the average test-set accuracy on the observed tasks.}
    \label{fig:accuracies}\vspace{-0.35em}
\end{figure*}
By examining Tab.~\ref{table:faa}, we can make the preliminary consideration that the considered regularization approach (LwF.MC) consistently underperforms compared to online replay-based methods. This observation aligns with~\cite{aljundi2019gradient, buzzega2020dark} and suggests that adopting a replay memory is crucial for achieving solid performance in CiCL.

As it only learns from its memory buffer, the offline training of GDumb allows it to observe few examples from all classes (old and new) jointly, thus avoiding issues related to bias by design. On Split \miniimagenet, which features a long sequence of tasks, this is sufficient to outperform methods that do not compensate for this effect (\eg ER, RPC). However, since it entirely discards the remaining data from the input stream, GDumb produces a lower FAA w.r.t.\ to most online-learning methods.

Among ER-based approaches, ER-ACE stands out as the most effective thanks to its loss, carefully designed to prevent interference between the learning of the current task and the replay of old data. This trait allows to protect previously acquired knowledge, resulting in lower FF metric.

On average, methods combining rehearsal and distillation achieve better performance w.r.t.\ simple replay. iCaRL limits forgetting consistently and achieves balanced accuracy on all seen tasks thanks to its nearest-mean-of-exemplars classifier. This is rewarding on the medium-length CIFAR-100 benchmark, but proves sub-optimal on both \miniimagenet and NTU-60 (due to forgetting on the former and to lack of fitting of the current task on the latter). Differently, LUCIR delivers a high accuracy on the last few encountered tasks, thus proving very effective on the short Split NTU-60 but struggling on longer sequences. While its performance is adequate on CIFAR-100, BiC is characterized by the highest FF on all other benchmarks, leading to an FAA score close to its parent method LwF.MC.

Our previous proposals DER and DER++ classify as strong baselines when combined with a large-enough memory buffer. However, due to the limitations explored in Sec.~\ref{sec:limitations}, they occasionally give in to approaches that contrast bias more effectively (ER-ACE, iCaRL, LUCIR for $\mathcal{M}_\text{size}$ $500$ on Split CIFAR-100; ER-ACE and LUCIR on Split NTU-60).

Compared to the current state-of-the-art, \XDER delivers higher accuracy and lower forgetting across all benchmarks. As one can observe from a close exam of its incremental accuracy values (Fig.~\ref{fig:accuracies}), the proposed enhancements lead to increased performance retention on past tasks, lifting its score significantly over competitors as training progresses.

To gain a deeper understanding, we further compare \XDER against its ablative baselines. By omitting to update the content of the memory buffer (\XDER w/o memory update), we see a significant drop in performance -- especially relevant for smaller $\mathcal{M}_\text{size}$.
Comparatively, the strategy adopted for preparing future logits is less influential. The proposed contrastive preparation loss of \XDER yields the lowest FF rates, validating our intuition to use past data to prepare future learning. Adopting the theoretically grounded but fixed design of \XDER w/\ RPC comes at a steady but non-negligible cost in performance across all benchmarks. X-DER w/\ CE future heads shows that dampening future heads by indiscriminately applying CE leads to a further decrease in accuracy; however, even this approach is still preferable to training \XDER without future heads, which is linked to higher FF metrics. This stresses the importance of preparing the model for future classes and suggests that using future heads and replaying their logits can act as a remedy against forgetting.

\section{Model Analysis}
\subsection{Towards Better ``Continual'' Teachers}
This section delves into the regularization strategy of \XDER: why are its responses so effective against forgetting of old tasks? We here build upon the seminal work of~\cite{menon2021statistical}, which has recently proposed a statistical background of Knowledge Distillation (KD) helping researchers and practitioners to gain novel insights on its effectiveness. Essentially, the authors assume that the teacher's response $\mathds{P}^{t}(y|x)$ constitutes an approximation of the true \textit{Bayes class-probability distribution} $\mathds{P}*(y|x)$, which represents the suitability of each class $y$ for a given $x$ (hence encoding confusions amongst the labels). With respect to one-hot targets, it is proven that minimizing the risk associated with $\mathds{P}*$ gives the student an objective with lower \textit{variance}, which aids generalization. Nevertheless, the true $\mathds{P}*$ cannot be accessed and an imperfect estimate has to be used (\eg the response of a teacher net). In that sense, \textit{the better} the estimation of the true Bayes probabilities, \textit{the higher} the generalization capabilities of a student learning through the corresponding risk.

In the following, we show how such a novel perspective can help to gain a new understanding of our approach.
\subsubsection*{Analysis of Secondary Information}
\begin{table}[t]
    \caption{Secondary information metrics (lower is better). \xmark~indicates no use of Knowledge Distillation (KD) while training, \cmark*~indicates KD of past logits only, \cmark~indicates KD of all logits (incl.\ future past).}
    \centering
    \rowcolors{2}{lightgray}{}
    \setlength{\tabcolsep}{4pt}
    \begin{tabular}{l|c|cccc}
    \toprule
    & \textbf{KD} & \multicolumn{2}{c}{\textbf{SS-ERR} $\boldsymbol{[\downarrow]}$} &  \multicolumn{2}{c}{\textbf{SS-NLL} $\boldsymbol{[\downarrow]}$}\\
    \midrule
    \multicolumn{2}{l}{\textbf{$\mathcal{M}_\text{size}$}} & 500 & 2000 & 500 & 2000 \\
    \midrule
    \textbf{ER}             & \xmark  & \simnormres{0.71} & \simnormres{0.68} & \simnormres{4.00} &  \simnormres{4.32} \\
    \textbf{ER-ACE}         & \xmark  & \simnormres{0.63} & \simnormres{0.60} & \simnormres{2.64} &  \simnormres{2.81} \\
    \midrule
    \textbf{DER}            & \cmark* & \simnormres{0.67} & \simnormres{0.64} & \simnormres{2.22} &  \simnormres{2.21} \\
    \textbf{DER++}          & \cmark* & \simnormres{0.67} & \simnormres{0.64} & \simnormres{2.22} &  \simnormres{2.25} \\
    \midrule
    \textbf{LUCIR}          & \cmark  & \simsecores{0.60} & \simsecores{0.59} & \simnormres{2.21} &  \simnormres{2.22} \\
    \textbf{iCaRL}          & \cmark  & \simsecores{0.60} & \simnormres{0.60} & \simsecores{1.90} &  \simsecores{1.94} \\
    \midrule
    \textbf{\XDER ~{\small w/o memory update}} & \cmark*  & \simnormres{0.64} & \simnormres{0.61} & \simnormres{2.14} &  \simnormres{2.10} \\
    \textbf{\XDER} & \cmark  & \simboldres{0.57} & \simboldres{0.56} & \simboldres{1.83} &  \simboldres{1.82} \\
    \bottomrule
    \end{tabular}
    \label{tab:brox_values}
    \vspace{-0.2cm}
\end{table}
In literature, the concept of Bayes class-probabilities has also been studied in terms of \textbf{secondary information}~\cite{yang2019training,mittal2021essentials} \ie for each non-maximum score, the model’s belief about the semantic cues of the corresponding class within the input image. Unsurprisingly, \textit{Yang et al.}\ in~\cite{yang2019training} identify the preservation of secondary information as a key property of KD: they empirically find that teachers with richer secondary information lead to students that generalize better. However -- when dealing with catastrophic forgetting -- it can be challenging to capture rich secondary information, as the latter becomes available only as tasks progress.

Seeking to measure how effectively distinct CL approaches can leverage secondary information, we follow the setup proposed in~\cite{mittal2021essentials} and re-examine their performance on the test-set of CIFAR-100 but labeled in a different way: we indeed group its $100$ classes into their natural $20$ super-classes. According to the authors of~\cite{mittal2021essentials}, a model achieving a high classification score in this setup also retains better secondary information, as classes belonging to the same super-class can be assumed to have higher visual similarity than the ones belonging to different super-classes.

The retained secondary information can be quantified by two metrics~\cite{mittal2021essentials}: on the one hand, the \textbf{Secondary-Superclass Error} ($\operatorname{SS-ERR}$) equals $1$ minus the probability of predicting the right super-class. As the focus lies on secondary information, the maximum logit is always omitted during softmax computation. Otherwise, the \textbf{Secondary-Superclass NLL} ($\operatorname{SS-NLL}$) considers the negative log-likelihood when using super-classes as labels. 

From the results in Tab.~\ref{tab:brox_values}, \XDER, iCaRL and LUCIR consistently end up predicting the correct coarse classes (lower SS-ERR) and do so more confidently (lower SS-NLL). This is in line with our expectations: as these methods handle hindsight-learned similarities between newly discovered classes and old ones, the corresponding teaching signal leads the student toward richer secondary information. In contrast, DER(++) yield lower metrics due to: \textit{i)} their distillation targets neglecting logits of future past, \textit{ii)} the existence of a large bias towards the last seen classes. To verify the importance of \textit{i)}, we also run this evaluation on a version of \XDER that does not update its buffer logits; this results in metrics comparable to DER(++). ER -- which applies no distillation at all -- also experiences the issues \textit{i)} and \textit{ii)}; it indeed produces the highest metrics among the evaluated methods. On the contrary, ER-ACE -- which addresses \textit{ii)} by means of its segregated objective -- attains lower metrics closing in on DER(++). This highlights that bias control too plays a primary role in the emergence and conservation of secondary information.
\subsubsection*{Offline Training on Memory Buffer}
\begin{figure}
    \centering
    \includegraphics[width=.9\linewidth,keepaspectratio]{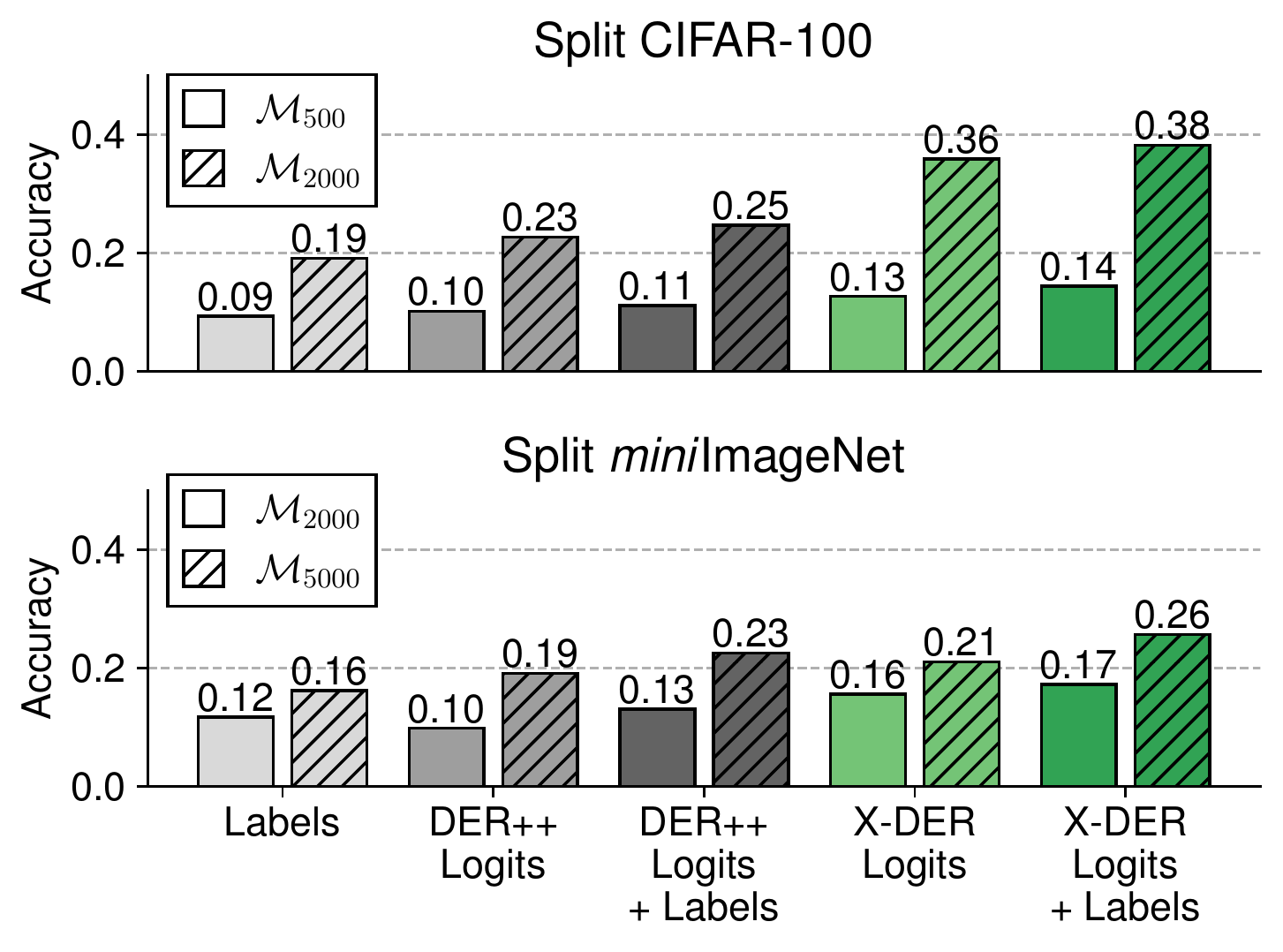}
    \caption{Accuracy of models trained from scratch on memory buffers of ER (\textit{Labels}), DER++ (\textit{Logits}, \textit{Both}), and \XDER (\textit{Logits}, \textit{Both}). The resulting accuracy represents a proxy of the informativeness of the memory buffer.}
    \label{fig:informativeness}
\end{figure}
Distinct rehearsal methods compared in Sec.~\ref{sec:exp} retain different summaries of the previously encountered knowledge: approaches such as ER, iCaRL, LUCIR, etc. employ labels for the recorded samples, DER(++) use the responses provided by the model at insertion time, while \XDER exploits responses that are updated as future past logits become reliable. As done in our previous work~\cite{buzzega2020dark} (which used the simpler Split CIFAR-10 dataset), we here aim to assess the amount of reliable information retained by these approaches. We then train a model from scratch only on the data available in final buffers constructed by ER, DER++, and \XDER. We compute the performance achieved by the resulting models after $70$ epochs of training and show the results on Split CIFAR-100 and Split \miniimagenet in Fig.~\ref{fig:informativeness}.

In line with the theoretical results of~\cite{menon2021statistical}, we observe that relying on logits yields lower generalization error w.r.t.\ learning from labels solely. In addition, the combination of hard and soft supervision signals leads to slight improvements both for DER++ and \XDER. Secondly, the use of updated logits of \XDER results in a steady improvement: when compared to DER++, we observe an average gain of $6\%$ (when using logits alone) and $6.25\%$ (combined with labels). Based on the considerations above, we attribute such an additional regularization effect to the exploitation of future past logits, which arguably drives the model towards a better estimate of the true \textit{Bayes} class-probabilities.
\subsubsection*{Calibration of Continual Learners: A New Perspective}
\begin{figure}[t]
    \centering
    \begin{tabular}{c}
    \includegraphics[width=.95\linewidth,keepaspectratio]{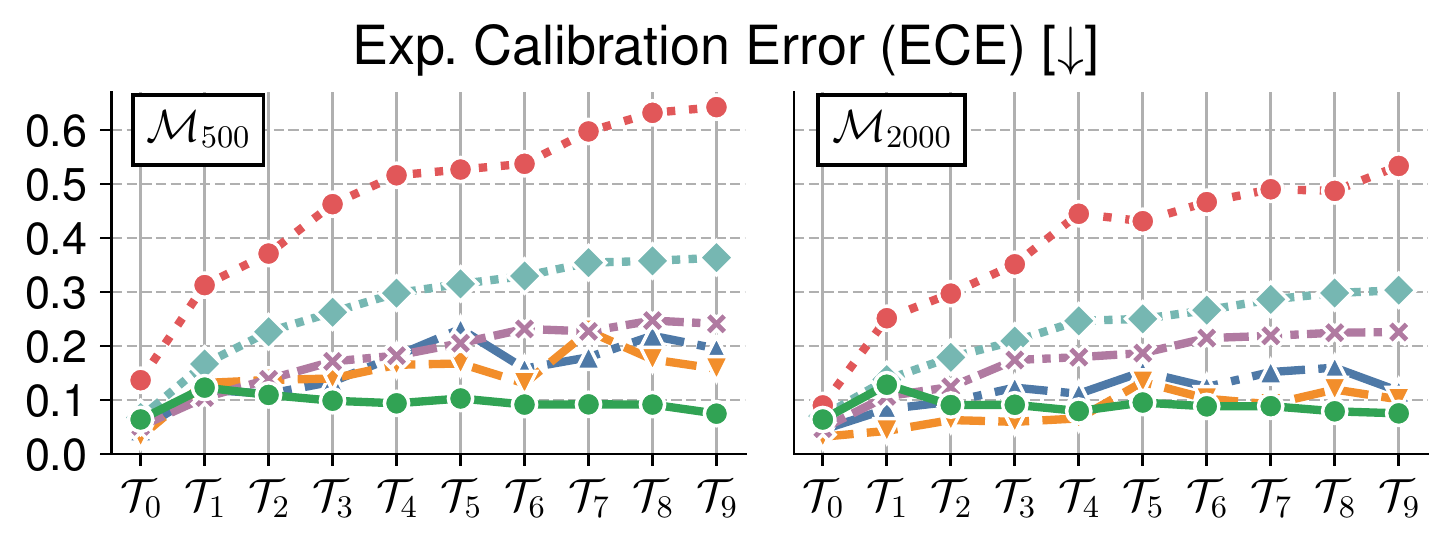} \\
    \includegraphics[width=0.9\linewidth,keepaspectratio,clip,trim=0 4.2cm 0 4.2cm]{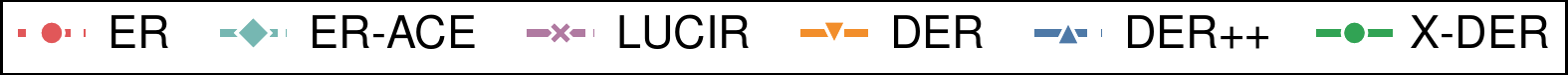}
    \end{tabular}
    \caption{
    Effect of several regularization methods on net calibration (Split CIFAR-100). While most of them degrade with lower memory size (left), \XDER yields robust performance.
    }
    \label{fig:calibration}
\end{figure}
Although~\cite{menon2021statistical} presents an appealing framework, it is still up for debate how to assess the quality of the approximation of the $\mathds{P}*$. \textit{Menon et al.} suggest that a coarse evaluation can be carried out by looking at the Expected Calibration Error (ECE)~\cite{guo2017calibration}. Remarkably, this provides a new light and foundation to the experiment conducted in our previous work~\cite{buzzega2020dark}: in fact, we already compared several replay strategies in terms of the induced ECE, which led us to ascribe the gains of DER(++) also to the higher calibration of the underlying network. 

On these premises, we here repeat the above-mentioned evaluation on top of our new proposal, \XDER. Fig.~\ref{fig:calibration} shows the results obtained on Split CIFAR-100: as can be seen, \XDER delivers a lower ECE compared to other approaches. If this finding could seem trivial when using as baseline \textit{one-hot teachers} such as ER, this holds as well for smoothed ones such as DER(++): in light of the previous considerations, we can now link the advantages of \XDER to a better estimation of the underlying Bayes class-probabilities.
\subsection{Effect of Future Preparation on Unseen Classes}
\begin{figure*}[t]
    \begin{center}
    \includegraphics[width=1.0\linewidth]{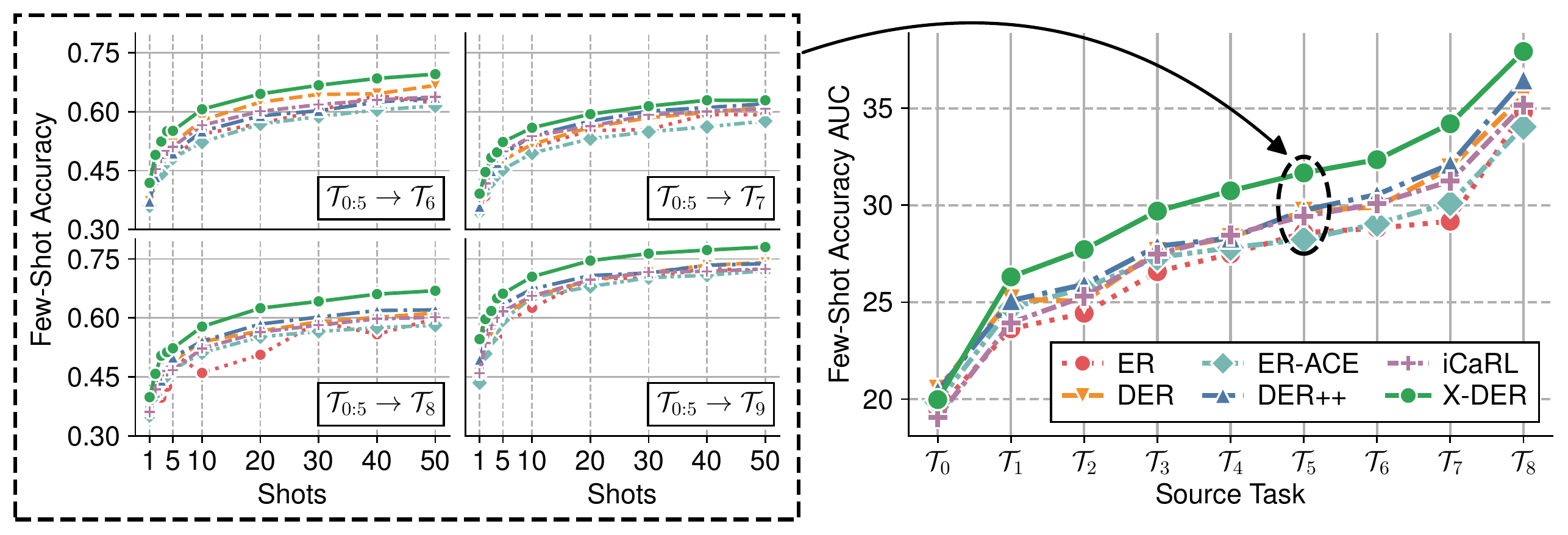}
    \end{center}
    \begin{subfigure}{.5\textwidth}
      \caption{}
      \label{fig:fewshot_left}
    \end{subfigure}
    \begin{subfigure}{.5\textwidth}
      \caption{}
      \label{fig:fewshot_right}
    \end{subfigure}\vspace{-0.35em}
    \caption{Analysis of generalization to unseen classes. (a) For each of the four remaining tasks of CIFAR-100, the performance \textit{vs} training-set size trend for different CL methods; (b) The curves describing the forward transfer at the end of every task.}
    \label{fig:fewshot}\vspace{-0.35em}
\end{figure*}
One of our main contributions regards the design of a pretext task to warm up unused heads; this way, we expect a \textit{gentler} adaption of the network to unseen data distributions, thus removing the need for dramatic updates to its parameters and therefore lowering the risk of forgetting. To verify whether this happens, we envision a setting where very few data of the incoming tasks are available: here, how well does the feature space spanned by future heads work? Does the technique introduced in Sec.~\ref{sec:futurepreparation} give the network any advantage when dealing with unseen data?

We conduct an in-depth investigation of these facets and let Fig.~\ref{fig:fewshot} provide a graphical summary of it. For multiple methods (such as ER, DER(++), and our \XDER), we firstly focus on a single snapshot and stop their training procedure right after the $6^{\text{th}}$ task of CIFAR-100 (see Fig.~\ref{fig:fewshot_left}). We then aim at measuring their performance on each of the remaining four tasks separately: we do that by fitting a  Nearest Neighbor (NN) classifier on top of the activations given by the corresponding future head. To draw a clear picture of the forward transfer delivered by those methods: \textit{i)} we repeat this evaluation at varying training set sizes (ranging from one to fifty shots per class); \textit{ii)} we freeze network parameters; so no fine-tuning steps are performed on data-points of the incoming task. As can be seen, \XDER is the method delivering the best results among other approaches.

Afterwards, we propose a more comprehensive evaluation, which takes into consideration the transfer from all encountered task (and not the $6^{\text{th}}$ task solely): here, we aim to assess how the capabilities linked to forward transfer evolve one task after the other. To do so -- for each observed task $t$ ranging from the first to the penultimate one -- we firstly define the performance curves $\operatorname{NN}_{t \rightarrow \tilde{t}}(k)$ over the unseen tasks $\tilde{t} > t$, where $k$ indicates the number of shots per class; to gain a clear understanding, Fig.~\ref{fig:fewshot_left} focuses on $t=6$ and depicts $\operatorname{NN}_{6 \rightarrow \tilde{t}}(k) \ \forall\tilde{t} \in \{7,8,9,10\}$. Subsequently, we summarize each curve with the \textbf{Area Under the Curve} $\operatorname{AUC}_{t \rightarrow \tilde{t}}$ and finally average the latter across $\tilde{t}$ (\eg $\operatorname{AUC}_{6} \triangleq \frac{1}{4}\sum_{\tilde{t}=7}^{10}\operatorname{AUC}_{6 \rightarrow \tilde{t}}$), thus providing an overall measure of generalization to all tasks that will be encountered onwards. 

In this respect, Fig.~\ref{fig:fewshot_right} reports the trend of the $\operatorname{AUC}_{t}$ for different rehearsal methods. As can be appreciated, we do not observe a clear distinction in their performance when focusing on the earliest tasks. Instead, the AUC curve of \XDER widens the gap as the number of seen tasks increases (it scales better to the number of seen tasks). This suggests that the more and diverse the data modalities present in the memory buffer, the higher the chances that optimizing Eq.~\ref{eq:selfsup} will lead to a good forward transfer to unseen data.
\subsubsection*{Pre-allocation of Future Tasks}
\label{sec:preallocation}
We have so far supposed (see Sec.~\ref{sec:xder}) that the overall number of tasks $T$ can be known in advance. This allows us to instantiate a last fully-connected layer large enough to accommodate the logits for all seen and unseen classes. However, in practical and real scenarios, we may not know how many tasks will be encountered from the outset: hence, it can be brought into question whether our approach can still be applied to those settings. 

We here discuss a straightforward modification that enables the number of future tasks to be unknown. We initially set up the last layer to expose $\tilde{t} + 1$ classification heads: precisely, the one dedicated to the first task and the remaining $\tilde{t}$ ones to future tasks. In addition, we instantiate a new head at the end of each task, thus guaranteeing that one head (at least) is always available to the incoming task.

\begin{figure}[t!]
    \centering
    \includegraphics[width=\linewidth,keepaspectratio]{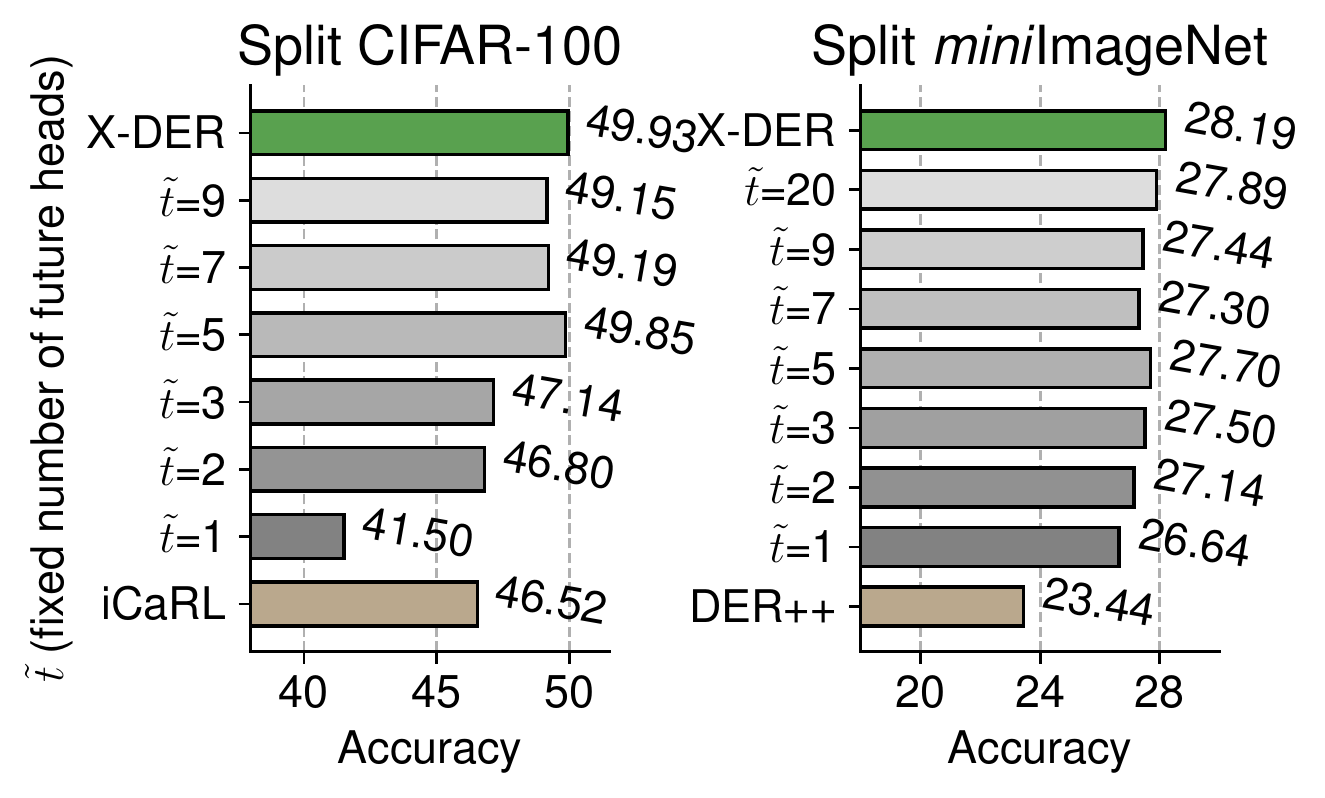}
    \caption{On Split CIFAR-100 and Split \miniimagenet, analysis evaluating how the number of heads pre-allocated in preparation of future tasks affects the final average accuracy.}
    \label{fig:dynhead}
\end{figure}

Fig.~\ref{fig:dynhead} depicts how such a modification affects performance and the impact of the hyperparameter $\tilde{t}$ controlling the number of pre-allocated heads. We draw the following conclusions: \textit{i)} given the slight gap in performance between \XDER and the proposed variant, the overall number of tasks does not seem an essential information for achieving good results; \textit{ii)} a higher number of pre-allocated heads positively influences the final average accuracy. This latter finding suggests that future logits also play a role against forgetting: we conjecture that rehearsal of \textit{non-coding} logits might represent an additional guard against forgetting, as they still embody a reflection of past neural activities.
\subsection{On the Geometry of the Local Minimum}
\subsubsection*{The effectiveness of flat minima in Continual Learning}
We investigated in~\cite{buzzega2020dark} the relation between the nature of the attained local minima and the generalization capabilities linked to them. We indeed conjectured that flatness around a loss minimizer represents a remarkable property for CL settings: intuitively, a loss region tolerant towards local displacements favors later optimization trajectories that entail a less severe drop in performance for old tasks. As a proof of concept, we used two common metrics (recalled later in this section) to characterize the geometry of the minima: DER(++) -- the methods that performed best on the benchmarks of~\cite{buzzega2020dark} -- also exhibit favorably flatter minima.

However, we consider such a matter still nebulous and worthy of further discussion. In this respect, the authors of~\cite{mirzadeh2020understanding} assessed the impact of different training regimes on forgetting and stated that the latter strongly correlates with the curvature of the loss function around the minimum of each task. In practice, they made use of some strategies known to affect the width of the minima (\eg higher initial learning rates, dropout, small batch sizes, etc.) and observed that these lead to a \textit{stable regime} that further mitigates forgetting. Similar arguments have been raised as well in~\cite{cha2020cpr}.

\begin{figure}[t]
    \centering
    \includegraphics[width=.9\linewidth,keepaspectratio]{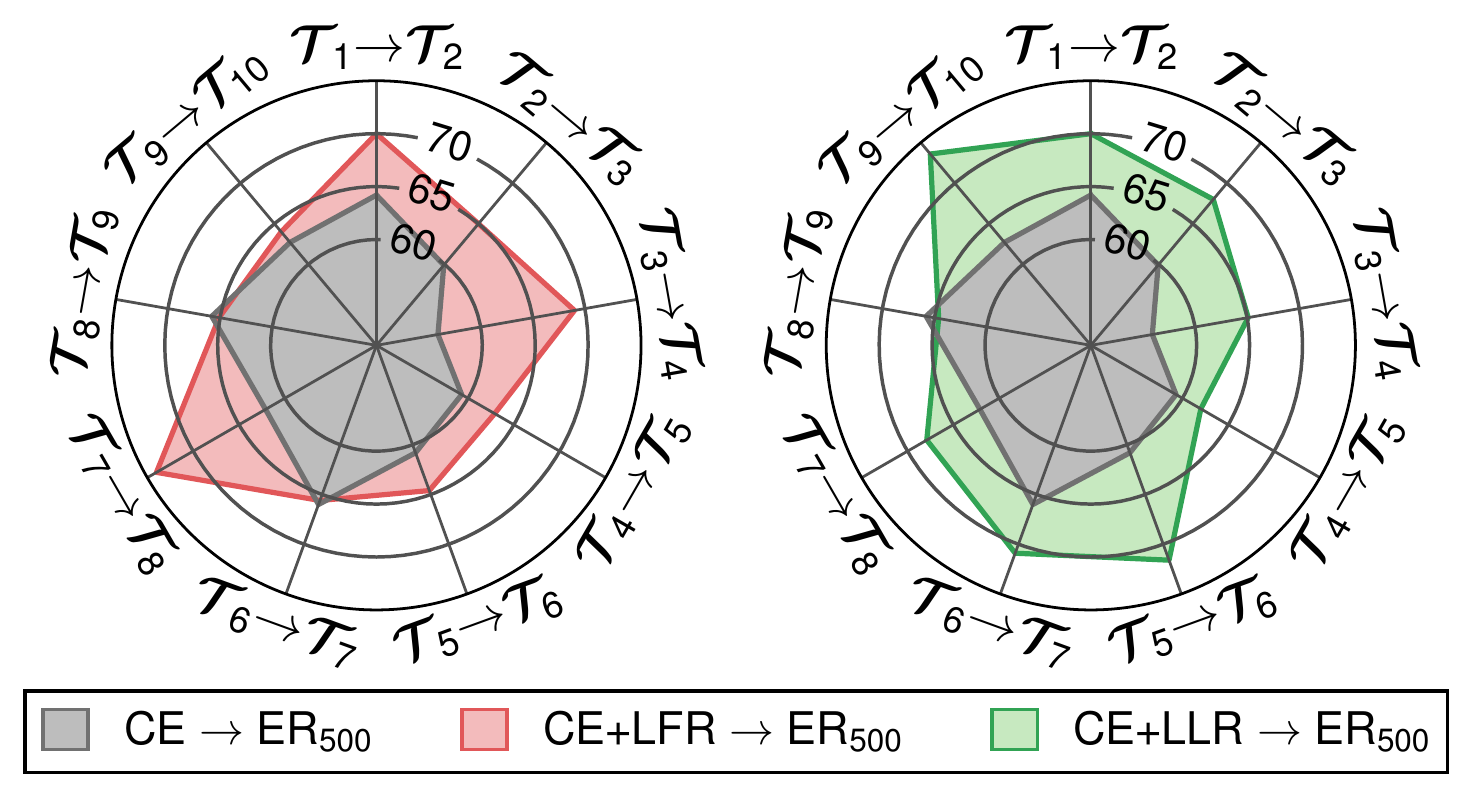}
    \caption{For nine sequences of two tasks of CIFAR-100, the final accuracy on the test-set of the former task. LFR and LLR, which encourage a \textit{stable} training regime, lead to higher retention of performance.}
    \label{fig:radarplot}
\end{figure}

In this work, we contribute again to this topic with a more targeted evaluation: given a sequence of two tasks, we deliberately drive the optimization towards a wider minimum during the former (\textit{stable regime}). Differently from~\cite{mirzadeh2020understanding}, we directly make sure the network reaches a wide minimum by introducing a tailored term in the loss function. In this regard, we evaluate two distinct approaches:
\begin{itemize}
    \item \textbf{Local Flatness Regularizer} (LFR)~\cite{xu2020adversarial}, which seeks to minimize the $\ell_1$ norm of the loss gradients w.r.t.\ a \textit{malign} example, the latter forged so that: \textit{i)} it lies in a $\varepsilon$-neighborhood centered on a given (benign) example; \textit{ii)} it maximizes the norm of the gradients. The authors prove that the robustness towards such kind of attack favorably relates to the flatness of the loss surface.
    \item \textbf{Local Linearity Regularizer} (LLR)~\cite{qin2019adversarial}, which promotes loss smoothness around the local neighborhood. As before, it consists of a regularization term that depends upon adversarial examples: on these inputs -- supposing a smooth and approximately linear loss surface -- the first-order Taylor expansion represents a good approximation of the value of the loss function; therefore, LLR simply seeks to minimize the error one would commit when using such an approximation.
\end{itemize}
\noindent We hence train the network on one task (pairing cross-entropy loss with loss surface regularization) and then measure the forgetting entailed by a CL method -- for the sake of simplicity, Experience Replay -- at the end of the second task. As a baseline, we consider the results achieved when neglecting regularization during the former task (\textit{plastic regime}): namely, it corresponds to optimizing with plain Stochastic Gradient Descent (SGD) followed by ER during the following task. We conduct this evaluation on top of nine pairs of adjacent tasks of CIFAR-100. Fig.~\ref{fig:radarplot} highlights the effect of the regularization imposed by LFR and LLR on forgetting: the results delivered by SGD (grayed out) are always upper-bounded by its regularized counterparts. 

We consider this as additional evidence showing the benefits  of a stable training regime in CL settings. It corroborates the intuition behind the effectiveness of those approaches based on self-distillation (\eg iCaRL, LwF, DER(++), etc.) -- which are known~\cite{zhang2019your,zhang2021self} to lead to such a regime.
\subsubsection*{Measuring the Flatness}
\label{sec:flatness}
\begin{figure*}[t]
    \centering
    \begin{tabular}[t]{cc}
    \includegraphics[valign=T,width=.4752\linewidth,keepaspectratio]{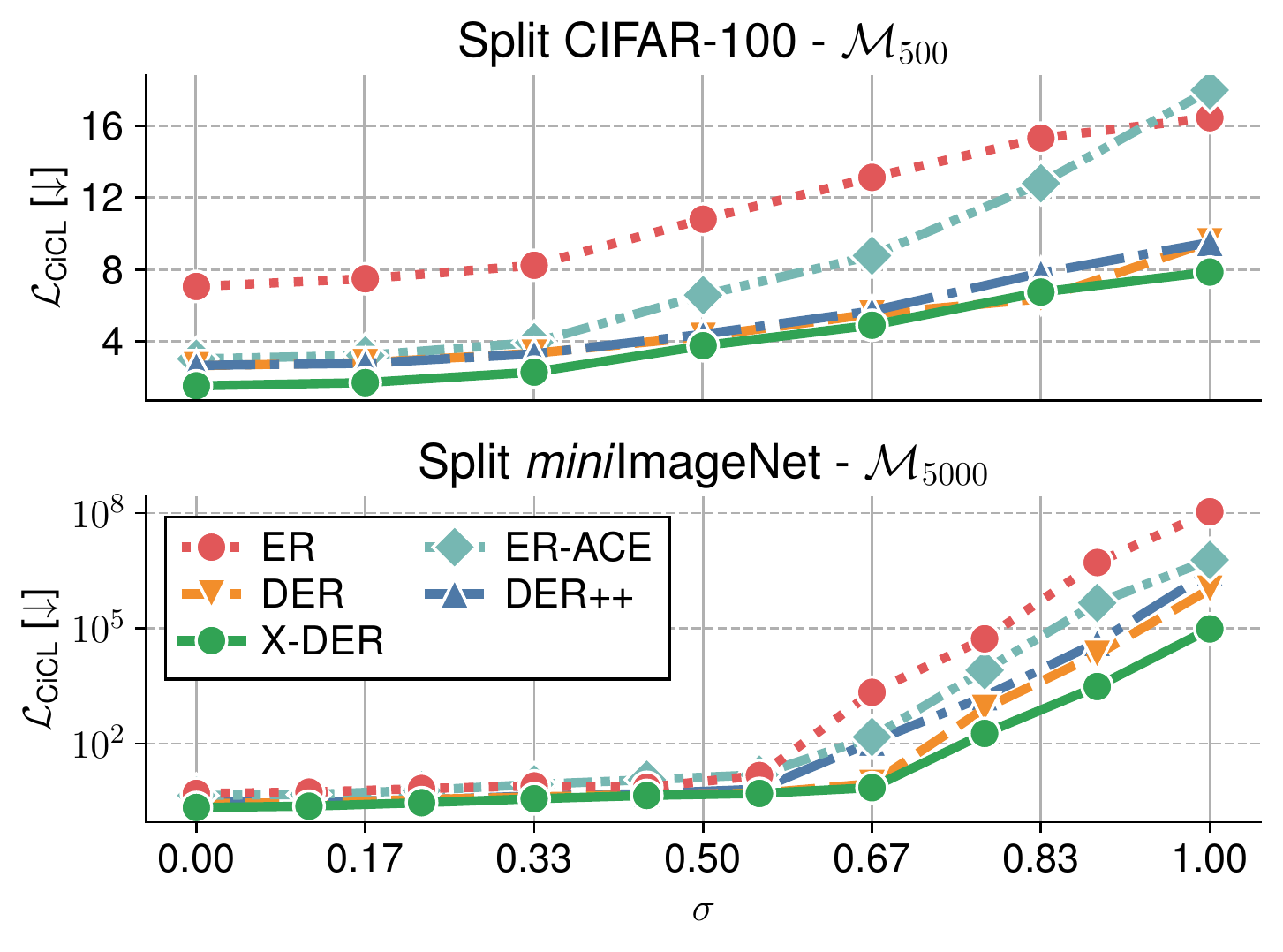} & 
    \includegraphics[valign=T,width=.4554\linewidth,keepaspectratio]{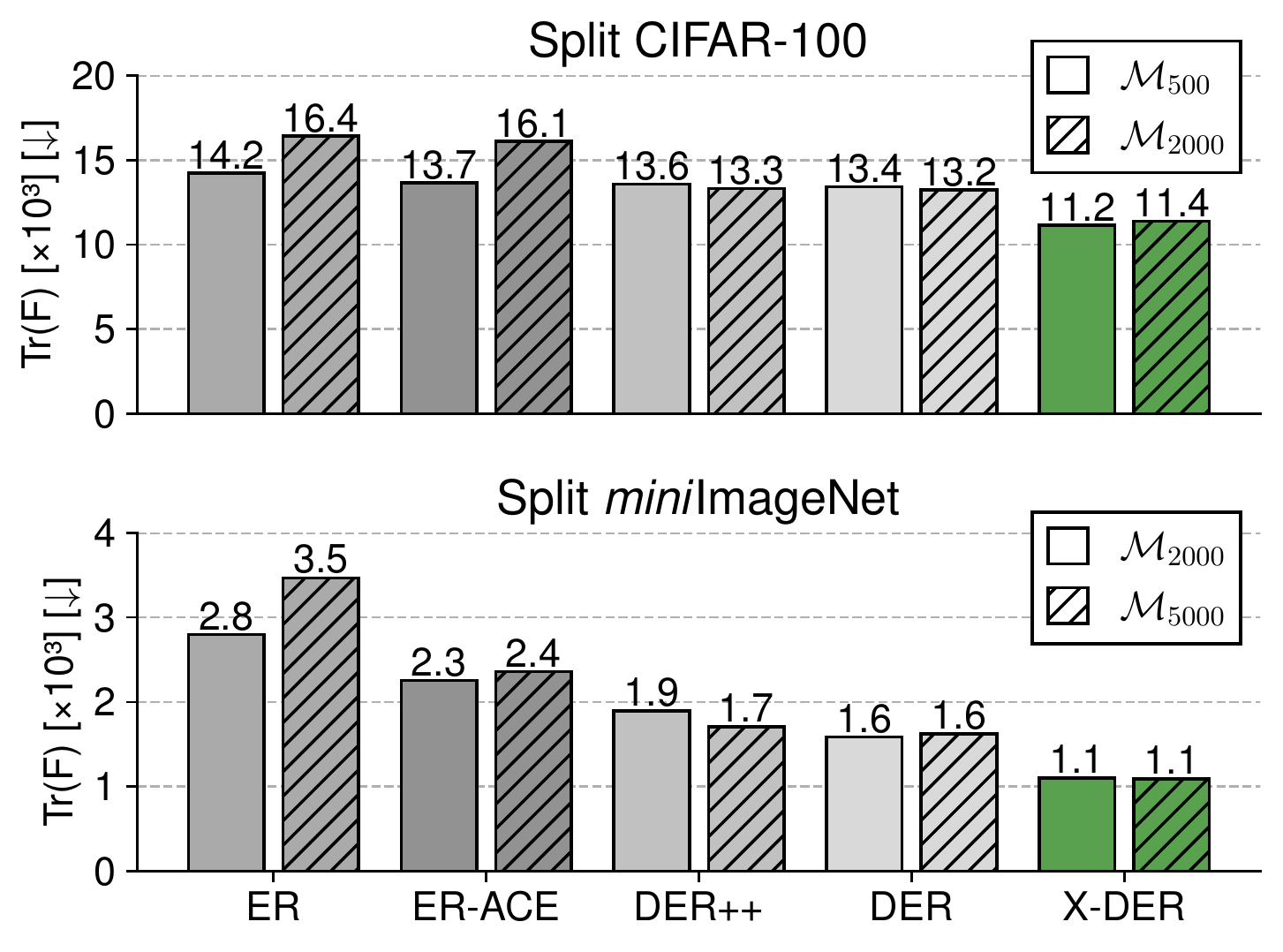} \\
    (a) & (b)
    \end{tabular}\vspace{-0.35em}
    \caption{Analysis of minima attained by distinct approaches: (a) for increasing $\sigma$, the value assumed by the loss function when Gaussian noise is applied to network weights -- tolerance towards noise is a sign of a flatter solution; (b) the sum of the eigenvalues of the Fisher Information, which models the curvature of the loss function around the solution.}
    \label{fig:flatness}\vspace{-0.35em}
\end{figure*}
Based on the above, we provide two quantitative evaluations illustrating the stability and flatness of the optima we observe for \XDER and other CL approaches.

Firstly, we measure how weight perturbations affect the \textbf{expected loss} $\mathcal{L}_{\text{CiCL}}$ w.r.t.\ to the training set~\cite{neyshabur2017exploring, keskar2017large}:
\begin{equation}
\label{eq:noisyl}
    \mathcal{L}_{\sigma} \triangleq \frac{1}{T} \sum_{i=0}^{T-1} \mathop{\mathds{E}}_{\substack{(x,y) \sim \mathcal{T}_i\\\tilde{\theta} \sim \mathcal{N}(\theta, \sigma)}}\big[\mathcal{L}(f(x; \tilde{\theta}), y)\big];
\end{equation}
specifically, we follow the hints of~\cite{li2018visualizing,neyshabur2017exploring} and weigh the perturbation according to the magnitude of parameters ($\sigma_i = \alpha |\theta_i|$), thus preventing degenerate solutions~\cite{neyshabur2017exploring}. With reference to Fig.~\ref{fig:flatness}a, it can be seen that logit-replay based models such as DER(++) and \XDER consistently preserve a lower value for Eq.~\ref{eq:noisyl}. Among them, \XDER exhibits a higher tolerance to perturbations especially in the high-$\sigma$ regime, which suggests that its attained minima are overall harder to disrupt when compared to the other methods.

A complementary flatness measure~\cite{chaudhari2017entropy, jastrzkebski2018three, keskar2017large} examines the eigenvalues of the Hessian of the overall loss function $\nabla_\theta^2\mathcal{L}_{\text{CiCL}}$. While the latter is intractable, it can be approximated by computing the empirical Fisher Information Matrix on the training set~\cite{chaudhari2017entropy,kirkpatrick2017overcoming}:
\begin{equation*}
    F \triangleq \frac{1}{T}\sum_{i=0}^{T-1}\mathop{\mathds{E}}_{(x,y)\sim \mathcal{T}_i}\big[\nabla_\theta \mathcal{L}(f(x; \theta), y) \nabla_\theta \mathcal{L}(f(x; \theta), y)^{\operatorname{T}}\big].
\end{equation*}
As in~\cite{buzzega2020dark}, we estimate the sum of the eigenvalues of $F$ through the trace of the matrix $\operatorname{Tr}(F)$, reported in Fig.~\ref{fig:flatness}b. Even according to this metric, DER(++) and \XDER reach flatter minima w.r.t.\ other approaches. Remarkably, \XDER produces lower $\operatorname{Tr}(F)$ values, suggesting that its improved accuracy can be linked to the local geometry of the loss.
\subsection{Model Explanation}
\label{sec:mexpl}
The goal of this subsection is to explore what lies behind the last layer of the network. In an attempt to investigate the quality and meaning of learned representations, we put the emphasis on three approaches (\ie ER, DER, and \XDER) and evaluate the effect of these regularization strategies on the explanations provided by the corresponding models.
\subsubsection*{Analysis of Model Explanations for Primary Targets}
Inspired by the investigation carried out in~\cite{cheng2020explaining}, we here aim to assess the quality of the \textit{visual concepts} encoded in the intermediate layers of the network. More precisely, we are interested in assessing whether the use of Knowledge Distillation leads to more refined visual concepts in regimes of catastrophic forgetting.

As stated by the authors of~\cite{cheng2020explaining}, a precise and well-established definition of visual concepts as well as the ways these can be quantified remain elusive matters. In this regard, we first assess the acquisition of task-relevant information (\ie what concerns the main subject of the image): considering the true class, does the model ascribe its score to the expected spatial location? How much does its explanation overlap with the foreground region?

To answer these questions, we take into consideration the evaluation protocol proposed in~\cite{zhang2018top}, termed \textbf{pointing game}, which was conceived to characterize the spatial selectiveness of a saliency map in the localization of target objects. The procedure is as follows: given the explanation map yielded by the learner, we check whether the point with the maximum score falls into the object region (usually defined through annotated segmentation maps); in the positive case, we have a \textit{hit} (a \textit{miss} otherwise). The localization capabilities can be finally quantified by the average Pointing Accuracy:
\begin{figure}[t]
    \centering
    \begin{tabular}{c}
    \includegraphics[width=0.90\linewidth,keepaspectratio,clip,trim=0cm 0cm 0cm 0cm]{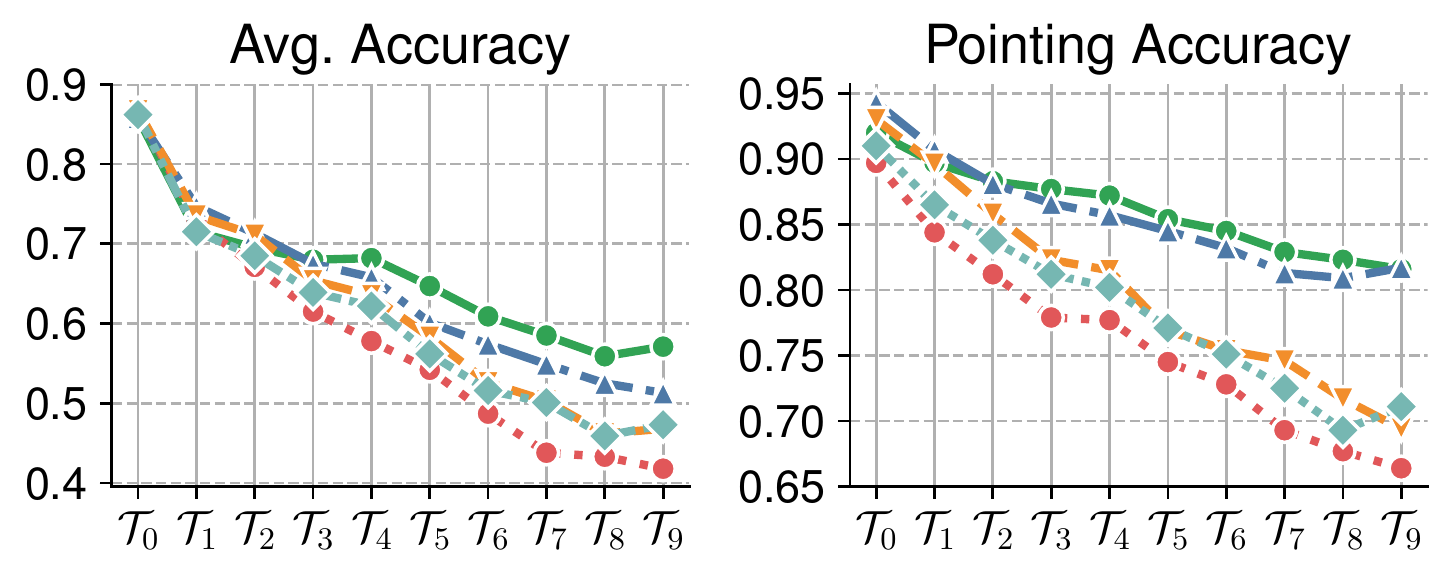} \\
    \includegraphics[width=0.9\linewidth,keepaspectratio,clip,trim=0cm 4.2cm 0cm 4.2cm]{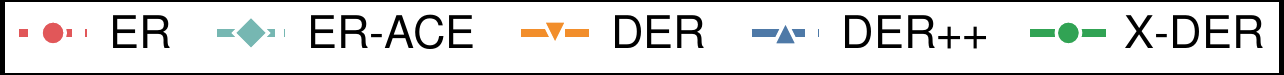}
    \end{tabular}
    \caption{On Split CUB-200, the average test-set accuracy (left) and the average pointing accuracy (right).
    }
    \label{fig:cub}
\end{figure}
\begin{equation}
\operatorname{PA} = \frac{1}{T \cdot |\mathcal{Y}|} \sum_{y=0}^{T \cdot |\mathcal{Y}| - 1} \frac{\#\operatorname{hit}_y}{\#\operatorname{hit}_y + \#\operatorname{miss}_y}. 
\end{equation}
\begin{figure*}[t]
    \centering
    \setlength{\tabcolsep}{0.15em}
    \def\arraystretch{0.5}
    \begin{tabular}{cp{0.015\textwidth}llllcp{0.015\textwidth}llll}
    Input & & $\quad \ \ \ \mathcal{T}_{0}$ & $\quad \ \ \ \mathcal{T}_{3}$ & $\quad \ \ \ \mathcal{T}_{6}$ & $\quad \ \ \ \mathcal{T}_{9}$ & Input & & $\quad \ \ \ \mathcal{T}_{0}$ & $\quad \ \ \ \mathcal{T}_{3}$ & $\quad \ \ \ \mathcal{T}_{6}$ & $\quad \ \ \ \mathcal{T}_{9}$\\
    &&&&&&&&&&\\
    & & \testprediction{1} & \testprediction{1} & \testprediction{0} & \testprediction{0} & & & \testprediction{1} & \testprediction{0} & \testprediction{1} & \testprediction{0}\\
    \cubimage{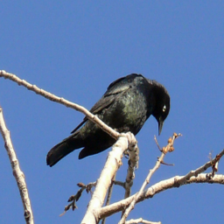} & \rotatebox{90}{\quad \ \textbf{ER}} &
    \cubgradimage{204}{er}{0} & \cubgradimage{204}{er}{3} & \cubgradimage{204}{er}{6} & \cubgradimage{204}{er}{9} & \cubimage{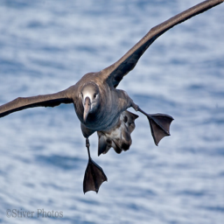} & \rotatebox{90}{\quad \ \textbf{ER}} & \cubgradimage{15}{er}{0} & \cubgradimage{15}{er}{3} & \cubgradimage{15}{er}{6} & \cubgradimage{15}{er}{9} \\
    & & \testprediction{1} & \testprediction{0} & \testprediction{0} & \testprediction{0} & & & \testprediction{0} & \testprediction{0} & \testprediction{0} & \testprediction{0}\\
    & \rotatebox{90}{\quad \ \ \textbf{DER}} & \cubgradimage{204}{der}{0} & \cubgradimage{204}{der}{3} & \cubgradimage{204}{der}{6} & \cubgradimage{204}{der}{9} & & \rotatebox{90}{\quad \ \textbf{DER}} & \cubgradimage{15}{der}{0} & \cubgradimage{15}{der}{3} & \cubgradimage{15}{der}{6} & \cubgradimage{15}{der}{9}\\
    & & \testprediction{1} & \testprediction{1} & \testprediction{1} & \testprediction{1} & & & \testprediction{1} & \testprediction{1} & \testprediction{1} & \testprediction{1} \\
    & \rotatebox{90}{\quad \textbf{\XDER}}& \cubgradimage{204}{xder}{0} & \cubgradimage{204}{xder}{3} & \cubgradimage{204}{xder}{6} & \cubgradimage{204}{xder}{9} & & \rotatebox{90}{\quad \ \textbf{\XDER}} & \cubgradimage{15}{xder}{0} & \cubgradimage{15}{xder}{3} & \cubgradimage{15}{xder}{6} & \cubgradimage{15}{xder}{9}\\
    \end{tabular}
    \caption{Considering some examples of the first task of Split CUB-200, the evolution of explanation maps as tasks progress. Green and red bars indicate whether the model predicts the right class.}
    \label{fig:attention}
\end{figure*}    
Since the datasets considered in the previous section do not come with segmentation maps, we move to \textbf{Split CUB-200}~\cite{chaudhry2019efficient,yu2020semantic}, which consists of photos depicting $200$ bird species that are split into $10$ disjoint tasks (for the sake of brevity, we provide the technical details in App.~\ref{app:cub}). On top of that, we extract explanation maps through the Grad-CAM algorithm~\cite{selvaraju2019grad} and use them to compute the resulting pointing accuracy, which is reported in Fig.~\ref{fig:cub} along with the average classification accuracy. Compared to other approaches, \XDER is less prone to forgetting the reasoning behind its predictions, as also highlighted by some qualitative examples shown in Fig.~\ref{fig:attention}.
\subsubsection*{Analysis of Model Explanations for Secondary Targets}
\begin{figure}[t]
    \centering
    \setlength{\tabcolsep}{0.25em}
    \def\arraystretch{0.5}
    \begin{tabular}{cccc}
    \includegraphics[width=0.20\linewidth,keepaspectratio,clip,trim=0cm 0cm 0cm 0cm]{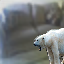} & \includegraphics[width=0.20\linewidth,keepaspectratio,clip,trim=0cm 0cm 0cm 0cm]{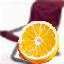} & \includegraphics[width=0.20\linewidth,keepaspectratio,clip,trim=0cm 0cm 0cm 0cm]{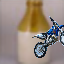} & \includegraphics[width=0.20\linewidth,keepaspectratio,clip,trim=0cm 0cm 0cm 0cm]{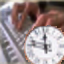}\\
    \textcolor{color1bg}{couch} & \textcolor{color1bg}{chair} & \textcolor{color1bg}{bottle} & \textcolor{color1bg}{keyboard} \\
    \textcolor{color2bg}{bear} & \textcolor{color2bg}{orange} & \textcolor{color2bg}{motorcycle} & \textcolor{color2bg}{clock} \\
    \end{tabular}
    \caption{Synthetic images obtained by stitching patches of COCO 2017 (Green) examples on top of CIFAR-100 (Red).}
    \label{fig:cocoimgs}
\end{figure}
\begin{figure}[t]
    \centering
    \def\arraystretch{0.8}
    \begin{tabular}{c}
    \includegraphics[width=0.90\linewidth,keepaspectratio,clip,trim=0cm 0cm 0cm 0cm]{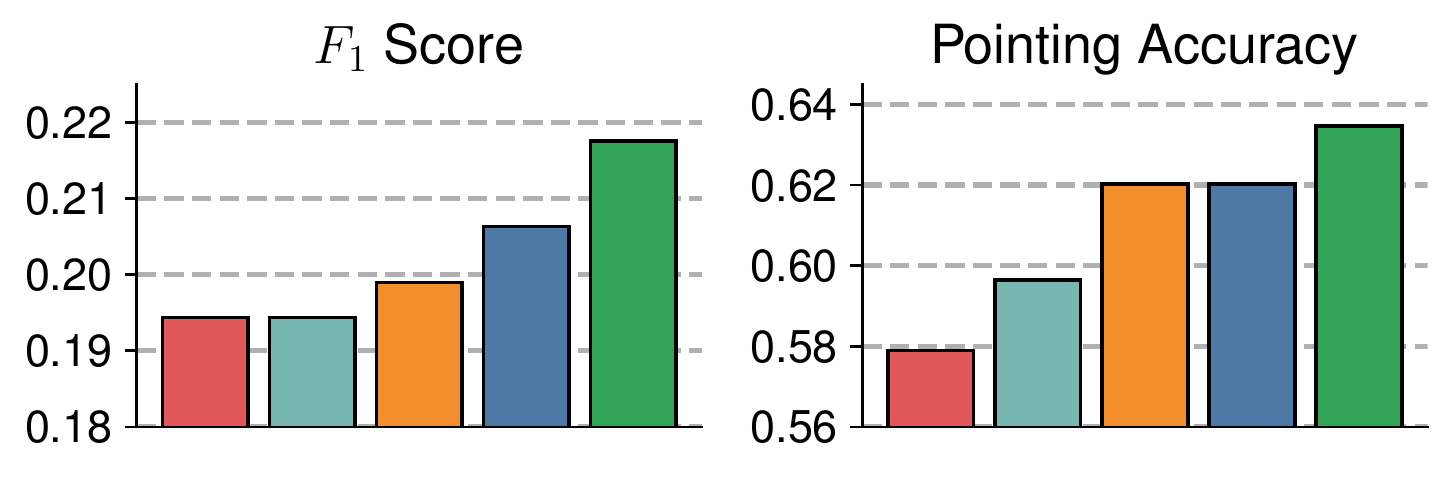} \\
    \includegraphics[width=0.9\linewidth,keepaspectratio,clip,trim=0cm 4.2cm 0cm 4.2cm]{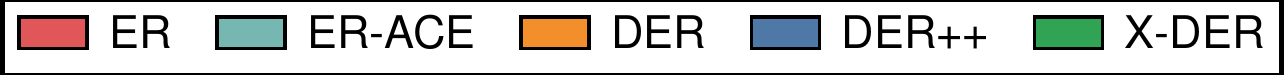}
    \end{tabular}
    \caption{Performance on secondary targets, expressed in terms of F$_{1}$ and pointing accuracy.}
    \label{fig:coco}
\end{figure}
We previously investigated the model's ability to impute its prediction to the right evidence, thus evaluating how much its representation disentangles the main subject of the image from background clutter. However, there is nothing that prevents a similar analysis on top of the secondary objects that may appear in the input image: in that sense, is secondary information limited to the main target or does it also encode insights regarding minor targets? If so, is the model able to locate these objects? If a model can correctly encode the presence of multiple objects in its response, we also expect its activation maps to convey meaningful information for the purpose of their localization.

With the aim of investigating these facets, we present a novel tailored procedure that starts with pre-training on Split CIFAR-100. We then leverage a synthetic dataset obtained by stitching small image patches from COCO 2017~\cite{lin2014microsoft} on examples of CIFAR-100\footnote{We limit to patches belonging to the set of classes shared between CIFAR-100 and COCO 2017. Moreover, we facilitate the stitching by using a $2$x-upscaled version of CIFAR-100 obtained through the CAI super-resolution API~\cite{schuler2019}. Additional details are reported in App.~\ref{app:coco}.}. As shown in Fig.~\ref{fig:cocoimgs}, these patches are cut through ground-truth segmentation masks and pasted on CIFAR-100 images to simulate secondary semantic content. Finally, we exploit the \textbf{linear evaluation} protocol~\cite{chen2020simple} to assess the representation quality of secondary targets: the parameters of the network are frozen and only a linear classifier is trained on top of its features.

We compare the performance of several methods in terms of $F_1$ score and pointing accuracy for the \textit{stitched} secondary targets. As confirmed by the results shown in Fig.~\ref{fig:coco}, the approaches relying on Knowledge Distillation perform better according to both considered metrics. Notably, \XDER stands out, thus providing a further confirmation that it can retain richer secondary information.

\section{Conclusion}
This paper reviewed Dark Experience Replay~\cite{buzzega2020dark}, our previously proposed Continual Learning method combining rehearsal and Knowledge Distillation. Upon a preliminary examination, we showed it discards valuable information about the semantic relations between old and novel classes. We also found that it suffers from a classification bias, overemphasizing the most recently acquired knowledge.

We then proposed \textbf{eXtended-DER} (\textit{a.k.a.}\ \XDER), which introduces multiple innovations (\eg memory content editing) addressing these issues: through experiments across multiple datasets, we showed that \XDER delivers higher performance and outperforms the current state of the art. Further, we presented a comprehensive analysis that goes beyond the mere final accuracy and provides an all-round validation. We in fact offered several explanations of its effectiveness against forgetting (\eg the knowledge inherent its memory buffer, the geometry of minima, the high retention of secondary information, etc.). Moreover, our results indicate that our future preparation technique favorably arranges the model to classes that are yet to be seen.

We finally envision several directions for future works: we strongly believe that overcoming the standard schema (embodied in the \textit{stability vs. plasticity} dilemma) with the guess of incoming tasks can favorably foster new ideas and advances in the field. Due to its potential applicability to a variety of CL approaches, we feel there is room for the proposals of novel strategies for mimicking future data distributions, which will be the scope of our future works.

% or


% \sigma(\ell_{gt} - \operatornamewithlimits{max}_{j\in\mathcal{Y}_\mathcal{F}} \ell_j - m


% \section{}

\ifCLASSOPTIONcompsoc
  \section*{Acknowledgments}
\else
  \section*{Acknowledgment}
\fi
The authors would like to thank Silvia Cascianelli for the constructive feedback she provided about the editing and revision of the paper.

This work has been supported in part by the InSecTT project, funded by the Electronic Component Systems for European Leadership Joint Undertaking under grant agreement 876038. The Joint Undertaking receives support from the European Union’s Horizon 2020 research and innovation programme and AU, SWE, SPA, IT, FR, POR, IRE, FIN, SLO, PO, NED and TUR. The document reflects only the author’s view and the Commission is not responsible for any use that may be made of the information it contains.
\ifCLASSOPTIONcaptionsoff
  \newpage
\fi



\bibliographystyle{IEEEtran}
\bibliography{IEEEabrv,content_rev/references.bib}

{
\small
\vspace{-3.5em}
\begin{IEEEbiography}[{\includegraphics[width=1in,height=1.25in,clip,keepaspectratio]{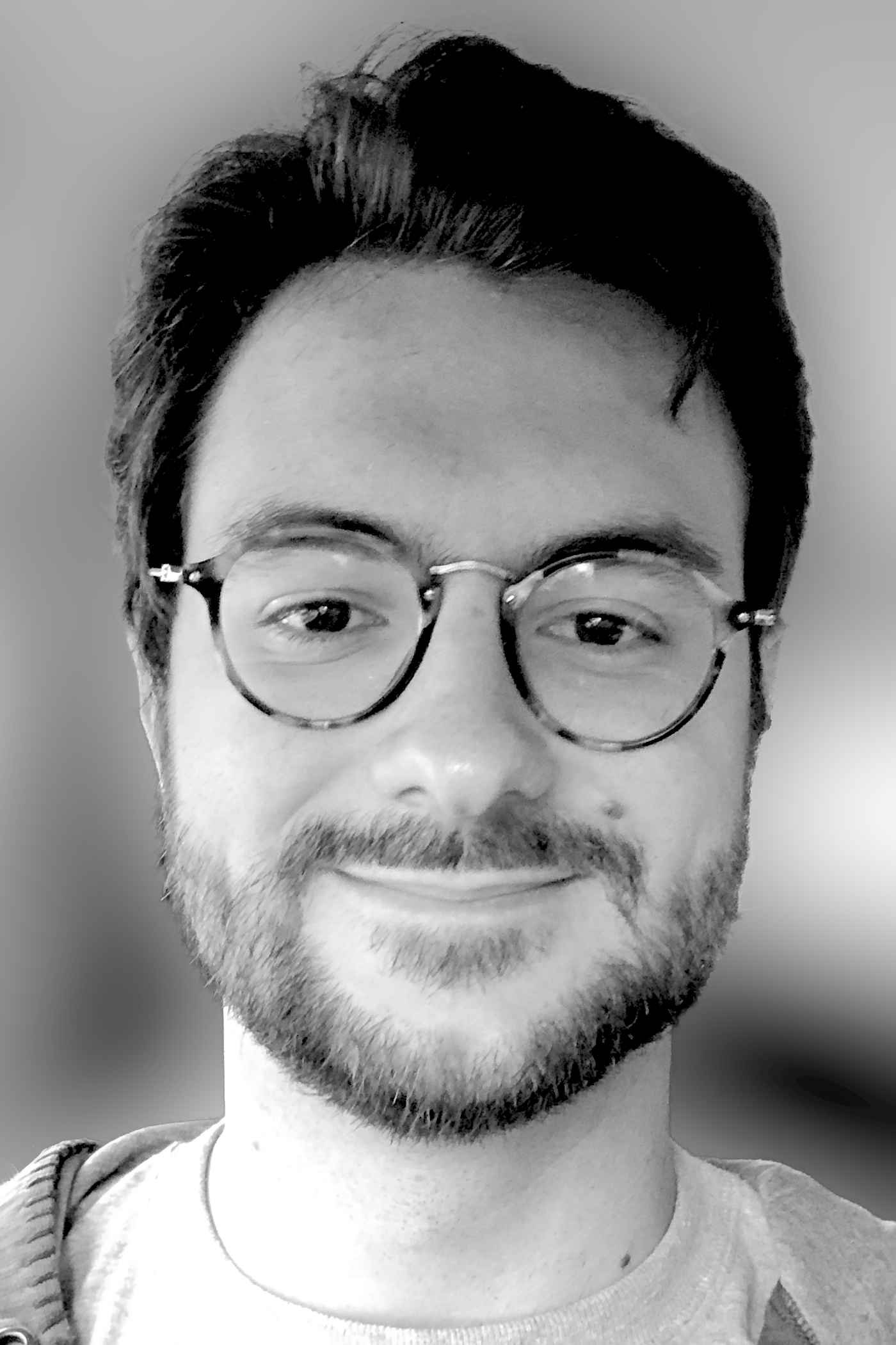}}]{Matteo Boschini}
received his master's degree in 2018 from Alma Mater Studiorum - University of Bologna, Italy. He is currently pursuing a Ph.D.\ degree at the University of Modena and Reggio Emilia, Italy, within the AImageLab research group. His research interests include machine learning, continual learning, and computer vision. He is a Grd.\ Student Member of the IEEE.
\end{IEEEbiography}
\vspace{-3.5em}
\begin{IEEEbiography}[{\includegraphics[width=1in,height=1.25in,clip,keepaspectratio]{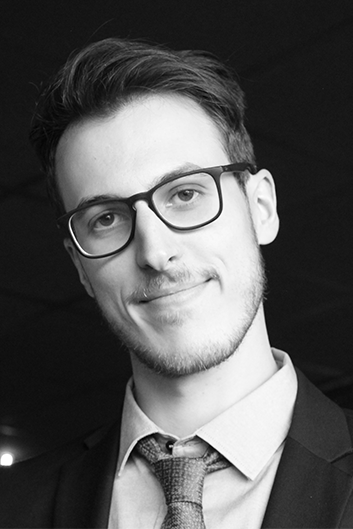}}]{Lorenzo Bonicelli}
is currently pursuing a Ph.D.\ degree at the University of Modena and Reggio Emilia, Italy, after receiving a bachelor's degree and a master's degree at the same university in 2018 and 2020 respectively. His current and past research interests include machine learning, deep learning, and the recent advances in continual learning and geometric deep learning.
\end{IEEEbiography}
\vspace{-3.5em}
\begin{IEEEbiography}[{\includegraphics[width=1in,height=1.25in,clip,keepaspectratio]{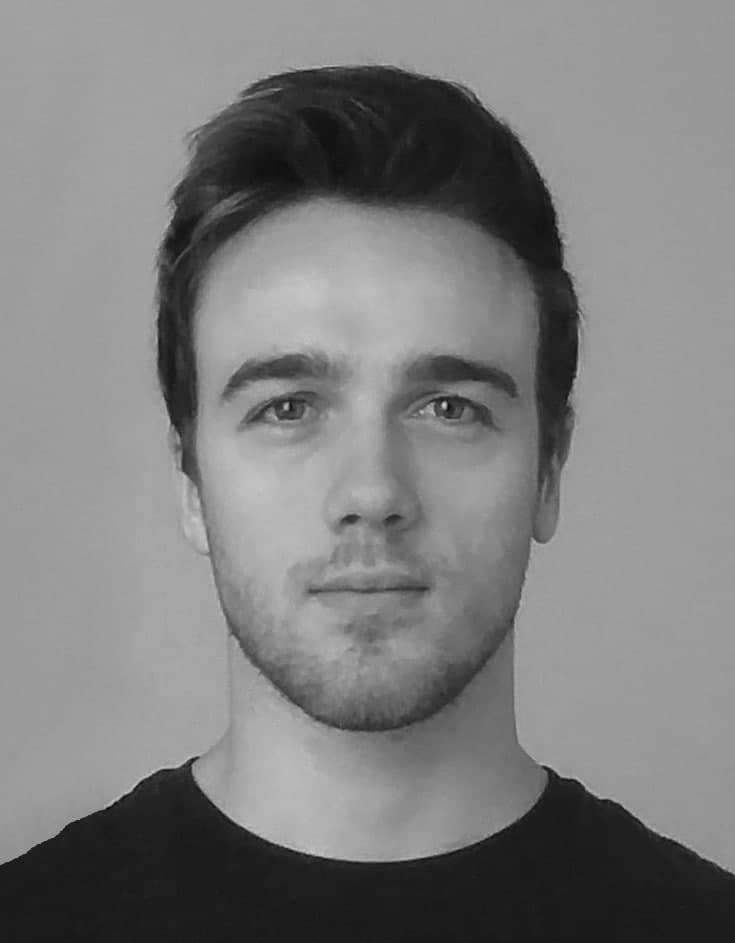}}]{Pietro Buzzega}
graduated in Computer Engineering at the University of Modena and Reggio Emilia in 2019. His thesis and later research works were mainly focused on machine learning, continual learning and anomaly detection. He is now working as a computer vision engineer at Covision Lab.
\end{IEEEbiography}
\vspace{-3.5em}
\begin{IEEEbiography}[{\includegraphics[width=1in,height=1.25in,clip,keepaspectratio]{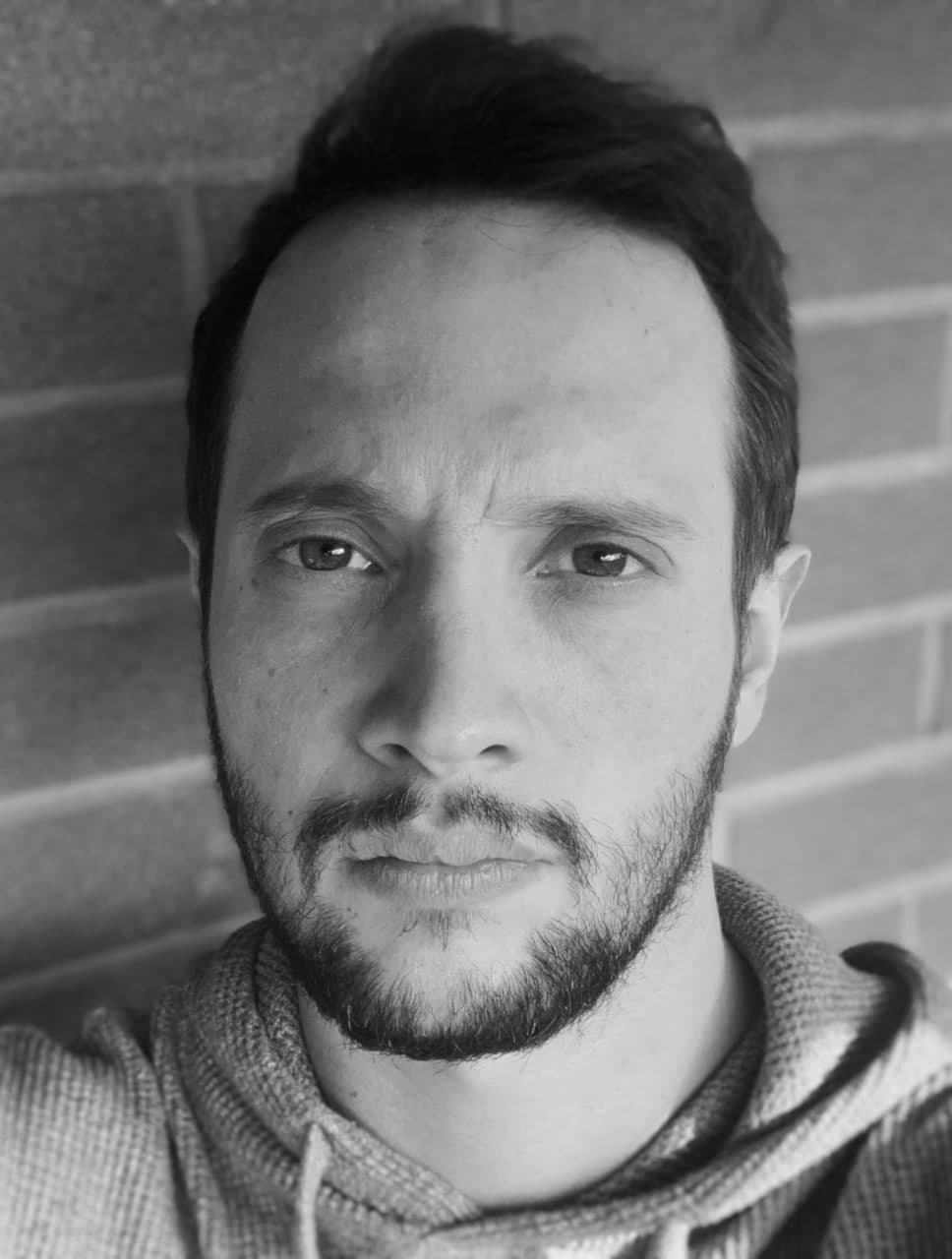}}]{Angelo Porrello}
obtained a Master's Degree in Computer Engineering in 2017 from the University of Modena and Reggio Emilia. He pursued a Ph.D. programme in ICT in the three-year period 2019-2021; currently, he is a Research Fellow within the AImageLab Group at the Department of Engineering “Enzo Ferrari”. His research interests focus on Deep Learning techniques: more precisely on Continual Learning, Re-Identification, and Anomaly Detection.
\end{IEEEbiography}
\vspace{-3.5em}
\begin{IEEEbiography}[{\includegraphics[width=1in,height=1.25in,clip,keepaspectratio]{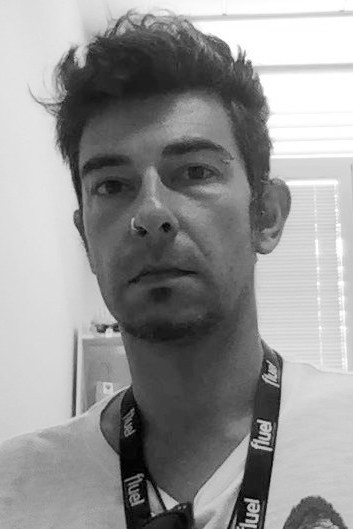}}]{Simone Calderara}
received a computer engineering master's degree in 2005 and the Ph.D. degree in 2009 from the University of Modena and Reggio Emilia, where he is currently an assistant professor within the AImageLab group. His current research interests include computer vision and machine learning applied to human behavior analysis, visual tracking in crowded scenarios, and time series analysis for forensic applications. He is a member of the IEEE.
\end{IEEEbiography}
}
\vfill

\appendices
\section{\changed{Experimental Illustration of L1}}
\label{app:l1otherexamples}
\begin{figure}[h]
    \centering
    \includegraphics[width=0.48\textwidth]{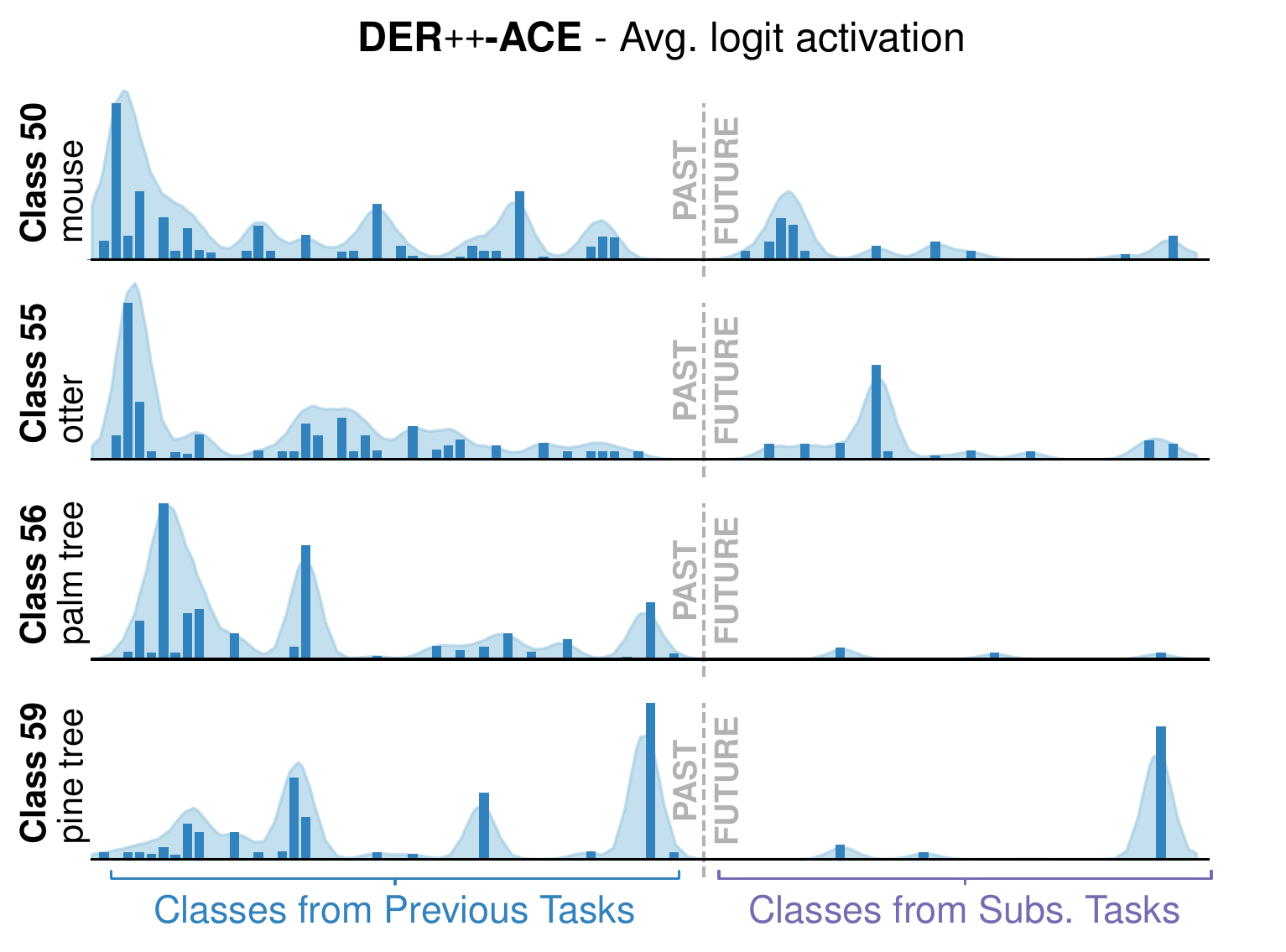}
    \caption{\changed{An illustration of DER(++)'s blindness for future past classes. For 4 different classes from the $5^\text{th}$ task of Split CIFAR-100, we depict the class-wise average predictions of the network at the end of the last task. The model has an artifactual preference towards classes shown in earlier tasks.}
    }
    \label{fig:fwdbwd_a}
\end{figure}
\subsection{\changed{DER(++) have a blind spot for future past}}
\changed{In this section, we propose an experimental proof to illustrate the first limitation presented in Sec.~\ref{sec:limitations} (L1 - DER(++) have a blind spot for future past). This limitation stems from DER(++) replaying previous network responses stored at a given point in time; these targets have shown to be more informative than plain ground-truth labels, as they portray how the network beliefs distribute across secondary classes. However, at the time examples are inserted into the memory buffer, the network has yet to encounter classes that are contained in future tasks. As a consequence, later replay cannot inform the model whether past examples hold any similarity to incoming classes. }
% \textit{b)} an insight for the class \quotationmarks{\textit{mouse}} that covers the 
    % different output landscape of the Joint Training (JT), DER\texttt{++}-ACE and \XDER. As done in \cite{mittal2021essentials}, we have masked the logit corresponding to the right class, to let the secondary information emerge.
    % Results are reported for both forward ($1\rightarrow100$) and reversed ($100\rightarrow1$) class orders. The markers highlight the top-5 activation for JT.
    % JT activation are reasonably consistent when training in either order; in sharp contrast, DER\texttt{++}-ACE fails at capturing the similarity between classes $A$ and $B$, with $B$ shown in a later task w.r.t.\ $A$. Instead, \XDER shows more consistent activation pattern when training in either order, thanks to the correct update of future past targets in replay. The markers highlight the top-5 activation for JT.

\changed{This rather complex effect is illustrated experimentally in Fig.~\ref{fig:fwdbwd_a}, which depicts for 4 classes (\textit{i.e.}; \quotationmarks{\textit{mouse}}, \quotationmarks{\textit{palm tree}}, \quotationmarks{\textit{otter}}, and \quotationmarks{\textit{pine tree}}) the average predictions produced by the model on the test set of Split CIFAR-100 at the end of the last task. As can be seen, for each class, DER++-ACE\footnote{\changed{In this experiment, we equip DER++ with Asymmetric Cross-Entropy (ACE)~\cite{caccia2021reducing}, as it compensates for the bias issue described in L2 (otherwise, the effect of bias would overshadow L1).}} emphasizes mostly the relations with classes that belong to previous tasks. Such a result cannot be attributed to a particular choice of the order in which classes are encountered: as shown in Fig.~\ref{fig:add_fwdbwd_n1} and Fig.~\ref{fig:add_fwdbwd_n2} (DER++-ACE is reported in the second row of each quadrant), when reversing that order, the model \textit{vice versa} emphasizes the relations (orange bars) with classes that in the original setting were neglected (as they belonged to subsequent tasks).}

\begin{figure}
    \begin{tabular}{c}
     \includegraphics[width=0.98\columnwidth]{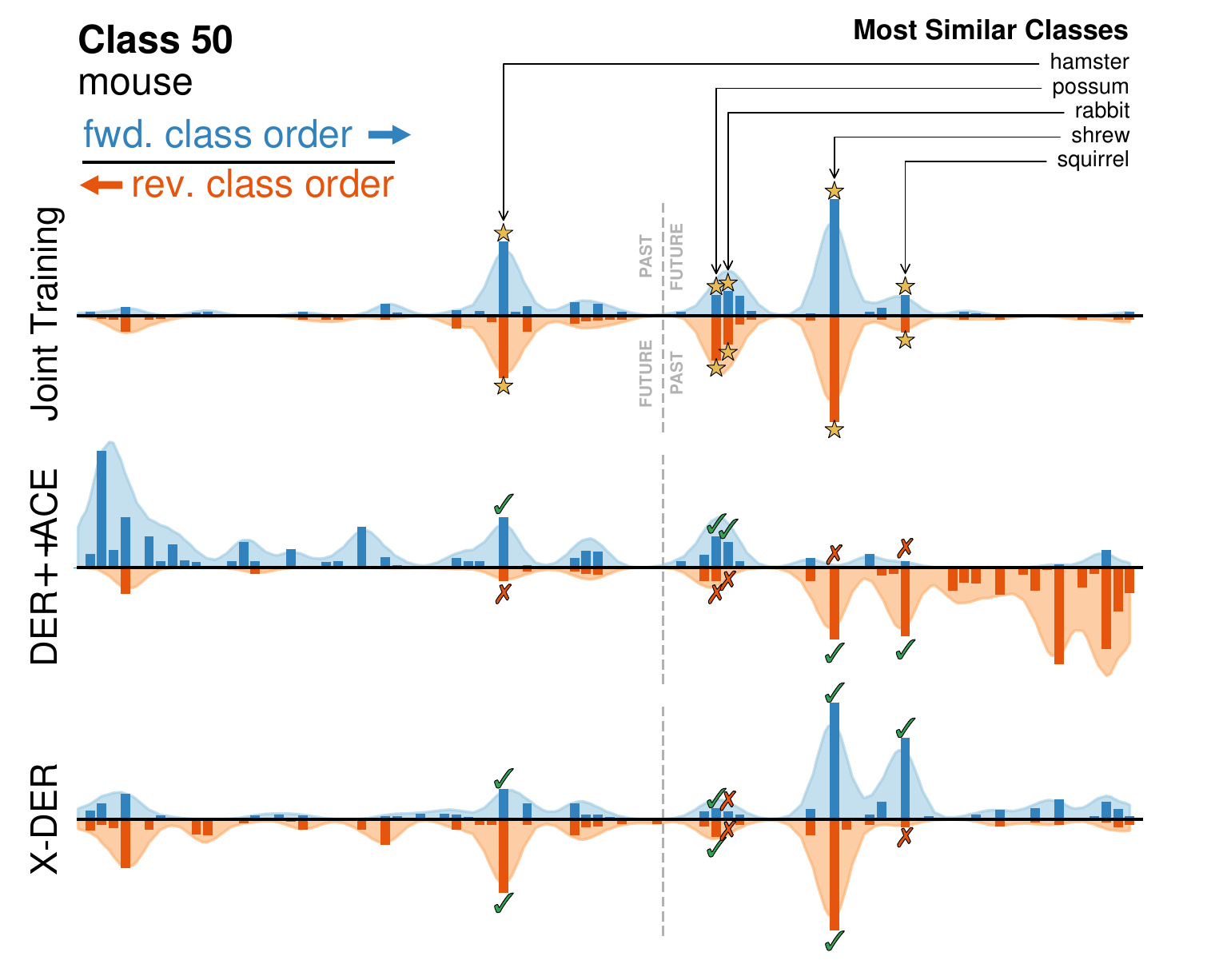}\\
     \includegraphics[width=0.98\columnwidth]{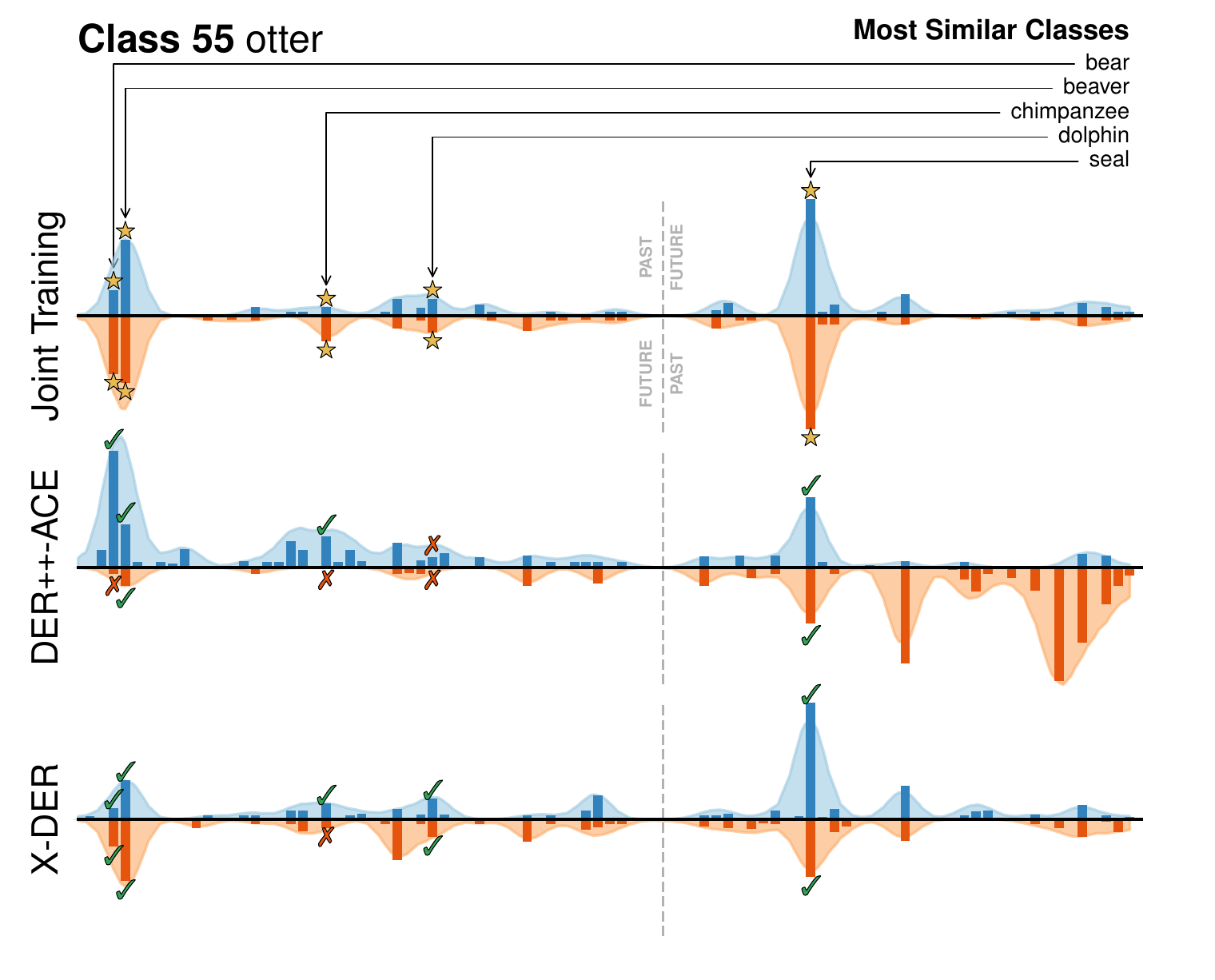}\\
    \end{tabular}
    \caption{\changed{Additional visual insights regarding the issue of future-past classes handling.}}
    \label{fig:add_fwdbwd_n1}
\end{figure}
\begin{figure}
    \begin{tabular}{c}
      \includegraphics[width=0.98\columnwidth]{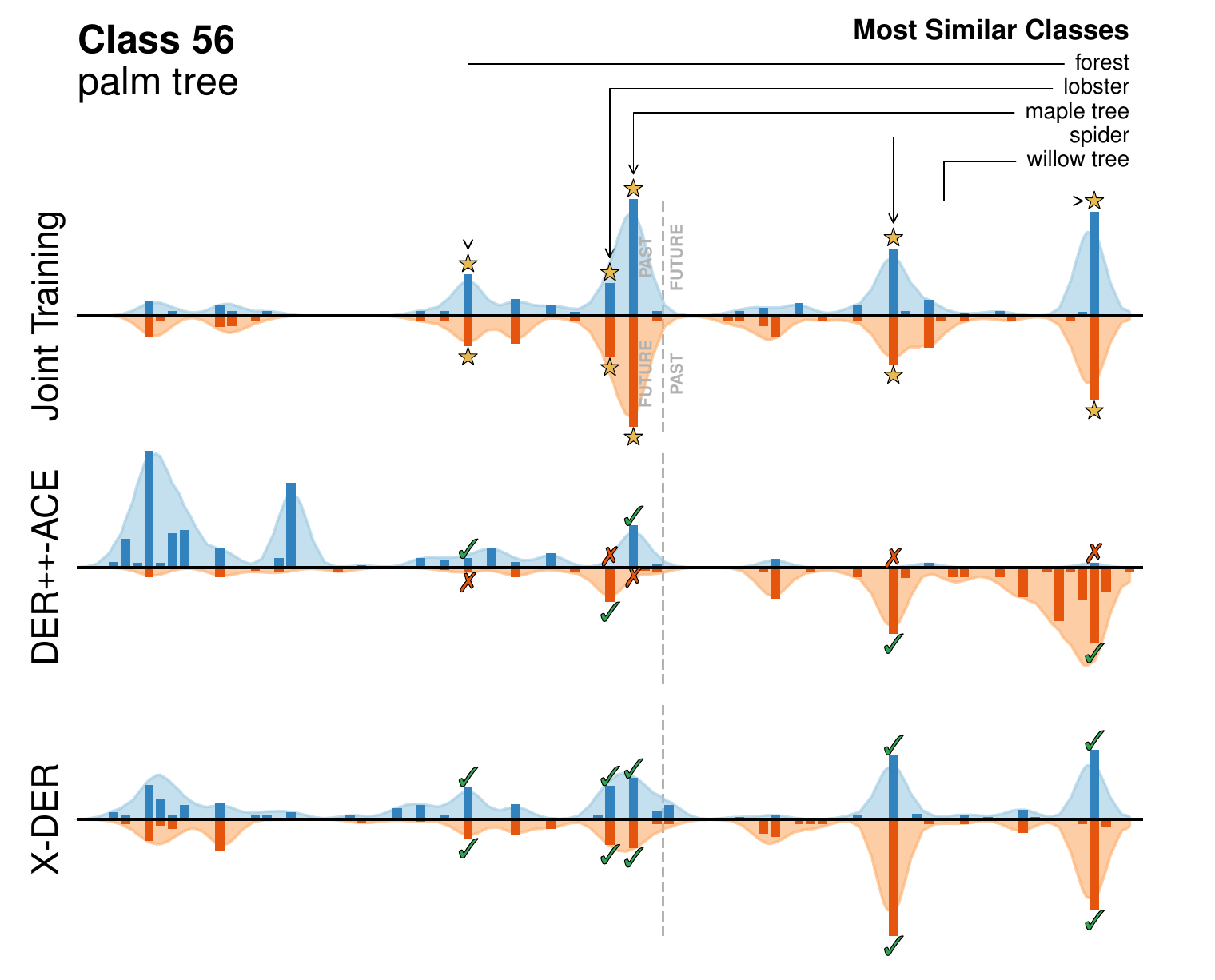}\\
      \includegraphics[width=0.98\columnwidth]{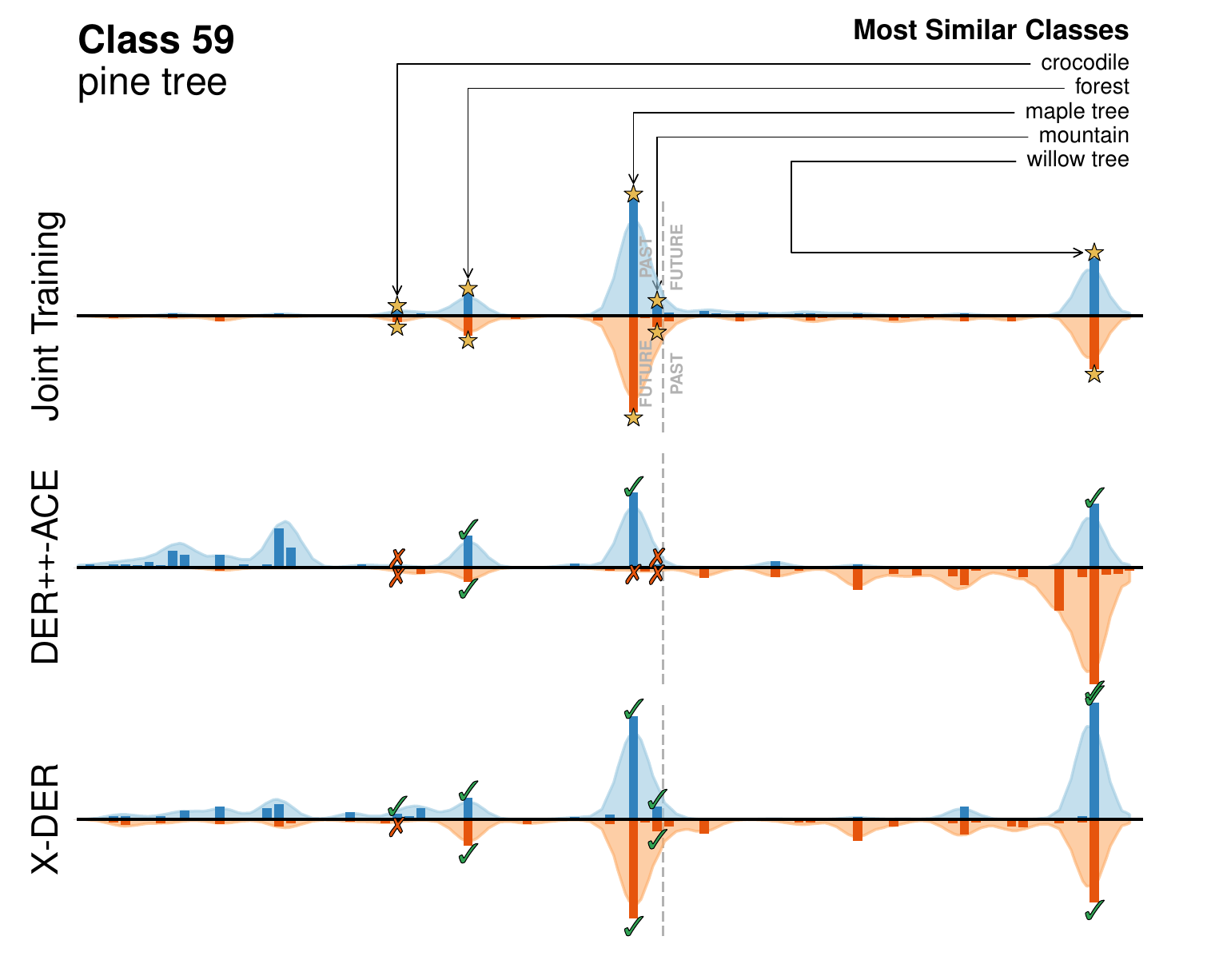}
    \end{tabular}
    \caption{\changed{Continuation of Fig.~\ref{fig:add_fwdbwd_n1}}.}
    \label{fig:add_fwdbwd_n2}
\end{figure}
\subsection{\changed{\XDER compensates this shortcoming}}
\changed{Here, we examine whether \XDER is successful in dealing with the issue discussed in the previous section. For each of the four classes examined in Fig.~\ref{fig:fwdbwd_a}, we report in Fig.~\ref{fig:add_fwdbwd_n1} and Fig.~\ref{fig:add_fwdbwd_n2} the average output distribution delivered by DER++-ACE, \XDER and the Joint Training (JT).}

\changed{Let the reader focus on the class \quotationmarks{mouse} (top left quadrant) and consider the output of JT: the latter acts upper-bound (it does not learn continually) and provides us with reference inter-class similarities. In this case, it reveals that \textit{mouse} instances mostly activate the logits associated with \textit{hamster}s and \textit{shrew}s. We provide the results for two settings: in the first one, classes are shown according to the usual order (\textit{forward order}, blue bars); in the second one, we reverse that order (\textit{backward order}, orange bars). As expected, JT does not deliver different distributions for the two orders.}

\changed{Instead, DER++-ACE shows differentiated behaviors: in forward order, it captures only the \textit{mouse}-\textit{hamster} similarity; \textit{vice versa}, it sees only the \textit{mouse}-\textit{shrew} similarity when trained in reverse order. This is a result of L1: when training on classes in a forward order, \textit{hamster}, \textit{mouse} and \textit{shrew} are learned during the $3^\text{rd}$, $5^\text{th}$ and $7^\text{th}$ incremental task of Split CIFAR-100 respectively. Accordingly, DER++-ACE replay targets include the association of \textit{mouse} examples with high logits for the \textit{hamster} class, but do not encode any meaningful information regarding the \textit{shrew} class, which is learned in a later task. The opposite happens when a reverse order is applied and \textit{shrew}s are learned before \textit{mice}.}

\changed{Instead, \XDER shows a more consistent activation pattern when training in either order and is effective in capturing similarities with future classes as well as past ones. We ascribe this peculiarity to the correct update of future past targets in replay.}
% This effect is consistent for multiple classes, as confirmed by the results in the other quadrants of Fig.~\ref{fig:add_fwdbwd}.}
\section{\changed{Pseudocode of XDER}}
\label{app:pseudocode}
\changed{In this section we provide the algorithms of X-DER (Alg.~\ref{alg:xder}) and its internal procedure that keeps update the logits stored in the memory buffer (Alg.~\ref{alg:fpastalgo}).}
\begin{figure}[h]
\centering
\hfill
\begin{algorithm}[H]
\small
\caption{- \textbf{\changed{eXtended Dark Experience Replay (X-DER)}}}
\label{alg:xder}
\begin{algorithmic}
  \STATE {\bfseries Input:} tasks $\mathcal{T}_0,\mathcal{T}_1,\dots,\mathcal{T}_{T-1}$; learner $f(\cdot;\theta)$; buffer memory $\mathcal{M}$ with bounded capacity $\#\mathcal{M}$; data augmentation procedures $\operatorname{aug}$ and $\operatorname{str\_aug}$ (\textit{strong}); loss coefficients $\alpha, \beta, \lambda, \eta$; scalar margin $m$; learning rate lr;
  \vspace{0.5em}
  \STATE $\mathcal{M} \gets \{\}$
  \FOR{\texttt{$c$} \textbf{in} $0,1,\dots,T-1$}
      \FOR{\texttt{$x,y$} \textbf{in} \texttt{$\mathcal{T}_c$}}
          \STATE \hspace{9em}\textbf{Sep. Cross Entropy (Eq. 10 and 13)}
          \vspace{-0.7em}
          \STATE $x^{m}, \ell^{m}, y^{m} \gets \operatorname{sample\_from}(\mathcal{M})$\\
          \STATE $\ell \gets f(\operatorname{aug}({x)};\theta)$
          \STATE $\ell' \gets f(\operatorname{aug}({x^{m})};\theta)$
          \STATE\tikzmark{start1}$\mathcal{L}_{\operatorname{S-CE}} \gets \operatorname{CE}(\operatorname{softmax}(\ell_{\operatorname{pr[c]}}), y) + \newline \hspace*{5em} \beta \cdot
          \operatorname{CE}(\operatorname{softmax}(\ell_{\operatorname{pa[c]}}'),y_{m})$ \hfill \tikzmark{end1}
          \vspace{0.7em}
          \STATE $\mathcal{M} \gets \operatorname{logits \_update}(\mathcal{M}, x^{m}, \ell', y^{m})$ \quad \textbf{see Alg.~\ref{alg:fpastalgo}}
          \vspace{0.3em}
          \STATE \hspace{10.5em}\textbf{Dark Experience Replay (Eq. 5)}
          \vspace{-0.7em}
          \STATE $x^{m}, \ell^{m}, y^{m} \gets \operatorname{sample\_from}(\mathcal{M})$
          \STATE\tikzmark{start2}$\ell' \gets f(\operatorname{aug}({x^{m})};\theta)$ \hfill\tikzmark{end2}
          \STATE $\mathcal{L}_{\operatorname{DER}} \gets \alpha\left\lVert \ell^{m} - \ell' \right\rVert^2_2$
          \vspace{0.7em}
          \STATE $\mathcal{M} \gets \operatorname{logits \_update}(\mathcal{M}, x^{m}, \ell', y^{m})$
          \vspace{0.3em}
          \STATE \hspace{13em}\textbf{Future preparation (Eq. 9)}
          \vspace{-0.7em}
          \STATE $x^{m}, \_, y^{m} \gets \operatorname{sample\_from}(\mathcal{M})$
          \STATE $\mathcal{X}, \mathcal{Y} \gets [x,x^{m}], [y,y^{m}]$
          \STATE\tikzmark{start3}$\mathcal{X}, \mathcal{Y} \gets [\operatorname{str\_aug}(\mathcal{X}),\operatorname{str\_aug}(\mathcal{X})], [\mathcal{Y},\mathcal{Y}]$\hfill\tikzmark{end3}
          \STATE $\mathcal{L}_{FP} \gets \mathcal{L}_{FP}(\mathcal{X},\mathcal{Y})$
          \vspace{0.3em}
          \STATE \hspace{10.5em}\textbf{Past/Future Constraints (Eq. 11)}
          \vspace{-0.7em}
          \STATE $x^{m}, \_, y^{m} \gets \operatorname{sample\_from}(\mathcal{M})$
          \STATE $\mathcal{X}, \mathcal{Y} \gets [x,x^{m}], [y,y^{m}]$
          \STATE $\ell \gets f(\operatorname{aug}(\mathcal{X});\theta)$
          \STATE $\ell_{\operatorname{gt}} \coloneqq \{\operatorname{one-hot}(y^{(n)}) \bigcdot \ell^{(n)}\}_{n=1}^{|\mathcal{X}|}$ \newline \hspace*{5em} $\triangleright$ logits of ground-truth classes
          \STATE \tikzmark{start4}$\ell_{\operatorname{pa-max}} \coloneqq \{\operatornamewithlimits{max}_{j\in\operatorname{pa}[c]}\ell_{j}\}_{n=1}^{|\mathcal{X}|}$ \newline \hspace*{5em} $\triangleright$ maximum past logits\hfill\tikzmark{end4} 
          \STATE $\ell_{\operatorname{fu-max}} \coloneqq \{\operatornamewithlimits{max}_{j\in\operatorname{fu}[c]}\ell_{j}\}_{n=1}^{|\mathcal{X}|}$ \newline \hspace*{5em} $\triangleright$ maximum future logits
          \STATE $\mathcal{L}_{\text{PFC}} \gets \max(0,\ell_{\operatorname{pa-max}} - \ell_{\operatorname{gt}} + m) + \newline \hspace*{5em} \max(0,\ell_{\operatorname{fu-max}} - \ell_{\operatorname{gt}} + m)$
          \vspace{0.7em}
          \STATE $\mathcal{L}_{\operatorname{F}} \gets \lambda \mathcal{L}_{\operatorname{FP}} + \eta \mathcal{L}_{\operatorname{PFC}}$
          \STATE $\mathcal{L}_{\operatorname{X-DER}} \gets \mathcal{L}_{\operatorname{S-CE}} + \mathcal{L}_{\operatorname{DER}} + \mathcal{L}_{\operatorname{F}}$ \quad $\triangleright$ \textbf{overall loss (Eq. 12)}
          \STATE $\theta \gets \theta - \text{lr} \cdot \nabla_{\theta}\mathcal{L}_{\operatorname{X-DER}}$ \quad $\triangleright$ \textbf{gradient step}
      \ENDFOR
      \vspace{0.1em}
      \STATE \hspace{10.5em}\textbf{Insertion  of the new items into $\mathcal{M}$}
      \vspace{-0.7em}
      \STATE $x, y \gets \operatorname{sample\_from}(\mathcal{T}_c, \text{num\_items}=\frac{\#\mathcal{M}}{c+1})$
      \STATE $\ell \gets f(\operatorname{aug}({x)};\theta)$
      \STATE\tikzmark{start5}$\mathcal{M} \gets \operatorname{remove\_items} (\mathcal{M}, \text{num\_items}=\frac{\#\mathcal{M}}{c+1})$\hfill\tikzmark{end5}
      \STATE $\mathcal{M} \gets \mathcal{M} \cup (x, \ell, y)$
  \vspace{0.3em}
  \ENDFOR
\end{algorithmic}
\TextboxSepCE[0]{start1}{end1}{}
\TextboxDer[0]{start2}{end2}{}
\TextboxFutPrep[0]{start3}{end3}{}
\TextboxFutConstr[0]{start4}{end4}{}
\TextboxMemInsert[0]{start5}{end5}{}
\end{algorithm}
\end{figure}
\begin{figure}[h]
\centering
\hfill
\begin{algorithm}[H]
\small
\caption{\changed{Update of logits of future past}}
\label{alg:fpastalgo}
\begin{algorithmic}
  \STATE {\bfseries Input:} memory buffer $\mathcal{M}$; examples $X$; logits $\ell$; labels $y$; attenuation rate $\gamma$ (default $0.8$)
  \vspace{0.5em}
  \FOR{($x$, $\ell$, $y$) \textbf{in} ($\mathcal{X}$, $\mathbf{\ell}$, $\mathcal{Y}$)}
      \STATE $i \gets \operatorname{get\_index}(\mathcal{M}, x)$
      \STATE $\ell^m \gets \operatorname{get\_stored\_logits}(\mathcal{M}, i)$
      \STATE $\ell_{\operatorname{gt}}^m \gets \operatorname{one-hot}(y)\bigcdot\ell^m$
      \STATE $\ell_{\operatorname{fpmax}} \gets \operatornamewithlimits{max}_{j\in{\operatorname{fp}[c;c]}} \ell_j$ \hspace*{1em} $\triangleright$ maximum future past logit
      \STATE $h \gets \operatorname{min} (\gamma \frac{\ell^{m}_{\operatorname{gt}}}{\ell_{\operatorname{fpmax}}}, 1)$ \hspace*{2.2em} $\triangleright$ compute the rescaling factor
      \STATE $\ell_{j}^m \gets h \cdot \ell_j \quad \forall j\in{\operatorname{fp}[c;c]}$ 
      \STATE $\mathcal{M} \gets \operatorname{store\_logits}(\mathcal{M},i,\ell^m)$ \hspace*{1em} $\triangleright$ save new logits into $\mathcal{M}$
  \ENDFOR
\end{algorithmic}
\end{algorithm}
\end{figure}
\section{Details on Split-NTU60}
\label{app:ntu}
The NTU-RGB+D dataset~\cite{shahroudy2016ntu} consists of $56,578$ video sequences depicting $60$ distinct actions, involving up to two persons. Each sample is captured by $3$ angled Kinect 2 cameras and made available as RGB videos, IR videos, Depth Map sequences and 3D skeletal data.

The latter consists of the 3D-space coordinates of $25$ body joints at each frame. We read the raw data as done by the official codebase for~\cite{yan2018spatial}, which represents each sample as a $S \times T \times J \times B$ tensor, where $S=3$ is the number of dimensions in space, $T=300$ is the number of frames per sequence, $J=25$ is the number of tracked skeletal joints and $B=2$ is the maximum number of bodies in the scene. For the sake of computational efficiency, we further down-sample the $T$ axis by interpolating it to $80$ frames. We observe that doing so yields a dramatic wall-clock time improvement, with no hindrance to the performance of the employed backbone network.

In our experiments, we adopt the following order:
\begin{itemize}
    \item \textit{task1}: put on glasses, cross hands in front, falling down, staggering, fan self, salute, throw, punch/slap, pushing, put palms together;
    \item \textit{task2}: hand waving, sneeze/cough, type on a keyboard, nod head/bow, check time (from watch), brush teeth, hugging, wipe face, eat meal, shaking hands;
    \item \textit{task3}: put on jacket, reading, take off a shoe, put on a shoe, tear up paper, reach into pocket, point to something, drink water, kicking something, sit down;
    \item \textit{task4}: headache, pat on back, play with phone/tablet, writing, stand up, back pain, walking towards, shake head, walking apart, touch pocket;
    \item \textit{task5}: take off glasses, point finger, brush hair, taking a selfie, giving object, take off jacket, take off a hat/cap, kicking, phone call, hopping;
    \item \textit{task6}: rub two hands, nausea/vomiting, jump up, clapping, drop, chest pain, neck pain, put on a hat/cap, cheer up, pick up.
\end{itemize}
We adopt the \textit{cross-subject} data-split~\cite{shahroudy2016ntu}, which reserves distinct subjects for the train and test data, resulting in $40,091$ and $16,487$ training and validation samples respectively.
\section{Backbone Architectures}
\label{app:backbone}
In this section, we present the Backbone Architectures used for the experiments in this paper. For our experiments on \textbf{Split CIFAR-100}, we follow~\cite{rebuffi2017icarl} and use \textbf{ResNet18}~\cite{he2016deep} with the original number of filters and an initial $3 \times 3$ kernel.

\begin{table}[H]
    \centering
    \rowcolors{1}{}{lightgray}
    \begin{tabular}{ccc}
        \toprule
        \multicolumn{3}{c}{\textbf{ResNet18} -- input 
        size $32 \times 32$}\\
        \midrule
        \rowcolor{white}Layer Name & Output Size & Layer Details \\
        \midrule
        \texttt{conv\_1} & $32 \times 32$ & \makecell{Conv2D $3 \times 3, 64$\\BatchNorm2D} \\
        
        \rowcolor{lightgray}\texttt{layer\_1} & $32 \times 32$ & \Gape[0pt]{\makecell{Conv2D $3 \times 3, 64$\\BatchNorm2D}$\bigg\} \times 4$} \\

\end{tabular}
\end{table}
\begin{table}[H]
    \centering
    \rowcolors{1}{}{lightgray}
    \begin{tabular}{ccc}

        \texttt{layer\_2} & $16 \times 16$ & \makecell{Conv2D $3 \times 3, 128, \text{stride}\ 2$\\BatchNorm2D \\ \makecell{Conv2D $3 \times 3, 128$\\BatchNorm2D}$\bigg\}\times 3$}\\
        
        \texttt{layer\_3} & $8 \times 8$ & \Gape[0pt]{\makecell{Conv2D $3 \times 3, 256, \text{stride}\ 2$\\BatchNorm2D \\ \makecell{Conv2D $3 \times 3, 256$\\BatchNorm2D}$\bigg\}\times 3$}}\\
        
        \texttt{layer\_4} & $4 \times 4$ & \makecell{Conv2D $3 \times 3, 512, \text{stride}\ 2$\\BatchNorm2D\\ \makecell{Conv2D $3 \times 3, 512$\\BatchNorm2D}$\bigg\}\times 3$}\\         
        
         \texttt{classifier} & $1$ & \makecell{AveragePool2D \\ $100\text{-d}$ FullyConnected Classifier} \\
         \bottomrule
    \end{tabular}
\end{table}

\noindent We refer the reader to the original paper for additional details on the way these constituting blocks compose (with particular reference to \textit{skip connections}).

For \textbf{Split \miniimagenet}, we adopt \textbf{EfficientNet-B2}~\cite{tan2019efficientnet}, whose architecture is summarized in the following.

\begin{table}[H]
    \centering
    \rowcolors{1}{}{lightgray}
    \begin{tabular}{ccc}
        \toprule
        \multicolumn{3}{c}{\textbf{EfficientNet-B2} -- input 
        size $84 \times 84$}\\
        \midrule
        \rowcolor{white}Layer Name & Output Size & Layer Details \\
        \midrule
        \texttt{stem} & $42 \times 42$ & \makecell{Conv2D $3 \times 3, 32, \text{stride}\ 2$\\BatchNorm2D} \\
        
        \texttt{block\_1} & $42 \times 42$ & \Gape[0pt]{\makecell{MBConv $3 \times 3, 16, \text{ER}\ 1$\\
                                                       MBConv $3 \times 3, 16, \text{ER}\ 1$}} \\
        
        \texttt{block\_2} & $21 \times 21$ & \makecell{MBConv $3 \times 3, 24, \text{ER}\ 6, 
        \text{stride}\ 2$\\MBConv $3 \times 3, 24, \text{ER}\ 6\big\} \times 2$} \\
        
        \texttt{block\_3} & $11 \times 11$ & \Gape[0pt]{\makecell{MBConv $5 \times 5, 48, \text{ER}\ 6,
        \text{stride}\ 2$\\MBConv $5 \times 5, 48, \text{ER}\ 6\big\} \times 2$}} \\
        
        \texttt{block\_4} & $6 \times 6$ & \makecell{MBConv $3 \times 3, 88, \text{ER}\ 6, 
        \text{stride}\ 2$\\MBConv $3 \times 3, 88, \text{ER}\ 6 \big\} \times 3$} \\
        
        \texttt{block\_5} & $6 \times 6$ & \Gape[0pt]{\makecell{MBConv $3 \times 3, 120, \text{ER}\ 6$\\MBConv $3 \times 3, 120, \text{ER}\ 6 \big\} \times 3$}} \\
        
        \texttt{block\_6} & $3 \times 3$ & \makecell{MBConv $3 \times 3, 208, \text{ER}\ 6, 
        \text{stride}\ 2$\\MBConv $3 \times 3, 208, \text{ER}\ 6 \big\} \times 4$} \\
        
        \texttt{block\_7} & $3 \times 3$ & \Gape[0pt]{\makecell{MBConv $3 \times 3, 352, \text{ER}\ 6$ 
        \\MBConv $3 \times 3, 352, \text{ER}\ 6$}} \\
        
        \texttt{head} & $3 \times 3$ & \makecell{Conv2D $1 \times 1, 1408$\\BatchNorm2D} \\
        
         \texttt{classifier} & $1$ & \Gape[0pt]{\makecell{AveragePool2D \\ $100\text{-d}$ FullyConnected Classifier}} \\
         \bottomrule
    \end{tabular}
\end{table}

\noindent With \textit{MBConv}, we indicate the mobile inverted bottleneck building block introduced by MobileNet~\cite{sandler2018mobilenetv2,tan2019mnasnet}, which is structured as follows.

\begin{table}[H]
    \centering
    \rowcolors{1}{}{lightgray}
    \begin{tabular}{ccc}
        \toprule
        \multicolumn{3}{c}{\textbf{MBConv} $k \times k, n_f, \text{ER}\ e, \text{stride}\ s$ -- input size $m \times m$, features $n_{i}$ }\\
        \midrule
        \rowcolor{white}Layer Name & Output Size & Layer Details \\
        \midrule
        \texttt{expansion}& $m \times m$ & \makecell{\makecell{Conv2D $1 \times 1, n_{i} \cdot e$\\BatchNorm2D}$\bigg\}\ \text{if}\ e > 1$}\\
        \texttt{depthwise\_conv} & $m/s \times m/s$ & \Gape[0pt]{\makecell{Conv2D $k \times k, n_{if}, \text{stride}\ s$\\
        BatchNorm2D}}\\
        \makecell{\texttt{squeeze\_and} \\\texttt{\_excitation}} & $1 \times 1$ & \makecell{AveragePool2D\\Conv2D $1 \times 1, n_{if} \cdot 0.25$\\
        Conv2D $1 \times 1, n_{i} \cdot e$}\\
        \texttt{projection} & $m/s \times m/s$ & 
        \Gape[0pt]{\makecell{Conv2D $1 \times 1, n_{f}$\\
        BatchNorm2D}}\\
        \bottomrule
    \end{tabular}
\end{table}

\noindent We refer the reader to the original paper for additional information on the way the constituting blocks compose (with particular reference to \textit{skip connections} and on how \textit{squeeze and excitation} layers concur to the computation within each \textit{MBConv} layer).

When experimenting on \textbf{Split-NTU60}, we adopt a Graph-Convolutional Neural Network capable of handling the spatio-temporal graph data presented in this dataset. Due to its performance and efficiency, our choice falls on \textbf{EfficientGCN-B0}~\cite{song2021constructing}, which we only employ on joint data representing the input skeletons.

\begin{table}[H]
    \centering
    \rowcolors{1}{}{lightgray}
    \begin{tabular}{ccc}
        \toprule
        \multicolumn{3}{c}{\textbf{EfficientGCN-B0} -- input 
        size $300 \text{(time)} \times 25 \text{(space)}$}\\
        \midrule
        \rowcolor{white}Layer Name & Output Size & Layer Details \\
        \midrule
        \texttt{input\_stem} & $300 \times 25$ & \makecell{BatchNorm2D \\ 
                                                                    SpatialConv $3 \times 3, 64$\\
                                                                    TempConv $5 \times 5, 64$} \\
        
        \texttt{input\_block\_0} & $300 \times 25$ & \Gape[0pt]{\makecell{ 
                                                                    SpatialConv $3 \times 3, 48$\\
                                                                    AttentionLayer}} \\ 
        
        \texttt{input\_block\_1} & $300 \times 25$ & \makecell{ 
                                                                    SpatialConv $3 \times 3, 32$\\ 
                                                                    AttentionLayer} \\ 
        
        \texttt{main\_block\_0} & $150 \times 25$ & \Gape[0pt]{\makecell{ 
                                                                    SpatialConv $3 \times 3, 64$\\ 
                                                                    TempConv $5 \times 5, 64, \text{stride}\ 2$\\
                                                                    AttentionLayer}}\\ 
        
        \texttt{main\_block\_1} & $75 \times 25 $ & \makecell{ 
                                                                    SpatialConv $3 \times 3, 128$\\
                                                                    TempConv $5 \times 5, 128, \text{stride}\ 2$\\
                                                                    AttentionLayer}\\ 
        
         \texttt{classifier} & $1$ & \Gape[0pt]{\makecell{AveragePool3D \\ $60\text{-d}$ FullyConnected Classifier}} \\
         \bottomrule
    \end{tabular}
    \label{tab:my_label}
\end{table}

\noindent For additional details on the implementation of Attention, Spatial and Temporal Convolutions in this backbone, we kindly refer the reader to the original paper.
\section{Additional training details}
\subsection{Data Augmentation Techniques}
In our experimental section, all competitors make use of data augmentations on both the input stream of data and (independently~\cite{buzzega2020rethinking}) the memory buffer.
For Split CIFAR-100 and \miniimagenet, we apply a full-size random crop after 4-pixel padding, followed by random horizontal flip and normalization.
Differently, for Split NTU-60, we follow~\cite{li2017skeleton} and apply random Gaussian noise to the 3D coordinates of the skeletal joints with $(\mu=0.01, \sigma=0.02)$, followed by random rotation to the 3D coordinates of the skeletal joints about the x-, y- and z-axis between $-30^{\circ}$ and $30^{\circ}$. Finally, all data points are normalized.
\subsection{Hyperparameter Choice}
\label{app:hyper}
The following is a full list of all the considered hyperparameter values for the experiments of Tab.~\ref{table:faa}, with the chosen configuration highlighted in bold. Our selection is the result of a grid search performed on a validation set obtained by sampling $10\%$ of the training set.

To ensure fairness, we exclude the number of epochs, the \textit{lr}-schedule steps, the batch size and the replay batch size from the list of parameters; instead, we fix them on a per-dataset basis for all evaluated methods.
\begin{tcolorbox}[colback=gray!65,halign=center, leftrule=0pt, rightrule=0pt, arc=0mm, toprule=0pt, bottomrule=0pt, bottom=0pt, top=0pt, boxrule=0pt,colframe=gray!15]\noindent\textbf{Split CIFAR-100}\\\end{tcolorbox}
\begin{description}
\item[SGD:] lr: [\textbf{0.03}, 0.1], mom: [\textbf{0}, 0.9], wd: [\textbf{0}, 1e-05]
\item[LwF.MC:] lr: [0.01, \textbf{0.03}, 0.1, 0.3], mom: [\textbf{0}, 0.9], wd: [0, 1e-05, \textbf{5e-04}]
\end{description}
\begin{tcolorbox}[colback=gray!25,halign=center, leftrule=0pt, rightrule=0pt, arc=0mm, toprule=0pt, bottomrule=0pt, bottom=0pt, top=0pt, boxrule=0pt,colframe=gray!15]\noindent$\mathcal{M}_{500}$\end{tcolorbox}
\begin{description}
\item[ER:] lr: [0.03, \textbf{0.1}], mom: [\textbf{0}, 0.9], wd: [\textbf{0}, 1e-05]
\item[GDumb:] lr$_{\text{max}}$: [\textbf{0.05}], lr$_{\text{min}}$: [\textbf{5e-04}], $\alpha_{\text{cutmix}}$: [\textbf{1}], epochs$_{\text{fitting}}$: [\textbf{250}, 50], mom: [0, \textbf{0.9}], wd: [\textbf{1e-06}]
\item[ER-ACE:] lr: [\textbf{0.03}, 0.3], wd: [\textbf{0}, 1e-05], mom: [\textbf{0}, 0.9]
\item[RPC:] lr: [0.01, \textbf{0.1}], wd: [\textbf{0}, 1e-05], mom: [\textbf{0}, 0.9]
\item[BiC:] $\tau$: [\textbf{2}], epochs$_{\text{BiC}}$: [\textbf{250}], lr: [\textbf{0.03}, 0.1, 0.3], wd: [\textbf{0}, 1e-05], mom: [\textbf{0}, 0.9]
\item[iCaRL:] lr: [0.03, 0.1, \textbf{0.3}], wd: [0, \textbf{1e-05}], mom: [\textbf{0}, 0.9]
\item[LUCIR:] $\lambda_{\text{base}}$: [\textbf{5}], mom: [0, \textbf{0.9}], $k$: [\textbf{2}], epochs$_{\text{fitting}}$: [0, \textbf{20}], lr: [\textbf{0.03}, 0.1, 0.3], lr$_{\text{fitting}}$: [\textbf{0.01}], $m$: [\textbf{0.5}], wd: [\textbf{0}, 1e-05]
\item[DER:] $\alpha$: [0.1, \textbf{0.3}, 0.5], lr: [\textbf{0.03}, 0.1, 0.3], wd: [\textbf{0}, 1e-05], mom: [\textbf{0}, 0.9]
\item[DER++:] $\beta$: [\textbf{0.5}, 0.8], $\alpha$: [\textbf{0.1}, 0.2, 0.3, 0.5], lr: [\textbf{0.03}, 0.1, 0.3], wd: [\textbf{0}, 1e-05], mom: [\textbf{0}, 0.9]
\item[X-DER w/o memory update:] $m$: [\textbf{0.3}], $\beta$: [\textbf{0.8}], $\gamma$: [\textbf{0.85}], $\lambda$: [\textbf{0.05}], $\eta$: [\textbf{0.001}], $\alpha$: [\textbf{0.3}], lr: [\textbf{0.03}], $\tau$: [\textbf{5}], wd: [0, 1e-05], mom: [\textbf{0}, 0.9]
\item[X-DER w/o future heads:] $m$: [0.2, \textbf{0.7}], $\beta$: [0.5, \textbf{0.8}, 0.9], $\gamma$: [\textbf{0.85}], $\eta$: [\textbf{0.001}, 0.01], $\alpha$: [0.1, \textbf{0.3}, 0.6], lr: [\textbf{0.03}], wd: [\textbf{0}, 1e-05], mom: [\textbf{0}, 0.9]
\item[X-DER w/ CE future heads:] $m$: [\textbf{0.3}], $\beta$: [\textbf{0.5}], $\gamma$: [\textbf{0.85}], $\eta$: [\textbf{5e-03}, 0.01], $\alpha$: [\textbf{0.1}], lr: [\textbf{0.03}], wd: [\textbf{0}, 1e-05], mom: [\textbf{0}, 0.9]
\item[X-DER w/ RPC future heads:] $m$: [\textbf{0.3}], $\beta$: [\textbf{0.5}, 0.8], $\gamma$: [\textbf{0.85}], $\eta$: [\textbf{0.001}], $\alpha$: [\textbf{0.1}, 0.3, 0.7], lr: [\textbf{0.03}, 0.3], wd: [\textbf{0}, 1e-05], mom: [\textbf{0}, 0.9]
\item[X-DER:] $m$: [0.2, 0.3, \textbf{0.7}], $\beta$: [\textbf{0.8}, 0.9], $\gamma$: [\textbf{0.85}], wd: [\textbf{0}, 5e-04, 1e-05, \textbf{1e-06}], $\lambda$: [0, 0.001, \textbf{0.05}, 0.1], $\eta$: [\textbf{0.001}, 0.01], $\alpha$: [0.1, \textbf{0.3}, 0.6, 0.7], lr: [5e-03, \textbf{0.03}, 0.1, 0.3], $\tau$: [10, 2, \textbf{5}], mom: [\textbf{0}, 0.9]
\end{description}
\begin{tcolorbox}[colback=gray!25,halign=center, leftrule=0pt, rightrule=0pt, arc=0mm, toprule=0pt, bottomrule=0pt, bottom=0pt, top=0pt, boxrule=0pt,colframe=gray!15]\noindent$\mathcal{M}_{2000}$\end{tcolorbox}
\begin{description}
\item[ER:] lr: [0.03, \textbf{0.1}], mom: [\textbf{0}, 0.9], wd: [\textbf{0}, 1e-05]
\item[GDumb:] lr$_{\text{max}}$: [\textbf{0.05}], lr$_{\text{min}}$: [\textbf{5e-04}], $\alpha_{\text{cutmix}}$: [\textbf{1}], epochs$_{\text{fitting}}$: [\textbf{250}, 50], mom: [\textbf{0}, 0.9], wd: [\textbf{1e-06}]
\item[ER-ACE:] lr: [\textbf{0.03}, 0.3], wd: [\textbf{0}, 1e-05], mom: [\textbf{0}, 0.9]
\item[RPC:] lr: [0.01, \textbf{0.1}], wd: [\textbf{0}, 1e-05], mom: [\textbf{0}, 0.9]
\item[BiC:] $\tau$: [\textbf{2}], epochs$_{\text{BiC}}$: [\textbf{250}], lr: [\textbf{0.03}, 0.1, 0.3], wd: [0, 1e-05, \textbf{5e-05}, 2e-04], mom: [\textbf{0}, 0.9]
\item[iCaRL:] lr: [0.03, 0.1, \textbf{0.3}], wd: [0, \textbf{1e-05}, 5e-05], mom: [\textbf{0}, 0.9]
\item[LUCIR:] $\lambda_{\text{base}}$: [\textbf{5}], mom: [0, \textbf{0.9}], $k$: [\textbf{2}], epochs$_{\text{fitting}}$: [0, \textbf{20}], lr: [\textbf{0.03}, 0.1, 0.3], lr$_{\text{fitting}}$: [\textbf{0.01}], $m$: [\textbf{0.5}], wd: [\textbf{0}, 1e-05]
\item[DER:] $\alpha$: [0.1, \textbf{0.3}, 0.5], lr: [\textbf{0.03}, 0.1, 0.3], wd: [\textbf{0}, 1e-05], mom: [\textbf{0}, 0.9]
\item[DER++:] $\beta$: [\textbf{0.5}, 0.8], $\alpha$: [\textbf{0.1}, 0.2, 0.3, 0.5], lr: [\textbf{0.03}, 0.1, 0.3], wd: [\textbf{0}, 1e-05], mom: [\textbf{0}, 0.9]
\item[X-DER w/o memory update:] $m$: [\textbf{0.3}], $\beta$: [\textbf{0.8}], $\gamma$: [\textbf{0.85}], $\lambda$: [\textbf{0.05}], $\eta$: [\textbf{0.001}], $\alpha$: [\textbf{0.3}], lr: [\textbf{0.03}], $\tau$: [\textbf{5}], wd: [\textbf{0}, 1e-05], mom: [\textbf{0}, 0.9]
\item[X-DER w/o future heads:] $m$: [\textbf{0.2}, 0.7], $\beta$: [0.5, 0.8, \textbf{0.9}], $\gamma$: [\textbf{0.85}], $\eta$: [0.001, \textbf{0.01}], $\alpha$: [0.1, 0.3, \textbf{0.6}], lr: [\textbf{0.03}], wd: [\textbf{0}, 1e-05], mom: [\textbf{0}, 0.9]
\item[X-DER w/ CE future heads:] $m$: [\textbf{0.3}], $\beta$: [\textbf{0.5}], $\gamma$: [\textbf{0.85}], $\eta$: [\textbf{5e-03}, 0.01], $\alpha$: [\textbf{0.1}], lr: [\textbf{0.03}], wd: [\textbf{0}, 1e-05], mom: [\textbf{0}, 0.9]
\item[X-DER w/ RPC future heads:] $m$: [\textbf{0.3}], $\beta$: [\textbf{0.5}, 0.8], $\gamma$: [\textbf{0.85}], $\eta$: [\textbf{0.001}], $\alpha$: [\textbf{0.1}, 0.3, 0.7], lr: [\textbf{0.03}, 0.3], wd: [\textbf{0}, 1e-05], mom: [\textbf{0}, 0.9]
\item[X-DER:] $m$: [\textbf{0.2}, 0.3, 0.7], $\beta$: [0.8, \textbf{0.9}], $\gamma$: [\textbf{0.85}], wd: [\textbf{0}, 5e-04, 1e-05, 1e-06], $\lambda$: [0, 0.001, \textbf{0.05}, 0.1], $\eta$: [0.001, \textbf{0.01}], $\alpha$: [0.1, 0.3, \textbf{0.6}, 0.7], lr: [5e-03, \textbf{0.03}, 0.1, 0.3], $\tau$: [10, 2, \textbf{5}], mom: [\textbf{0}, 0.9]
\end{description}
\begin{tcolorbox}[colback=gray!65,halign=center, leftrule=0pt, rightrule=0pt, arc=0mm, toprule=0pt, bottomrule=0pt, bottom=0pt, top=0pt, boxrule=0pt,colframe=gray!15]\noindent\textbf{Split \textit{mini}ImageNet}\\\end{tcolorbox}
\begin{description}
\item[SGD:] lr: [\textbf{0.03}, 0.1, 0.3], mom: [0, \textbf{0.9}], wd: [0, \textbf{1e-05}]
\item[LwF.MC:] lr: [0.1, \textbf{0.3}], mom: [\textbf{0}, 0.9], wd: [0, \textbf{1e-05}]
\end{description}
\begin{tcolorbox}[colback=gray!25,halign=center, leftrule=0pt, rightrule=0pt, arc=0mm, toprule=0pt, bottomrule=0pt, bottom=0pt, top=0pt, boxrule=0pt,colframe=gray!15]\noindent$\mathcal{M}_{2000}$\end{tcolorbox}
\begin{description}
\item[ER:] lr: [0.01, 0.03, \textbf{0.1}, 0.3], mom: [\textbf{0}, 0.9], wd: [\textbf{0}, 1e-05]
\item[GDumb:] lr$_{\text{max}}$: [\textbf{0.05}], lr$_{\text{min}}$: [\textbf{5e-04}], $\alpha_{\text{cutmix}}$: [\textbf{1}], epochs$_{\text{fitting}}$: [\textbf{250}, 50], wd: [0, \textbf{5e-05}, 5e-04], mom: [0.9]
\item[ER-ACE:] lr: [0.03, \textbf{0.1}], wd: [\textbf{0}, 1e-05], mom: [\textbf{0}, 0.9]
\item[RPC:] lr: [0.03, \textbf{0.1}, 0.3], wd: [\textbf{0}, 1e-05], mom: [\textbf{0}, 0.9]
\item[BiC:] $\tau$: [\textbf{2}], epochs$_{\text{BiC}}$: [\textbf{250}], lr: [0.01, \textbf{0.03}, 0.1, 0.3], wd: [0, \textbf{1e-05}], mom: [\textbf{0}, 0.9]
\item[iCaRL:] lr: [0.01, \textbf{0.1}, 0.3], mom: [0, \textbf{0.9}], wd: [\textbf{0}, 1e-04, 1e-05]
\item[LUCIR:] $\lambda_{\text{base}}$: [\textbf{5}], mom: [0, \textbf{0.9}], $k$: [\textbf{2}], wd: [0, \textbf{1e-05}, 1e-04], epochs$_{\text{fitting}}$: [\textbf{20}], lr: [0.01, 0.03, \textbf{0.1}, 0.3], lr$_{\text{fitting}}$: [0, \textbf{0.01}, 0.03], $m$: [\textbf{0.5}]
\item[DER:] $\alpha$: [0.3, \textbf{0.5}], lr: [0.03, \textbf{0.1}], wd: [\textbf{0}, 1e-05], mom: [\textbf{0}, 0.9]
\item[DER++:] $\beta$: [\textbf{0.8}], mom: [\textbf{0}, 0.9], $\alpha$: [\textbf{0.3}], lr: [0.03, \textbf{0.1}], wd: [\textbf{0}, 1e-05, 5e-05]
\item[X-DER w/o memory update:] $m$: [\textbf{0.3}], $\beta$: [0.5, \textbf{0.8}], $\gamma$: [\textbf{0.85}], $\lambda$: [\textbf{0.05}], $\eta$: [0, \textbf{0.01}, 1e-06, 1e-05], $\alpha$: [0.1, \textbf{0.3}], lr: [\textbf{0.1}], $\tau$: [\textbf{5}], wd: [\textbf{0}, 1e-05], mom: [\textbf{0}, 0.9]
\item[X-DER w/o future heads:] $m$: [\textbf{0.2}, 0.3, 0.7], $\beta$: [0.6, \textbf{0.8}], $\gamma$: [\textbf{0.85}], $\eta$: [0, 0.001, \textbf{0.01}], $\alpha$: [\textbf{0.1}, 0.3, 0.6], lr: [\textbf{0.03}, 0.1], wd: [\textbf{0}, 1e-05], mom: [\textbf{0}, 0.9]
\item[X-DER w/ CE future heads:] $m$: [0.3, \textbf{0.7}], $\beta$: [\textbf{0.8}], $\gamma$: [\textbf{0.85}], $\eta$: [0, \textbf{0.01}], $\alpha$: [\textbf{0.3}, 0.6], lr: [\textbf{0.1}], wd: [\textbf{0}, 1e-05], mom: [\textbf{0}, 0.9]
\item[X-DER w/ RPC future heads:] $m$: [\textbf{0.3}], $\beta$: [\textbf{0.5}], $\gamma$: [\textbf{0.85}], $\eta$: [0, \textbf{0.01}], $\alpha$: [\textbf{0.1}], lr: [\textbf{0.1}], wd: [\textbf{0}, 1e-05], mom: [\textbf{0}, 0.9]
\item[X-DER:] $m$: [0.2, 0.3, \textbf{0.7}], $\beta$: [0.5, \textbf{0.8}, 1, 1.3], $\gamma$: [\textbf{0.85}], $\lambda$: [\textbf{0.05}, 0.3], $\eta$: [0, 1e-04, 5e-03, \textbf{0.01}, 0.1, 1e-05, 1e-06], $\alpha$: [0.1, \textbf{0.3}, 0.5, 0.6, 0.8], lr: [0.03, \textbf{0.1}, 0.3], $\tau$: [\textbf{5}], wd: [\textbf{0}, 1e-05], mom: [\textbf{0}, 0.9]
\end{description}
\begin{tcolorbox}[colback=gray!25,halign=center, leftrule=0pt, rightrule=0pt, arc=0mm, toprule=0pt, bottomrule=0pt, bottom=0pt, top=0pt, boxrule=0pt,colframe=gray!15]\noindent$\mathcal{M}_{5000}$\end{tcolorbox}
\begin{description}
\item[ER:] lr: [0.01, 0.03, \textbf{0.1}, 0.3], mom: [\textbf{0}, 5e-04, 0.9], wd: [\textbf{0}, 1e-05]
\item[GDumb:] lr$_{\text{max}}$: [\textbf{0.05}], lr$_{\text{min}}$: [\textbf{5e-04}], $\alpha_{\text{cutmix}}$: [\textbf{1}], epochs$_{\text{fitting}}$: [\textbf{250}, 50], wd: [5e-04, \textbf{5e-05}], mom: [\textbf{0}, 0.9]
\item[ER-ACE:] lr: [0.03, \textbf{0.1}], wd: [\textbf{0}, 1e-05], mom: [\textbf{0}, 0.9]
\item[RPC:] lr: [\textbf{0.1}], wd: [\textbf{0}, 1e-05], mom: [\textbf{0}, 0.9]
\item[BiC:] $\tau$: [\textbf{2}], epochs$_{\text{BiC}}$: [\textbf{250}], lr: [\textbf{0.01}, 0.03, 0.1, 0.3], wd: [0, 1e-05], mom: [\textbf{0}, 0.9]
\item[iCaRL:] lr: [0.01, \textbf{0.1}, 0.3], mom: [0, \textbf{0.9}], wd: [\textbf{0}, 1e-04, 1e-05]
\item[LUCIR:] $\lambda_{\text{base}}$: [\textbf{5}], mom: [0, \textbf{0.9}], $k$: [\textbf{2}], wd: [0, 1e-04, \textbf{1e-05}], epochs$_{\text{fitting}}$: [\textbf{20}], lr: [0.01, 0.03, 0.1, \textbf{0.3}], lr$_{\text{fitting}}$: [0, 0.01, \textbf{0.03}], $m$: [\textbf{0.5}]
\item[DER:] $\alpha$: [0.3, \textbf{0.5}], lr: [0.03, \textbf{0.1}], wd: [\textbf{0}, 1e-05], mom: [\textbf{0}, 0.9]
\item[DER++:] $\beta$: [\textbf{0.8}], mom: [\textbf{0}, 0.9], $\alpha$: [\textbf{0.3}], lr: [0.03, \textbf{0.1}], wd: [\textbf{0}, 1e-05, 5e-05]
\item[X-DER w/o memory update:] $m$: [\textbf{0.3}], $\beta$: [0.5, \textbf{0.8}], $\gamma$: [\textbf{0.85}], $\lambda$: [\textbf{0.05}], $\eta$: [0, 0.01, 1e-05, \textbf{1e-06}], $\alpha$: [0.1, \textbf{0.3}], lr: [\textbf{0.1}], $\tau$: [\textbf{5}], wd: [\textbf{0}, 1e-05], mom: [\textbf{0}, 0.9]
\item[X-DER w/o future heads:] $m$: [0.2, \textbf{0.3}, 0.7], $\beta$: [0.6, \textbf{0.8}], $\gamma$: [\textbf{0.85}], $\eta$: [0, \textbf{0.001}, 0.01], $\alpha$: [0.1, 0.3, \textbf{0.6}], lr: [0.03, \textbf{0.1}], wd: [\textbf{0}, 1e-05], mom: [\textbf{0}, 0.9]
\item[X-DER w/ CE future heads:] $m$: [\textbf{0.3}, 0.7], $\beta$: [0.5, \textbf{0.8}], $\gamma$: [\textbf{0.85}], $\eta$: [0, \textbf{0.01}], $\alpha$: [\textbf{0.3}, 0.6], lr: [\textbf{0.1}], wd: [\textbf{0}, 1e-05], mom: [\textbf{0}, 0.9]
\item[X-DER w/ RPC future heads:] $m$: [\textbf{0.3}], $\beta$: [\textbf{0.5}, 0.8], $\gamma$: [\textbf{0.85}], $\eta$: [0, \textbf{0.01}], $\alpha$: [\textbf{0.1}, 0.5], lr: [\textbf{0.1}], wd: [\textbf{0}, 1e-05], mom: [\textbf{0}, 0.9]
\item[X-DER:] $m$: [0.2, \textbf{0.3}, 0.7], $\beta$: [0.5, \textbf{0.8}, 1, 1.3], $\gamma$: [\textbf{0.85}], $\lambda$: [\textbf{0.05}, 0.3], $\eta$: [0, 1e-04, 5e-03, \textbf{0.01}, 0.1, 1e-05, 1e-06], $\alpha$: [0.1, \textbf{0.3}, 0.5, 0.6, 0.8], lr: [0.03, \textbf{0.1}, 0.3], $\tau$: [\textbf{5}], wd: [\textbf{0}, 1e-05], mom: [\textbf{0}, 0.9]
\end{description}
\begin{tcolorbox}[colback=gray!65,halign=center, leftrule=0pt, rightrule=0pt, arc=0mm, toprule=0pt, bottomrule=0pt, bottom=0pt, top=0pt, boxrule=0pt,colframe=gray!15]\noindent\textbf{Split NTU-60}\\\end{tcolorbox}
\begin{description}
\item[SGD:] lr: [\textbf{0.1}, 0.3], wd: [0, 1e-05], mom: [0, \textbf{0.9}]
\item[LwF.MC:] lr: [0.03, \textbf{0.1}, 0.3], mom: [0, \textbf{0.9}], wd: [0, 1e-05]
\end{description}
\begin{tcolorbox}[colback=gray!25,halign=center, leftrule=0pt, rightrule=0pt, arc=0mm, toprule=0pt, bottomrule=0pt, bottom=0pt, top=0pt, boxrule=0pt,colframe=gray!15]\noindent$\mathcal{M}_{500}$\end{tcolorbox}
\begin{description}
\item[ER:] lr: [\textbf{0.1}, 0.3], wd: [0, 1e-05], mom: [0, \textbf{0.9}]
\item[GDumb:] lr$_{\text{max}}$: [\textbf{0.05}], lr$_{\text{min}}$: [\textbf{5e-04}], $\alpha_{\text{cutmix}}$: [\textbf{n/a}], epochs$_{\text{fitting}}$: [\textbf{256}], wd: [1e-04, \textbf{1e-06}], mom: [\textbf{0}, 0.9]
\item[ER-ACE:] lr: [0.01, \textbf{0.03}], wd: [\textbf{0}, 1e-05], mom: [\textbf{0}, 0.9]
\item[RPC:] lr: [\textbf{0.03}, 0.3], wd: [0, 1e-05], mom: [0, \textbf{0.9}]
\item[BiC:] $\tau$: [\textbf{2}], epochs$_{\text{BiC}}$: [\textbf{150}, 250], lr: [\textbf{0.01}, 0.03], wd: [0, 1e-05], mom: [\textbf{0}, 0.9]
\item[iCaRL:] lr: [0.1, \textbf{0.3}], wd: [0, 1e-05, \textbf{1e-06}, 5e-05], mom: [\textbf{0}, 0.9]
\item[LUCIR:] $\lambda_{\text{base}}$: [\textbf{5}], $k$: [\textbf{2}], epochs$_{\text{fitting}}$: [\textbf{20}], lr: [0.01, \textbf{0.03}, 0.1], lr$_{\text{fitting}}$: [\textbf{0.01}], $m$: [\textbf{0.5}], wd: [\textbf{0}, 1e-05], mom: [0, \textbf{0.9}]
\item[DER:] $\alpha$: [\textbf{0.3}, 0.5], lr: [0.001, 0.03, \textbf{0.1}, 0.3], wd: [0, 1e-05], mom: [0, \textbf{0.9}]
\item[DER++:] $\beta$: [0.5, \textbf{0.8}], $\alpha$: [\textbf{0.3}], lr: [\textbf{0.03}], wd: [0, 1e-05], mom: [0, \textbf{0.9}]
\item[X-DER w/o memory update:] $m$: [\textbf{0.3}], $\beta$: [\textbf{0.8}], $\gamma$: [\textbf{0.85}], $\lambda$: [\textbf{0.05}], $\eta$: [\textbf{0.03}], $\alpha$: [\textbf{0.3}], lr: [0.03], $\tau$: [\textbf{5}], wd: [\textbf{0}, 1e-05], mom: [0, \textbf{0.9}]
\item[X-DER w/o future heads:] $m$: [\textbf{0.3}], $\beta$: [\textbf{0.8}], $\gamma$: [\textbf{0.85}], $\eta$: [\textbf{0.01}], $\alpha$: [\textbf{0.3}], lr: [\textbf{0.03}], wd: [0, 1e-05], mom: [0, \textbf{0.9}]
\item[X-DER w/ CE future heads:] $m$: [\textbf{0.3}], $\beta$: [\textbf{0.8}], $\gamma$: [\textbf{0.85}], $\eta$: [\textbf{0.01}], mom: [0, \textbf{0.9}], $\alpha$: [\textbf{0.3}], lr: [0.001, \textbf{0.03}, 0.1], wd: [0, 1e-05]
\item[X-DER w/ RPC future heads:] $m$: [\textbf{0.3}], $\beta$: [\textbf{0.8}], $\gamma$: [\textbf{0.85}], $\eta$: [\textbf{0.01}], mom: [0, \textbf{0.9}], $\alpha$: [\textbf{0.3}], lr: [0.001, \textbf{0.03}, 0.1], wd: [0, 1e-05]
\item[X-DER:] $m$: [\textbf{0.3}], $\beta$: [0.5, \textbf{0.8}], $\gamma$: [\textbf{0.85}], $\lambda$: [\textbf{0.05}], $\eta$: [0, 5e-03, \textbf{0.05}], $\alpha$: [0.1, \textbf{0.3}], lr: [0.001, 0.01, 0.03, \textbf{0.1}], $\tau$: [\textbf{5}], wd: [\textbf{0}, 1e-05], mom: [0, \textbf{0.9}]
\end{description}

\section{Additional Experiments}
\label{app:additionalexp}
\subsection{\changed{Sensitivity analysis of hyperparameters}}
\label{app:sensitivity}
\begin{figure*}
    \centering
    \includegraphics[width=.66\textwidth]{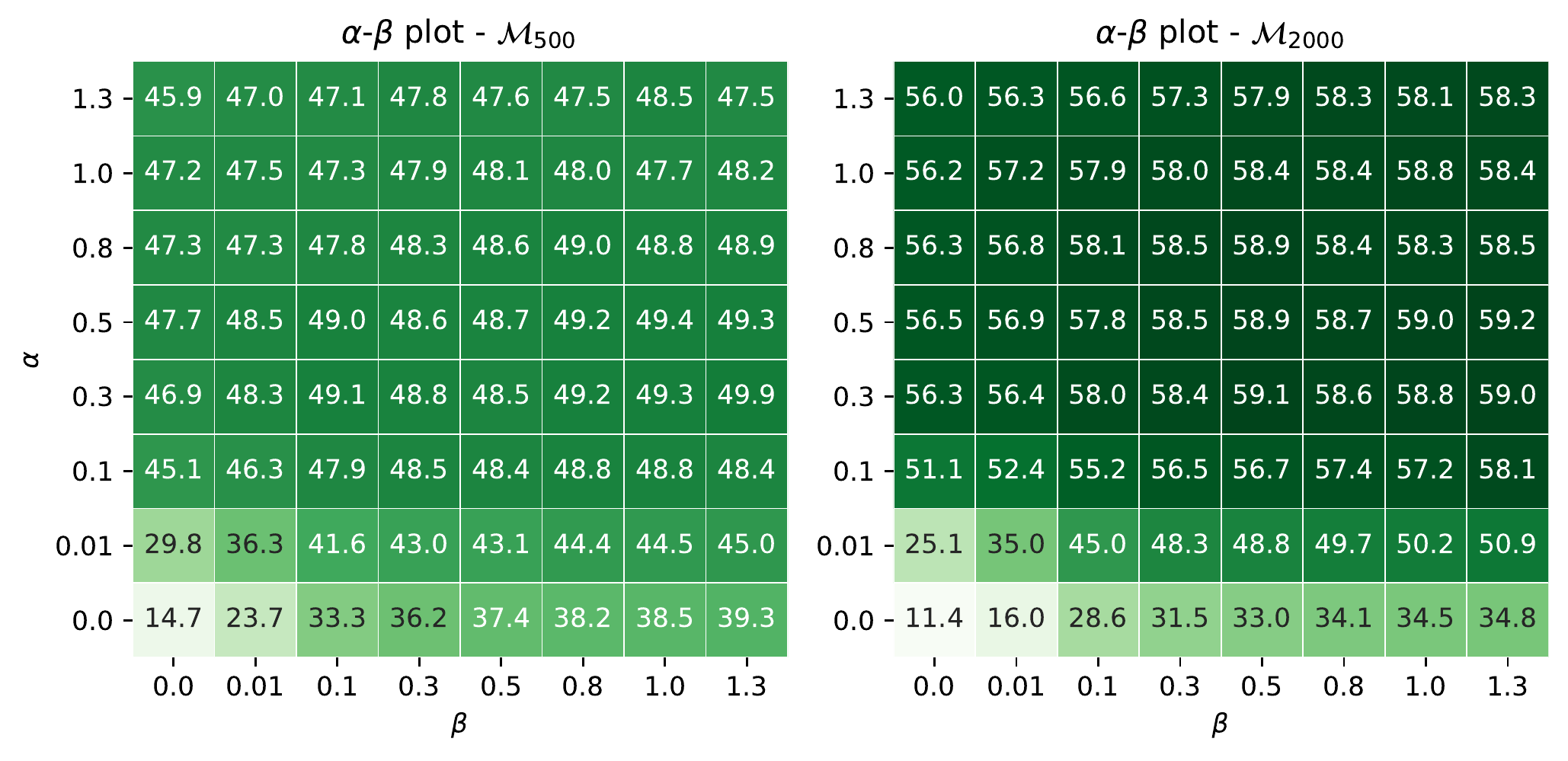}
    \caption{Joint sensitivity analysis of \XDER to parameters $\alpha$ and $\beta$ on Split CIFAR-100. Omitted parameters are taken at the best value as listed in App.~\ref{app:hyper}.}
    \label{fig:hypera}
\end{figure*}
\begin{figure*}
    \centering
    \includegraphics[width=.8\textwidth]{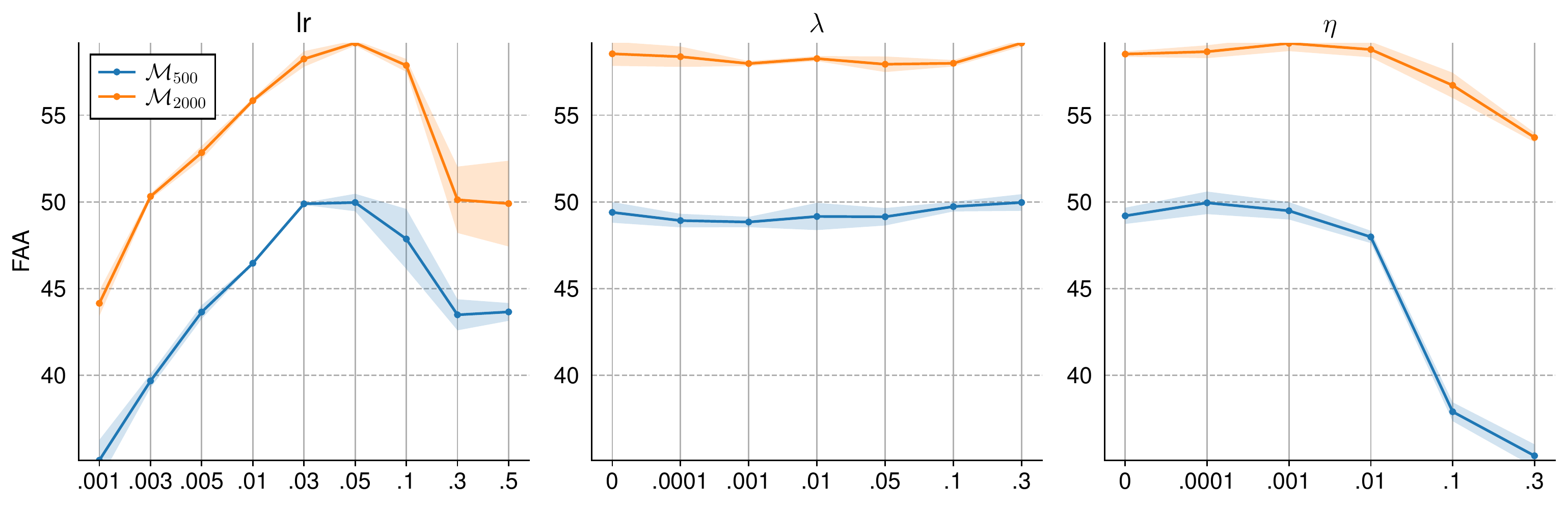}
    \caption{Sensitivity analysis of \XDER to parameters lr, $\lambda$ and $\eta$ on Split CIFAR-100.}
    \label{fig:hyperb}
\end{figure*}
\changed{In Fig.~\ref{fig:hypera} and \ref{fig:hyperb}, we investigate how the choice of different values for \textit{lr}, $\alpha$, $\beta$, $\lambda$ and $\eta$ influences the final average accuracy on Split CIFAR-100. The results in Fig.~\ref{fig:hypera} focus on the base parameters of Eq.~6 (main paper) and highlight that the model is rather robust w.r.t.\ $\beta$, provided that a high enough $\alpha$ ($\approx \geq 0.3$) is chosen. According to the results in Fig.~\ref{fig:hyperb}, $\eta$ (regulating the constraint losses of Eq.~11) must be taken low enough not to overconstrain the model; differently, $\lambda$ is stable enough, but benefits from higher values. As one may expect, the model is majorly affected by the tuning of the overall learning rate.}
\subsection{\changed{Memory and Time usage}}
\label{app:memtim}
\begin{figure*}[h]
    \centering
    \includegraphics[width=.99\textwidth]{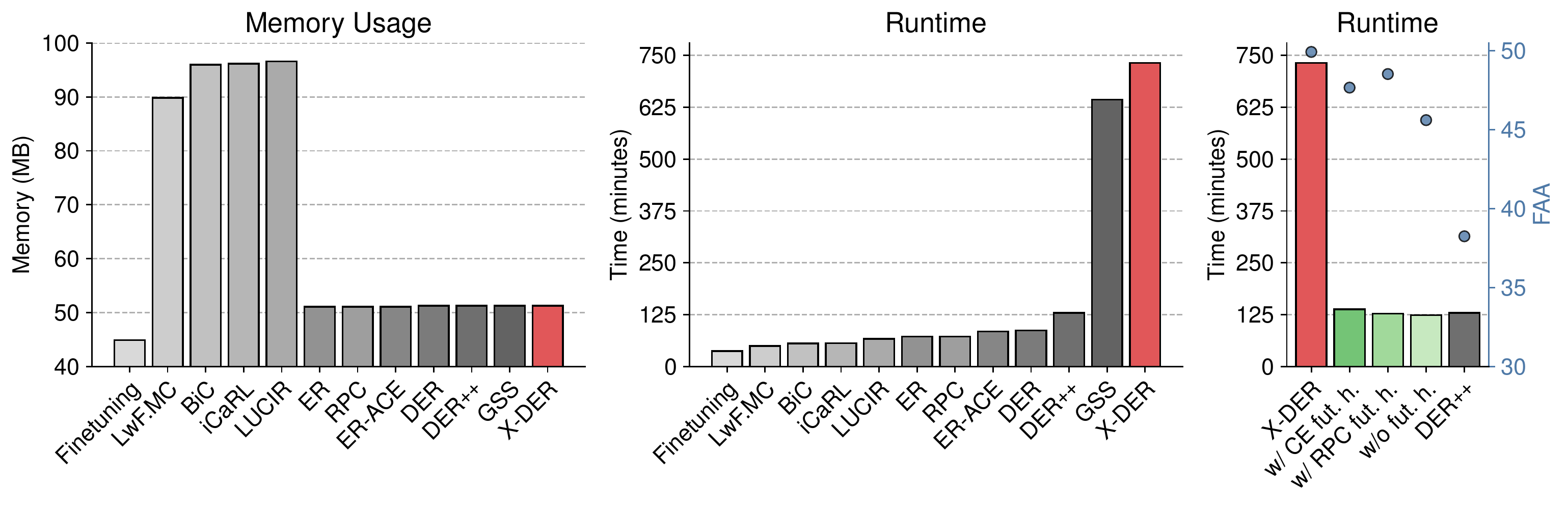}
    \vspace{-2em}
    \begin{center}
         \textcolor{white}{o}\hspace{7em} (a) \hspace{18em} (b) \hspace{14em} (c)\hspace{2em}\textcolor{white}{o}
    \end{center}
    \caption{\changed{Analysis of memory usage and execution time for the models in Sec.~5 on Split CIFAR-100, $\mathcal{M}_{500}$.}}
    \label{fig:runtimes}
\end{figure*}

\changed{
To complement our analysis in the main paper, we benchmark the models in Sec.~5 (Split CIFAR-100 - $\mathcal{M}_{500}$) in terms of memory footprint and execution time. Regarding the former (see Fig.~\ref{fig:runtimes}a), \XDER is in line with other rehearsal based methods such as DER(++) and ER(-ACE) and much less demanding than approaches using more than one backbone network (LwF, iCaRL, BiC and LUCIR).}

\changed{
Differently, Fig.~\ref{fig:runtimes}b shows that \XDER involves higher time complexity than its competitors, resulting in a $5\times$ runtime increase w.r.t.\ DER++ and being comparable to GSS~\cite{aljundi2019gradient}. We attribute such a phenomenon to the routine linked to the self-supervised future preparation (\textit{i.e.}; Eq.~8 and Eq.~9), which invokes the data augmentation procedure and forward propagation into the net several times. 
Indeed, as shown in Fig.~\ref{fig:runtimes}c, the baseline variants of \XDER introduced in Sec.~5 -- which avoid self-supervision for preparing future heads -- exhibit training times that are in line with existing approaches. Therefore, we advise using \XDER w/ RPC future heads when low training time is key, as it trades off very few points in accuracy for dramatically faster operation.}
\section{Secondary Targets: experiment details}
\label{app:coco}
\subsection{Details on Split CUB-200}
\label{app:cub}
We obtain the \textbf{Split CUB-200} dataset by splitting the $200$ classes of the \textbf{Caltech-UCSD Birds-200-2011}~\cite{wah2011caltech} dataset in $10$ $20$-fold tasks, each containing the same number of samples. The images are resized to $256\times 256$ and then randomly cropped to $224\times 224$ during training. We resize each sample to $224\times 224$ at inference. Further data augmentation is performed during training with random horizontal flip and $10$ degrees random rotation. We train the models for $70$ epochs on each tasks with \textbf{RAdam}~\cite{liu2019radam} and use ResNet-$18$ as backbone, scaling the learning rate by $0.2$ at epochs $20$, $40$, and $60$. Due to the low amount of data in CUB-200-2011, the backbone architecture is initially pre-trained on ImageNet as in~\cite{chaudhry2018efficient,cheng2020explaining}.
\subsection{Primary and secondary datasets}
Our proposed experiment on Secondary Targets in Sec.~\ref{sec:mexpl} exploits a synthetic dataset obtained by stitching secondary image patches from the COCO 2017 dataset on top of CIFAR-100 examples. The former are cut according to their ground-truth instance segmentation mask and superimposed on CIFAR images, upscaled beforehand through the CAI super-resolution API~\cite{schuler2019} to facilitate stitching.

For this evaluation, we only draw examples from the pool of classes shared by both datasets, namely: \textit{bed, cup, chair, clock, bicycle, elephant, apple, bowl, bottle, keyboard, motorcycle, train, bear, bus, mouse, couch, orange}.

The experiment is conducted on $50000$ train and $10000$ test examples generated in advance by sampling images from random pairs of the above-listed classes.
\subsection{Experimental details}
Models are initially pre-trained on the resized primary dataset (Split CIFAR-100 with images upsampled to $64\times 64$) to provide the base initialization. We then freeze the weights and only finetune the final classifier on the newly-generated dataset in a multi-label classification scenario. Specifically, we activate each logit with a sigmoid and seek to minimize the \textit{binary cross-entropy} loss function, using the SGD optimizer for $50$ epochs. After that, we evaluate the performance by means of the $F1$-score, computed on the secondary targets of the previously-generated test data.

\end{document}